\documentclass{article}

\usepackage{iclr2027_conference,times}
\iclrfinalcopy



\usepackage[utf8]{inputenc}
\usepackage[T1]{fontenc}

\usepackage{hyperref}
\usepackage{url}
\usepackage{booktabs}
\usepackage{amsfonts}
\usepackage{nicefrac}
\usepackage{microtype}
\usepackage[table]{xcolor}  %
\usepackage{colortbl}
\hypersetup{hidelinks}

\usepackage{amsmath,amssymb,amsthm,mathtools,bm}
\usepackage[font=small, skip=4pt]{caption}
\usepackage[capitalize, nameinlink]{cleveref}
\crefname{section}{Sec.}{Secs.}
\Crefname{section}{Sec.}{Secs.}
\crefname{subsection}{Sec.}{Secs.}
\Crefname{subsection}{Sec.}{Secs.}
\crefname{subsubsection}{Sec.}{Secs.}
\Crefname{subsubsection}{Sec.}{Secs.}
\crefname{table}{Tab.}{Tabs.}
\Crefname{table}{Tab.}{Tabs.}
\crefname{figure}{Fig.}{Figs.}
\Crefname{figure}{Fig.}{Figs.}
\crefname{equation}{Eq.}{Eqs.}
\Crefname{equation}{Eq.}{Eqs.}
\crefname{theorem}{Thm.}{Thms.}
\Crefname{theorem}{Thm.}{Thms.}
\crefname{proposition}{Prop.}{Props.}
\Crefname{proposition}{Prop.}{Props.}
\crefname{corollary}{Cor.}{Cors.}
\Crefname{corollary}{Cor.}{Cors.}
\crefname{definition}{Def.}{Defs.}
\Crefname{definition}{Def.}{Defs.}
\crefname{remark}{Rem.}{Rems.}
\Crefname{remark}{Rem.}{Rems.}
\crefname{algorithm}{Alg.}{Algs.}
\Crefname{algorithm}{Alg.}{Algs.}
\crefname{appendix}{App.}{Apps.}
\Crefname{appendix}{App.}{Apps.}

\usepackage{makecell}
\usepackage{pifont}
\usepackage{graphicx}
\graphicspath{{./}{../}}
\usepackage{adjustbox}
\usepackage{wrapfig}
\usepackage{float}
\usepackage{enumitem}

\usepackage{multirow}
\usepackage{algorithm}
\usepackage{algpseudocode}
\usepackage{needspace}

\makeatletter
\def\section{\@startsection{section}{1}{\z@}{-1.5ex plus -0.5ex minus -.2ex}%
  {1.1ex plus 0.3ex minus 0.2ex}{\large\sc\raggedright}}
\def\subsection{\@startsection{subsection}{2}{\z@}{-1.3ex plus -0.5ex minus -.2ex}%
  {0.6ex plus .2ex}{\normalsize\sc\raggedright}}
\makeatother
\AtBeginDocument{%
  \setlength{\abovedisplayskip}{4pt plus 2pt minus 2pt}%
  \setlength{\belowdisplayskip}{4pt plus 2pt minus 2pt}%
  \setlength{\abovedisplayshortskip}{0pt plus 2pt}%
  \setlength{\belowdisplayshortskip}{3pt plus 2pt minus 2pt}%
}

\newcommand{\cmark}{{\color{green!60!black}\ding{51}}}
\newcommand{\xmark}{{\color{red!70!black}\ding{55}}}

\newcommand{\distreal}{\textcolor{green!60!black}{Real}}
\newcommand{\distpseudo}{\textcolor{red!70!black}{Pseudo}}

\theoremstyle{plain}
\newtheorem{theorem}{Theorem}[section]
\newtheorem{proposition}[theorem]{Proposition}

\newtheorem{lemma}[theorem]{Lemma}
\crefname{lemma}{Lem.}{Lems.}
\Crefname{lemma}{Lem.}{Lems.}
\theoremstyle{definition}
\newtheorem{definition}[theorem]{Definition}
\theoremstyle{remark}
\newtheorem{remark}[theorem]{Remark}

\DeclareMathOperator{\sign}{sign}
\DeclareMathOperator{\arcsinh}{arcsinh}
\DeclareMathOperator{\arctanh}{arctanh}
\DeclareMathOperator{\arccosh}{arccosh}
\DeclareMathOperator{\Concat}{Concat}
\DeclareMathOperator*{\argmin}{argmin}
\DeclareMathOperator{\ReLU}{ReLU}

\newcommand{\LL}{\mathbb{L}}
\newcommand{\KK}{\mathbb{K}}
\newcommand{\PP}{\mathbb{P}}
\newcommand{\PV}{\mathbb{PV}}
\newcommand{\RR}{\mathbb{R}}
\newcommand{\Sn}{\mathbb{S}}
\newcommand{\KKn}{\mathbb{K}^{n}_{\mathrm{K}}}
\newcommand{\Expop}{\mathrm{Exp}}
\newcommand{\Logop}{\mathrm{Log}}
\newcommand{\PT}{\mathrm{PT}}
\newcommand{\oE}{\oplus_{\mathrm{E}}}
\newcommand{\otE}{\otimes_{\mathrm{E}}}
\newcommand{\oM}{\oplus_{\mathrm{M}}}
\newcommand{\otM}{\otimes_{\mathrm{M}}}
\newcommand{\FM}{\mathrm{FM}}
\newcommand{\KKd}[1]{\mathbb{K}^{#1}_{\mathrm{K}}}
\newcommand{\oS}{\mathbin{+_{\sigma}}}
\newcommand{\otS}{\mathbin{\cdot_{\sigma}}}
\newcommand{\oU}{\oplus_{\mathrm{U}}}
\newcommand{\otU}{\otimes_{\mathrm{U}}}
\newcommand{\gyr}{\mathrm{gyr}}
\newcommand{\sk}{\sqrt{-K}}
\newcommand{\ExpO}{\Expop_{\mathbf{0}}^{\KK}}
\newcommand{\LogO}{\Logop_{\mathbf{0}}^{\KK}}

\newcommand{\first}[1]{\textbf{#1}}
\newcommand{\second}[1]{\underline{#1}}

\newcommand{\proofref}[1]{\hyperref[#1]{[$\downarrow$]}}

\title{Klein Neural Networks++}

\author{%
  Ying Zhang\\
  University of Cambridge\\
  \texttt{yz2160@cam.ac.uk}\\
  \texttt{yingurt@outlook.com}
  \And
  Nicu Sebe\\
  University of Trento\\
  \texttt{niculae.sebe@unitn.it}
  \And
  Ziheng Chen\\
  Max Planck Institute for\\
  Intelligent Systems\\
  University of Trento\\
  \texttt{ziheng\_ch@163.com}
}

\renewcommand{\floatpagefraction}{0.9}
\renewcommand{\bottomfraction}{0.7}

\begin{document}

\maketitle
\lhead{Preprint}
\vspace{-0.25in}

\addtocontents{toc}{\protect\setcounter{tocdepth}{-1}}

\begin{abstract}
Hyperbolic neural networks (HNNs) have shown significant
advantages in representing hierarchical and tree-structured
data through their exponential volume growth. However,
existing methods rely almost exclusively on the Poincar\'{e}
ball and Lorentz hyperboloid, while the Klein model, another
classical representation, has long been unexplored.
The Klein model has Euclidean line-segment geodesics and a
closed-form Einstein midpoint, which hyperbolic attention already
uses and which, like the Lorentzian centroid, avoids the
iterative Fr\'{e}chet mean required for batch normalization.
In this work, we systematically explore its modeling
potential and build a Klein neural network framework of
closed-form operators, namely Klein MLR, Klein Busemann MLR,
Klein fully connected layer, Klein Busemann fully connected
layer, and Klein GyroBN. Klein demonstrates strong training stability under feature
clipping and curvature settings where the Poincar\'{e} and Lorentz reference
implementations diverge.
The closed-form Einstein midpoint in Klein GyroBN offers a favorable trade-off
between computational efficiency and performance against iterative Fr\'{e}chet means.
Across image classification and graph node
classification, the proposed Klein layers match their
Poincar\'{e}, Lorentz, and PV counterparts. Klein MLR heads further
achieve the fastest runtime among hyperbolic counterparts,
$1.13$--$1.20\times$ ahead of the next fastest head. A fully hyperbolic
Klein-ResNet trains $4.0\times$ faster than the reference HCNN
implementation on CIFAR-100 with a third of its GPU memory.
\end{abstract}

\section{Introduction}
\label{sec:introduction}

Hyperbolic geometry provides a natural representation for hierarchical and tree-like data due to its exponential volume growth, and has driven a growing body of \emph{hyperbolic neural networks} (HNNs) across applications, including computer vision~\citep{Khrulkov2020,Ermolov2022HVT,GhadimiAtigh2022,vanSpengler2023,Bdeir2024,Pal2025CEL,Sur2025HUFI,Liu2025HypCat,Bdeir2025Robust,He2025LResNet}, graph learning~\citep{Chami2019,Bachmann2020,Fu2024HypDiff,Sun2024GA,Li2024THGN}, natural language processing and knowledge graphs~\citep{Nickel2017,Ganea2018,Nickel2018,Gulcehre2019,Balazevic2019,Yang2025FT,Yang2024Hypformer,He2025HELM}, multimodal vision-language learning~\citep{Desai2023MERU}, recommendation systems~\citep{Yang2025HGformer}, astronomy~\citep{Chen2025Galaxy}, and genomic sequence learning~\citep{Zhou2021GeneExpr,Khan2025HGE}. These successes have driven work on hyperbolic space, principally on the Poincar\'{e} ball and the Lorentz hyperboloid with a recent extension to the Proper Velocity (PV) model~\citep{Chen2026PVNN}. The Poincar\'{e} ball and Lorentz hyperboloid have dominated recent work since they offer convenient Riemannian or gyrovector structures~\citep{Ganea2018,Chen2025gyrobn,Ungar2022} that facilitate neural network construction. However, the Poincar\'{e} ball and Lorentz hyperboloid are constrained spaces and can suffer from numerical instabilities. In particular, as Poincar\'{e}-ball embeddings approach the boundary, numerical computations become unstable and may cause vanishing gradients~\citep{Guo2022}.

The Klein ball is the third canonical hyperbolic model alongside the Poincar\'{e} ball and the Lorentz hyperboloid, originating from Einstein's special relativity~\citep{Ungar2022}. It admits an Einstein gyrovector space, paralleling the M\"{o}bius and proper-velocity gyrovector spaces of the Poincar\'{e} and PV models~\citep{Chen2026PVNN}. Among the three models, only the Klein ball admits Euclidean line segments as geodesics together with a closed-form Einstein midpoint~\citep[Ch.~6]{Ungar2022}. These properties make the Klein ball a natural geometry for batch-efficient hyperbolic neural networks, providing a closed-form alternative to the iterative Fr\'{e}chet means used for aggregation in Riemannian and gyrogroup batch normalization~\citep{Lou2020,Chen2025gyrobn}. This weighted average recurs throughout hyperbolic architectures as a batch center, few-shot prototype, message aggregator, and attention value mixture. Prior work already uses its Klein form in hyperbolic attention, while the Lorentzian centroid used in Lorentz batch normalization is the same average under an isometry~\citep{Gulcehre2019,Law2019,Bdeir2024}. Our native derivation removes coordinate conversion within Klein-HNN++, while the same closed-form average remains usable in Poincar\'{e} and Lorentz pipelines without adopting the Klein model. Beyond these uses~\citep{Gulcehre2019,Shimizu2021,Chen2025gyrobn}, recent work has begun to study the Klein model directly. \citet{Mao2024} develop Klein linear layers and a network built in the tangent space at the origin with a Euclidean classifier for graph node classification, whereas \citet{Zhang2026KleinHML} introduce an Einstein-gyrovector metric-learning method for fine-grained image classification. Neither study develops a unified, intrinsic layer framework with Klein-native geodesic and Busemann classifiers, fully connected counterparts, and batch normalization.

In this work, we propose Klein Neural Networks++ (Klein-HNN++), a systematic intrinsic Klein-based HNN framework, which stands to the Klein HNN of \citet{Mao2024} as HNN++~\citep{Shimizu2021} stands to HNN~\citep{Ganea2018}. We instantiate five intrinsic closed-form layers, Klein MLR, BMLR, FC, BFC, and GyroBN, via a Klein hyperplane distance, the Klein Busemann function, and an Einstein-midpoint acceleration of GyroBN, and assemble them with a Klein convolution and residual connection into a fully hyperbolic Klein-ResNet.
The contribution of Klein-HNN++ lies at the framework level. Most individual layers have counterparts in isometric hyperbolic models, but their coordinate expressions differ in computational cost and numerical behavior. In the Klein model, straight-line geodesics and Einstein gyrovector operations yield closed-form, batch-efficient implementations of the complete layer set, including a single-pass batch center (\cref{app:hyp-models}). Klein-HNN++ unifies these operators into an intrinsic toolkit whose computational efficiency and numerical operating range are evaluated below. Our main contributions are summarized as follows.
\begin{enumerate}[leftmargin=1.8em, itemsep=2pt, topsep=2pt]
\item We develop a new systematic toolkit of fundamental building blocks on the Klein ball, including MLR, Busemann MLR (BMLR), FC, Busemann FC (BFC), Klein GyroBN, and Klein-ResNet, which trains $4.0\times$ faster than the reference HCNN implementation on CIFAR-100 with a third of its GPU memory.
\item We show that the Klein layers train at every curvature from $|K|{=}0.01$ to $10$ and every clipping radius from $1$ to $10^{9}$, while the Poincar\'{e} reference diverges in six of these ten configurations and the Lorentz reference in two.
\item We derive a closed-form Klein batch normalization that bypasses the iterative Fr\'{e}chet means of prior gyrogroup BN via a single-pass Einstein midpoint, while matching the accuracy of converged Fr\'{e}chet means.
\item We validate the framework on numerical stability, image classification, link prediction, node classification, and ablations on the MLR head and the GyroBN gyromean, where Klein-HNN++ matches the accuracy of its Poincar\'{e}, Lorentz, and PV counterparts on image and node classification.
\end{enumerate}

\section{Preliminaries}
\label{sec:preliminary}

\textbf{Riemannian geometry~\citep{Lee2018}.}
Throughout, $\langle\cdot,\cdot\rangle$ denotes the standard Euclidean inner product and $\|\cdot\|$ the induced norm. A \emph{Riemannian manifold} $(\mathcal{M},g)$ is a smooth manifold equipped with a smoothly varying inner product $g_x$ on each tangent space $T_x\mathcal{M}$. We use $\Expop_x$ and $\Logop_x$ for the exponential map at $x$ and its local inverse, $\PT_{x\to y}$ for the parallel transport along the geodesic from $x$ to $y$, and $d(x,y)$ for the geodesic distance.

\textbf{Busemann function~\citep{Bridson1999}.}
Let $\gamma$ be a geodesic ray and let $d(\cdot,\cdot)$ denote the geodesic distance. The \emph{Busemann function} associated with $\gamma$ is defined as
\begin{equation}
\forall x\in\mathcal{M},\quad
B^{\gamma}(x) = \lim_{t\to\infty}\bigl(d(x,\gamma(t))-t\bigr),
\label{eq:busemann-def}
\end{equation}
and the limit exists on every hyperbolic manifold. In Euclidean space, the Busemann function associated with the geodesic $\gamma(t)=tv$ that starts at $\mathbf{0}$ with unit direction $v\in\Sn^{n-1}=\{v\in\RR^n\mid\|v\|=1\}$ is
\begin{equation}
\forall x\in\RR^n,\quad B^v(x) = -\langle x,v\rangle,
\label{eq:busemann-euclidean}
\end{equation}
which coincides with the inner product up to sign. Hence the Busemann function is the intrinsic generalization of the inner product to manifolds.

\textbf{Horosphere~\citep{Bridson1999}.}
The level sets of a Busemann function are called \emph{horospheres}, the hyperbolic counterpart of Euclidean hyperplanes. In Euclidean space, for a unit direction $v$, the hyperplanes $H^v_{\tau}=\{x\in\RR^n\mid\langle x,v\rangle=\tau\}$ with $\tau\in\RR$ are parallel. Analogously, for fixed $v$ and a hyperbolic manifold $\mathcal{M}$, the horospheres $H^v_{\tau}=\{x\in\mathcal{M}\mid B^v(x)=\tau\}$ are equidistant. When $\mathcal{M}$ is Euclidean, $B^v(x)=-\langle x,v\rangle$ recovers the linear hyperplane (with $\tau$ negated), so horospheres are the curved counterpart of linear half-space boundaries.

\begin{figure}[!ht]
\centering
\includegraphics[width=0.66\linewidth]{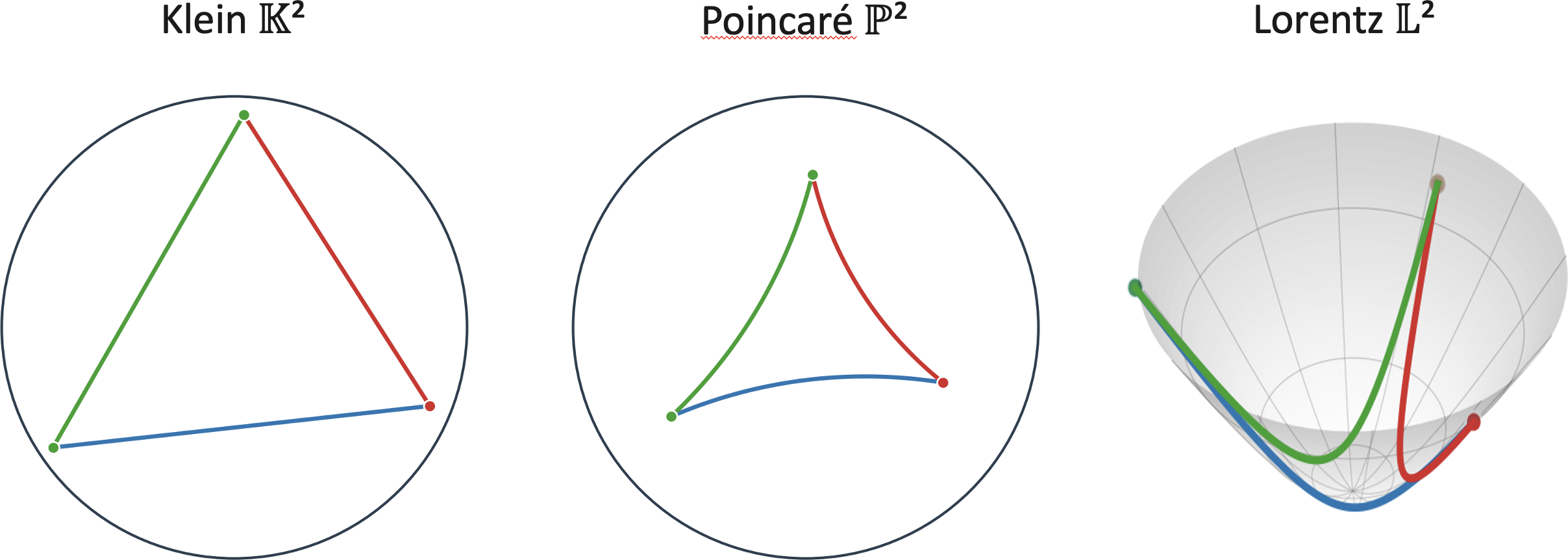}
\caption{Geodesic triangle on the three hyperbolic models ($K{=}{-}1$).}
\label{fig:three-geodesics}
\end{figure}

\textbf{Klein space~\citep{Ungar2022}.}
Hyperbolic space of constant curvature $K<0$ admits several equivalent models, the most popular of which are the Poincar\'{e} ball $\PP_K^n$ and the Lorentz hyperboloid $\LL_K^n$. The \emph{Klein model} $\KKn=\{x\in\RR^n\mid\|x\|^2<-1/K\}$~\citep[Ch.~3.8]{Ungar2022} originated in special relativity, parameterizing the relativistic velocity ball with $1/\sqrt{-K}$ as the speed of light. Its Riemannian metric is
\begin{equation}
g_x^{\KK}(u,v)
 = \frac{\langle u,v\rangle}{1+K\|x\|^2}
 - \frac{K\,\langle x,u\rangle\langle x,v\rangle}{(1+K\|x\|^2)^2},
 \qquad u,v\in T_x\KKn,
\label{eq:klein-metric}
\end{equation}
and we write $\gamma_x=1/\sqrt{1+K\|x\|^2}$ for the \emph{Lorentz factor}. Unlike the conformal Poincar\'{e} metric, $g_x^{\KK}$ has a non-isotropic component along $x$, but Klein geodesics are Euclidean line segments inside the open ball~\citep{Ungar2022}, in contrast to the circular arcs of $\PP_K^n$ and the hyperbolic sections of $\LL_K^n$, see \cref{fig:three-geodesics}. The Riemannian isometries connecting the three models appear in \cref{app:isometries}.

\textbf{Klein gyrovector~\citep{Ungar2022}.}
From an algebraic point of view, the Klein space forms a gyrovector space, which extends the Euclidean vector space to manifolds. Given $x,y\in\KKn$ and $t\in\RR$, Einstein gyroaddition $\oE$ and scalar gyromultiplication $\otE$ are defined as
\begin{align}
x\oE y &= \frac{1}{1-K\langle x,y\rangle}\!\left(
   x + \tfrac{1}{\gamma_x}\,y
  - \tfrac{K\gamma_x}{1+\gamma_x}\langle x,y\rangle\,x\right),
\label{eq:einstein-add}\\[2pt]
t\otE x &= \tfrac{1}{\sqrt{-K}}\,
  \tanh\!\bigl(t\arctanh(\sqrt{-K}\|x\|)\bigr)\tfrac{x}{\|x\|},
\quad (x\neq\mathbf{0}),\;\; t\otE\mathbf{0}=\mathbf{0}.
\label{eq:einstein-scalar}
\end{align}
In particular, the Einstein inverse is $\ominus_{\mathrm{E}}x=-x$, and the Einstein identity is the zero vector: $\mathbf{0}\oE x=x\oE\mathbf{0}=x$.

A summary of the manifold, metric, distance, exponential and logarithmic maps, and parallel transport on all three hyperbolic models, and detailed reviews of Riemannian geometry, the Busemann function and horospheres, gyrovector spaces, and the Poincar\'{e} ball and Lorentz hyperboloid models, are deferred to \cref{app:prelim}, with a side-by-side summary in \cref{tab:riem-ops-3model}.

\section{Klein Neural Networks}
\label{sec:klein-nn}

This section instantiates the core layers of a Klein neural network, namely Multinomial Logistic Regression (MLR), Busemann MLR (BMLR), Fully Connected (FC) and Busemann FC (BFC) maps, and batch normalization, and assembles them into convolutional residual networks (\cref{sec:klein-resnet}). We refer to this layer set as \emph{Klein Neural Networks++} (Klein-HNN++), following the progression from HNN to HNN++~\citep{Shimizu2021}. As defined by \citet{Chen2026,Shi2026}, a layer is intrinsic when its output is computed from quantities of the manifold rather than from a Euclidean operation in a tangent or ambient space (\cref{app:intrinsic}). Unlike the network of \citet{Mao2024}, built in the tangent space at the origin, Klein-HNN++ therefore provides a set of layers defined intrinsically on $\KKn$ for constructing full neural architectures.

\subsection{Klein Multinomial Logistic Regression}
\label{sec:klein-mlr}

As shown by~\citet{Lebanon2004,Ganea2018,Chen2024RMLR}, the Euclidean MLR $\mathrm{Softmax}(Ax+b)$ can be recast as a signed-margin layer
\begin{equation}
  p(y=k\mid x)\propto\exp(v_k(x)), \quad
  v_k(x) = \sign(\langle a_k,x-p_k\rangle)\,\|a_k\|\,d\!\left(x,H_{a_k,p_k}\right),
\label{eq:euc-mlr}
\end{equation}
with $H_{a_k,p_k}=\{x\in\RR^n\mid\langle a_k,x-p_k\rangle=0\}$. Following the Poincar\'{e} MLR \citep{Ganea2018}, we define the \emph{Klein hyperplane} parameterized by a base point $p\in\KKn$ and a normal direction $a\in T_p\KKn$ as the level set $\{x\in\KKn\mid\langle\Logop_p^{\KK}(x),a\rangle_p=0\}$. The following theorem gives an equivalent gyrovector form together with the point-to-hyperplane distance.

\begin{theorem}[Klein hyperplane and distance]\proofref{app:proof-hyp}
\label{thm:klein-hyp}
For $x,p\in\KKn$ and $a\in T_p\KKn$, the Klein hyperplane
admits the equivalence
\begin{equation}
H_{a,p}
 = \{x\in\KKn\mid\langle\Logop_p^{\KK}(x),a\rangle_p=0\}
 = \{x\in\KKn\mid\langle(-p)\oE x,w\rangle=0\},
\label{eq:klein-hyp}
\end{equation}
and the point-to-hyperplane distance is
\begin{equation}
d^{\KK}(y,H_{a,p})
 = \inf_{q\in H_{a,p}} d^{\KK}(y,q)
 = \frac{1}{\sqrt{-K}}\arcsinh\!\left(
    \frac{\sqrt{-K}\,\gamma_z\,|\langle z,w\rangle|}{\|w\|}\right),
\label{eq:klein-dist}
\end{equation}
where $z=(-p)\oE y$, $w=\PT_{p\to\mathbf{0}}^{\KK}(a)$, and
$\gamma_z=1/\sqrt{1+K\|z\|^2}$.
\end{theorem}

By \cref{thm:klein-hyp}, we define the $C$-class Klein MLR as
\begin{equation}
p(y{=}k\mid x)\propto\exp(v_k(x)), \quad
v_k(x) = \sign\!\bigl(\langle(-p_k)\oE x,w_k\rangle\bigr)\,
         \|a_k\|_{p_k}\,d^{\KK}\!\bigl(x,H_{a_k,p_k}\bigr),
\label{eq:klein-mlr-raw}
\end{equation}
parameterized by $p_k\in\KKn$ and $a_k\in T_{p_k}\KKn$ for each class $k$. This raw form has three drawbacks. It is over-parameterized, since the bias is absorbed into $p_k$. It couples the gyroaddition $(-p_k)\oE x$ across classes. It also constrains $p_k$ to $\KKn$, requiring Riemannian optimization. Following \citet{Shimizu2021}, we adopt the unidirectional parameterization $p_k=\Expop_{\mathbf{0}}^{\KK}(r_k\hat z_k)$, $a_k=\PT_{\mathbf{0}\to p_k}^{\KK}(z_k)$ with $z_k\in\RR^n$, $r_k\in\RR$, $\hat z_k=z_k/\|z_k\|$, which puts the margin of \cref{eq:klein-mlr-raw} in closed form.

\begin{theorem}[Klein MLR]\proofref{app:proof-mlr}
\label{thm:klein-mlr}
For $x\in\KKn$, the Klein MLR score for class $k$ is
\begin{equation}
v_k(x) = \frac{\|z_k\|}{\sqrt{-K}}\arcsinh\!\left(
  \gamma_x\!\left(\cosh(\sqrt{-K}\,r_k)\frac{\sqrt{-K}}{\|z_k\|}\langle x,z_k\rangle
  -\sinh(\sqrt{-K}\,r_k)\right)\right),
\label{eq:klein-mlr}
\end{equation}
with $\gamma_x=1/\sqrt{1+K\|x\|^2}$. As $K\to 0^-$,
$v_k(x)\to\langle x,z_k\rangle-r_k\|z_k\|$, the
Euclidean MLR of \cref{eq:euc-mlr}.
\end{theorem}

\cref{thm:klein-mlr} provides a closed-form geodesic Klein MLR via the signed point-to-Klein-hyperplane distance of \cref{eq:klein-dist}. Unlike the tangent-space head of~\citet{Mao2024}, it classifies with the Klein distance itself. The unidirectional parameterization $(z_k,r_k)$ is essential for batchability: \cref{eq:klein-mlr} depends on $x$ only through the inner product $\langle x,z_k\rangle$ and the scalar $\|x\|^2$ (via $\gamma_x$), implementable as a single matmul $XZ$, in contrast to the per-class $-p_k\oE x$ tensor that the raw \cref{eq:klein-mlr-raw} would require. This reparameterization preserves the full family of Klein geodesic hyperplanes. Indeed, setting $v_k(x)=0$ in \cref{eq:klein-mlr} gives $\langle x,\hat z_k\rangle=\tanh(\sqrt{-K}r_k)/\sqrt{-K}$, where $\hat z_k$ determines the hyperplane orientation and $r_k$ its signed offset. Hence every Klein geodesic hyperplane, including oblique ones, is represented by some $(\hat z_k,r_k)$. \cref{app:param-equiv} gives the exact conversion from the raw $(p_k,a_k)$ form and compares the two parameterizations empirically. The full derivation is in \cref{app:proof-mlr}.

\subsection{Klein Busemann MLR}
\label{sec:klein-bmlr}

The Euclidean signed margin $v_k(x)=\sign(\langle a_k,x{-}p_k\rangle)\,\|a_k\|\,d(x,H_{a_k,p_k})$ of \cref{eq:euc-mlr} is a point-to-hyperplane distance. \citet{Chen2026} equivalently write this as a point-to-horosphere distance through the Busemann function, see \cref{eq:busemann-def}, since horospheres recover Euclidean hyperplanes in the flat limit. Following the horosphere-based MLR of \citet{Chen2026}, we replace geodesic hyperplanes by horospheres as decision boundaries and define the $C$-class Klein Busemann MLR
\begin{equation}
p(y{=}k\mid x)\propto\exp(u_k(x)), \quad
u_k(x) = -\alpha_k B^{v_k}_{\KK}(x) + b_k,
\label{eq:klein-bmlr}
\end{equation}
with learnable $\alpha_k>0$, $v_k\in\Sn^{n-1}$, $b_k\in\RR$. Evaluating \cref{eq:klein-bmlr} requires the Klein Busemann function $B^v_{\KK}$ in closed form, which the following theorem provides.

\begin{theorem}[Klein Busemann function]\proofref{app:busemann-verify}
\label{thm:klein-busemann}
Let $v\in\Sn^{n-1}$ be a unit direction at the origin and let
$\gamma(t)=\Expop_{\mathbf{0}}^{\KK}(tv)$ be the unit-speed
Klein geodesic ray it generates. The Busemann function
$B^v_{\KK}:\KKn\to\RR$ associated with $\gamma$ admits the
closed form
\begin{equation}
B^v_{\KK}(x) = \frac{1}{\sqrt{-K}} \log\!\left(
  \frac{1-\sqrt{-K}\langle x,v\rangle}{\sqrt{1+K\|x\|^2}}\right),
  \qquad x\in\KKn.
\label{eq:klein-busemann}
\end{equation}
Its level sets are Klein horospheres (ellipsoids tangent to
the boundary sphere $\{\|y\|=1/\sqrt{-K}\}$ at $v/\sqrt{-K}$),
and $B^v_{\KK}(x)\to-\langle v,x\rangle$ as $K\to 0^-$.
\end{theorem}

The closed form \cref{eq:klein-busemann} is derived via the Klein--Lorentz isometry and independently cross-checked through the Klein--Poincar\'{e} isometry, with both derivations given in \cref{app:busemann-verify}. The dominant operation is $\sqrt{-K}\langle x,v\rangle$, which reduces to a single scaled matmul $\sqrt{-K}XV$ over a batch.

Substituting \cref{thm:klein-busemann} into \cref{eq:klein-bmlr} yields the explicit Klein BMLR logit
\begin{equation}
u_k(x) = \frac{\alpha_k}{\sqrt{-K}}\left[
  \tfrac{1}{2}\log(1+K\|x\|^2)
  -\log(1-\sqrt{-K}\langle x,v_k\rangle)\right] + b_k.
\label{eq:klein-bmlr-expanded}
\end{equation}
A first-order expansion as $K\to 0^-$ gives $u_k(x)\to\alpha_k\langle v_k,x\rangle+b_k$, recovering the Euclidean MLR exactly. See \cref{app:proof-bmlr}. Klein BMLR complements the geodesic MLR of \cref{sec:klein-mlr} by providing a horosphere-based formulation, and we evaluate both variants. The level sets of a Busemann function form an equidistant family, $d(H^{v}_{\tau_1},H^{v}_{\tau_2})=|\tau_2-\tau_1|$. Consequently, $|u_k(x)|/\alpha_k$ equals the geodesic distance from the level set through $x$ to the decision horosphere. This interpretation holds throughout the score range, whereas distinct geodesic hyperplanes do not form an equidistant family. \cref{app:busemann} summarizes the Euclidean-to-hyperbolic correspondence underlying this construction (\cref{fig:horosphere-vs-chord}).

\textbf{Discussion.}
\cref{tab:mlr-comparison} compares the logit form, parameter count, FLOP cost, and batch efficiency across hyperbolic MLR methods. Klein BMLR attains the lowest per-class FLOP count $C(2n{+}11)$ among methods with unconstrained parameters, and the Klein heads are the fastest in \cref{tab:mlr-runtime}, by $1.13$ to $1.20$ times. Per-method derivations and the FLOP-counting convention are in \cref{app:flop-convention}.

The gap follows from the model rather than the implementation. Klein geodesics are Euclidean line segments, so the pair $(z_k,r_k)\in\RR^n\times\RR$ specifies a decision boundary completely, whereas the same surface on $\PP_K^n$ is a sphere meeting the ideal boundary orthogonally and must be anchored at a manifold-valued point, retracted after every update (\cref{fig:boundary-models}, \cref{app:boundary-geometry}).

\begin{table}[!t]
\centering
\caption{Hyperbolic MLR methods. Dist is the geodesic point-to-hyperplane distance (\distreal{}) or a Euclidean surrogate (\distpseudo{}); compact params avoid extra manifold-valued parameters (\cref{app:mlr-column-defs}).}
\label{tab:mlr-comparison}
\renewcommand{\arraystretch}{1.25}
\setlength{\tabcolsep}{4pt}
\adjustbox{max width=\textwidth}{%
\begin{tabular}{>{\raggedright\arraybackslash}m{2.7cm} >{\centering\arraybackslash}m{5.7cm} c c c c c c}
\toprule
\textbf{Method} &
\makecell{\textbf{Logit} $u_k(x)$,\\$\forall k\!\in\!\{1,\ldots,C\}$} &
\textbf{Space} & \textbf{Dist} & \textbf{\#Params} &
\makecell{\textbf{Compact}\\\textbf{params}} &
\textbf{FLOPs} &
\makecell{\textbf{Batch}\\\textbf{eff.}} \\
\midrule
Euclidean MLR &
\makecell{$\langle a_k, x\rangle + b_k$\\[1pt]
{\scriptsize $a_k\!\in\!\RR^n,\, b_k\!\in\!\RR$}} &
$\RR^n$ & \distreal{} & $C(n{+}1)$ & \cmark & $C(2n)$ & \cmark \\
\midrule
Poincar\'{e} MLR~\citep{Ganea2018} &
\makecell{$\frac{\lambda_{p_k}^K\|a_k\|}{\sqrt{-K}}\,\arcsinh(\alpha)$\\[1pt]
$\alpha\!=\!\frac{2\sqrt{-K}\,\langle {-}p_k\!\oplus_{\!M}\! x,\, a_k\rangle}
  {(1{+}K\|{-}p_k\!\oplus_{\!M}\! x\|^2)\|a_k\|}$\\[1pt]
{\scriptsize $p_k\!\in\!\PP_K^n,\; a_k\!\in\! T_{p_k}\!\PP_K^n$}} &
$\PP_K^n$ & \distreal{} & $C(2n)$ & \xmark &
$C(19n{+}27)$ & \xmark \\
\midrule
Poincar\'{e} MLR~\citep{Shimizu2021} &
\makecell{$\frac{2\alpha_k}{\sqrt{-K}}\,\arcsinh\bigl(\alpha{-}\beta\bigr)$\\[1pt]
{\scriptsize $\alpha\!=\!\lambda_x^K\!\sqrt{-K}\langle x,v_k\rangle
  \cosh(2\sqrt{-K}b_k)$}\\[1pt]
{\scriptsize $\beta\!=\!(\lambda_x^K{-}1)\sinh(2\sqrt{-K}b_k)$}\\[1pt]
{\scriptsize $\alpha_k\!>\!0,\; v_k\!\in\!\Sn^{n-1},\; b_k\!\in\!\RR$}} &
$\PP_K^n$ & \distreal{} & $C(n{+}2)$ & \cmark & $C(2n{+}46)$ & \cmark \\
\midrule
Pseudo-Busemann MLR~\citep{Nguyen2025} &
\makecell{$-d(x,p_k)\,\frac{B^{v_k}({-}p_k\!\oplus_{\!M}\! x)}
  {\|{-}p_k\!\oplus_{\!M}\! x\|}$\\[1pt]
{\scriptsize $p_k\!\in\!\PP_K^n,\; v_k\!\in\!\Sn^{n-1}$}} &
$\PP_K^n$ & \distpseudo{} & $C(2n)$ & \xmark &
$C(18n{+}30)$ & \xmark \\
\midrule
Lorentz MLR~\citep{Bdeir2024} &
\makecell{$\frac{1}{\sqrt{-K}}\sign(\alpha)\,\beta
  \bigl|\arcsinh\!\bigl(\sqrt{-K}\frac{\alpha}{\beta}\bigr)\bigr|$\\[1pt]
{\scriptsize $z_k\!\in\!\RR^n,\; b_k\!\in\!\RR$}} &
$\LL_K^n$ & \distreal{} & $C(n{+}1)$ & \cmark & $C(4n{+}51)$ & \cmark \\
\midrule
BMLR~\citep{Chen2026} &
\makecell{$-\alpha_k B^{v_k}(x) + b_k$\\[1pt]
{\scriptsize $\alpha_k\!>\!0,\; v_k\!\in\!\Sn^{n-1},\; b_k\!\in\!\RR$}\\[1pt]
{\scriptsize \emph{(P, $(p_k,v_k)$ / $v_k$-unit / L)}}} &
\makecell{$\PP_K^n$\\$\PP_K^n$\\$\LL_K^n$} & \distreal{} & $C(n{+}2)$ & \cmark &
\makecell{$C(4n{+}10)$\\$C(2n{+}12)$\\$C(2n{+}12)$} & \cmark \\
\midrule
\rowcolor{blue!8}
\textbf{Klein MLR} \cref{sec:klein-mlr} &
{\makecell{$\frac{\|z_k\|}{\sqrt{-K}}\arcsinh(\gamma_x\alpha)$\\[1pt]
{\scriptsize $\alpha\!=\!\cosh(\sqrt{-K}r_k)\,\sqrt{-K}\,\|z_k\|^{-1}\langle x,z_k\rangle
  {-}\sinh(\sqrt{-K}r_k)$}\\[1pt]
{\scriptsize $z_k\!\in\!\RR^n,\; r_k\!\in\!\RR$}}} &
$\KKn$ & \distreal{}
& $C(n{+}1)$ & \cmark
& $C(2n{+}14)$ & \cmark \\
\rowcolor{blue!8}
\textbf{Klein BMLR} \cref{sec:klein-bmlr} &
{\makecell{$-\frac{\alpha_k}{\sqrt{-K}}\log\!\left(
  \frac{1{-}\sqrt{-K}\langle x,v_k\rangle}{\sqrt{1{+}K\|x\|^2}}\right)+b_k$\\[1pt]
{\scriptsize $\alpha_k\!>\!0,\; v_k\!\in\!\Sn^{n-1},\; b_k\!\in\!\RR$}}} &
$\KKn$ & \distreal{}
& $C(n{+}2)$ & \cmark
& $C(2n{+}11)$ & \cmark \\
\bottomrule
\end{tabular}%
}
\end{table}

\subsection{Klein Fully Connected and Busemann Fully Connected Layers}
\label{sec:klein-fc}

\citet{Shimizu2021} recast the Euclidean FC layer $y=Ax+b$ as $m$ signed-distance equations $\sign(\langle e_k,y\rangle)\,d(y,H_{e_k,\mathbf{0}}) = \langle a_k,x-p_k\rangle$ for $1\le k\le m$, with $e_k$ the $k$-th standard basis vector. \citet{Chen2026} rewrite the inner product as the Busemann score $-\alpha_k B^{v_k}(x)+b_k$, replacing the geodesic signed-distance with a horosphere-based one. We unify both via the Klein analogue
\begin{equation}
\sign(y_k)\,d^{\KK}(y,H_{e_k,\mathbf{0}}) = s_k(x), \quad 1\le k\le m,
\label{eq:klein-fc-def}
\end{equation}
which defines the Klein FC $\mathcal{F}:\KKn\to\KK_K^m$, where $s_k$ is the Klein FC logit $v_k$ from \cref{eq:klein-mlr} or the Klein BFC logit $u_k$ from \cref{eq:klein-bmlr-expanded}, and admits a unified closed form.

\begin{theorem}[Klein FC and BFC]\proofref{app:proof-fc}
\label{thm:klein-fc}
For $x\in\KKn$, the output $y=\mathcal{F}(x)\in\KK_K^m$ of
\cref{eq:klein-fc-def} is
\begin{equation}
y_k = \frac{\sinh(\sqrt{-K}\,s_k(x))}
           {\sqrt{-K}\;\sqrt{1+\sum_{j=1}^{m}\sinh^2(\sqrt{-K}\,s_j(x))}},
\quad 1\le k\le m,
\label{eq:klein-fc}
\end{equation}
with
\begin{equation}
s_k(x) =
\begin{cases}
v_k(x), & \text{Klein FC},\\
u_k(x), & \text{Klein BFC}.
\end{cases}
\label{eq:klein-fc-cases}
\end{equation}
Klein FC and Klein BFC use $m(n{+}1)$ and $m(n{+}2)$ parameters, respectively, and both recover the Euclidean
FC as $K\to 0^-$. The proof, together with a comparison of FC
vs.\ BFC costs, is in \cref{app:proof-fc}.
\end{theorem}

The shared denominator is the Lorentz time component required to project the output back to the Klein ball. Because it rescales all coordinates equally, it preserves every pairwise ratio $y_i/y_j$. Further details are provided in \cref{app:fc-denominator}.

\textbf{Generalization with activation.}
The Euclidean FC with activation $y=\sigma(Ax+b)$ applies $\sigma$ to the linear preactivation $\langle a_k,x\rangle+b_k$, which under the recast above equivalently appears as $\sigma(s_k(x))$ on the RHS of \cref{eq:klein-fc-def}. We extend the Klein FC/BFC by the same substitution, and \cref{eq:klein-fc} becomes
\begin{equation}
y_k = \frac{\sinh\!\bigl(\sqrt{-K}\,\sigma(s_k(x))\bigr)}
           {\sqrt{-K}\;\sqrt{1+\sum_{j=1}^{m}\sinh^2\!\bigl(\sqrt{-K}\,\sigma(s_j(x))\bigr)}},
\quad 1\le k\le m.
\label{eq:klein-fc-act}
\end{equation}

\textbf{Discussion.}
Each per-output logit reads directly off \cref{tab:mlr-comparison}, and the sinh-normalized Klein output map adds a shared $18m{+}2$ FLOPs. Klein BFC costs $2nm{+}29m{+}2n{+}4$ FLOPs and Lorentz BFC-L $2nm{+}30m{+}1$, so Klein BFC saves one FLOP per output but spends $2n{+}2$ on $\gamma_x$, and is cheaper exactly when $m>2n+3$. \cref{tab:fc-comparison} in \cref{app:fc-bfc-cost} lays the full comparison out side by side, extending \citet[Tab.~3]{Chen2026} with our Klein FC and Klein BFC.

\subsection{Klein Batch Normalization}
\label{sec:klein-bn}

Euclidean Batch Normalization (BN) centers a batch using its mean, normalizes its dispersion using the batch variance, and applies learnable scale and bias parameters. GyroBN~\citep{Chen2025gyrobn} lifts these operations to manifolds through their gyrogroup counterparts. We instantiate this template on the Klein ball, first with Fr\'{e}chet statistics and then with a closed-form Einstein midpoint.

For a batch $\{x_i\}_{i=1}^m\subset\KKn$, the corresponding Fr\'{e}chet mean and variance are
\begin{equation}
\mu_{\mathrm{F}} = \FM\bigl(\{x_i\}_{i=1}^m\bigr)
 = \argmin_{y\in\KKn}\frac{1}{m}\sum_{i=1}^{m} d^{\KK}(x_i,y)^2,
\qquad
v^2 = \frac{1}{m}\sum_{i=1}^{m} d^{\KK}(x_i,\mu_{\mathrm{F}})^2.
\label{eq:klein-frechet-stats}
\end{equation}
Substituting Einstein addition and scalar multiplication into the GyroBN template yields the per-sample update
\begin{equation}
\forall i\le m,\quad
\tilde{x}_i \leftarrow
\overbrace{\beta\,\oE}^{\text{Biasing}}
\!\!\left[\,
  \overbrace{\tfrac{\gamma}{\sqrt{v^2}+\varepsilon}\otE}^{\text{Scaling}}
  \!\!\left(\,
    \overbrace{(-\mu_{\mathrm{F}})\oE x_i}^{\text{Centering}}
  \right)
\right],
\label{eq:klein-gyrobn}
\end{equation}
with learnable $\beta\in\KKn$ and $\gamma\in\RR_+$, recovering Euclidean BN as $K\to 0^-$. As shown by~\citet{Chen2025gyrobn}, Klein GyroBN has the following homogeneity properties:
\begin{align}
\mu\!\bigl(\{\beta\oE x_i\}\bigr)
 &= \beta\oE\mu\!\bigl(\{x_i\}\bigr),
 \label{eq:bn-homog-mean}\\
\tfrac{1}{m}\sum_{i=1}^m d^{\KK}(t\otE x_i,\mathbf{0})^2
 &= t^2\cdot\tfrac{1}{m}\sum_{i=1}^m d^{\KK}(x_i,\mathbf{0})^2,
 \label{eq:bn-homog-disp}
\end{align}
for all $\beta\in\KKn$ and $t\in\RR$, so that \cref{eq:bn-homog-mean} guarantees centering and biasing exactly shift the gyromean while \cref{eq:bn-homog-disp} ensures the scaling step exactly controls the dispersion, reproducing the Euclidean BN normalization guarantee under the Klein geodesic metric. The full inference algorithm, which adapts the GyroBN~\citep{Chen2025gyrobn} to the Klein ball, is in \cref{alg:klein-gyrobn} of \cref{app:klein-gyrobn-algo}.

\textbf{Closed-form acceleration via the Einstein midpoint.}
The Fr\'{e}chet mean $\mu_{\mathrm{F}}$ has no closed form and requires an iterative solver~\citep{Lou2020}, slowing training~\citep{Bdeir2024}. We substitute $\mu_{\mathrm{F}}$ in \cref{eq:klein-gyrobn} by the closed-form \emph{Einstein midpoint}~\citep{Ungar2022}
\begin{equation}
\mu_{\mathrm{E}} = \frac{\sum_{i=1}^{m}\gamma_{x_i}\,x_i}
                       {\sum_{i=1}^{m}\gamma_{x_i}},
\qquad
\gamma_{x_i} = \frac{1}{\sqrt{1+K\|x_i\|^2}},
\label{eq:einstein-midpoint}
\end{equation}
which lies in $\KKn$ at $O(mn)$ cost and matches the iterative Fr\'{e}chet mean within seed noise on node classification. \citet{Gulcehre2019} already use this midpoint as a closed-form aggregation in hyperbolic attention, and the Lorentzian centroid of Lorentz batch normalization~\citep{Bdeir2024} is the same point under the isometry to the Lorentz model. The Lorentz weights increase with radius, giving boundary points greater influence on the Einstein midpoint. We use the midpoint for single-pass batch centering and apply feature clipping to control this sensitivity. See \cref{sec:exp-bn-ablation} for empirical results and \cref{app:einstein-frechet} for the full analysis and engineering considerations.

\subsection{Klein Convolution and Residual Networks}
\label{sec:klein-resnet}

An operation $f$ of a Euclidean residual network can be moved to $\KKn$ in two ways. The first maps the points to $\RR^n$ with a bijection $\varphi$, applies $f$ there and maps the result back, $\varphi^{-1}\circ f\circ\varphi$. We call it the tangent version when $\varphi=\LogO$, the inverse of $\ExpO(u)=\tanh(\sk\|u\|)\,u/(\sk\|u\|)$, as in the M\"{o}bius version of \citet[Def.~3.2]{Ganea2018} and the Einstein version of \citet[Def.~8]{Mao2024}, and the spatial version when $\varphi(x)=\sigma(x)=\gamma_x x$ is the spatial part of the matching Lorentz point (\cref{lem:klein-lorentz}), as in the space addition and the Lorentz ReLU of HCNN~\citep{Bdeir2024}. The second builds the operation from the Klein geometry itself, as for the Klein FC, the pointwise ReLU, $\oE$ and $\otE$. \cref{app:ops-ablation} compares the options for each operation in Klein-ResNet.

\textbf{Addition.}
A residual connection adds the output of a block to its input, but the sum of two points of $\KKn$ can leave the ball. The most direct choice is the Einstein addition $\oE$, the Klein counterpart of the M\"{o}bius addition used in the Poincar\'{e} ResNet~\citep{vanSpengler2023}. The other two options are the tangent and spatial versions of ordinary vector addition, $\ExpO(\LogO x+\LogO y)$ and
\begin{equation}
x\oS y=\sigma^{-1}\bigl(\sigma(x)+\sigma(y)\bigr),\qquad
r\otS x=\sigma^{-1}\bigl(r\,\sigma(x)\bigr),\qquad
\sigma^{-1}(v)=\frac{v}{\sqrt{1-K\|v\|^2}},
\label{eq:klein-res}
\end{equation}
where the spatial version is the space addition of HCNN~\citep[\S4.4]{Bdeir2024}. Klein-ResNet uses $\oS$ instead of $\oE$, since it is at least as accurate as the other two and slightly faster (\cref{app:proof-res,app:ops-ablation}).

\textbf{Activation.}
Each block needs a nonlinearity that maps $\KKn$ into itself. Besides the tangent version $\ExpO\circ\ReLU\circ\LogO$ and the spatial version $\sigma^{-1}\circ\ReLU\circ\sigma$, which is the Lorentz ReLU of HCNN~\citep[Eq.~(13)]{Bdeir2024}, ReLU can act pointwise on the Klein coordinates, which keeps every point in the ball since $\|\ReLU(x)\|\le\|x\|$. Klein-ResNet uses this pointwise ReLU, which of the three keeps points closest to the origin, away from the boundary where $\gamma_x$ diverges, and needs no $\exp$, $\log$ or $\tanh$ (\cref{prop:relu}).

\textbf{Convolution.}
A convolution applies one linear map to every $k\times k$ receptive field $\mathcal{N}(q)$. The tangent and spatial versions transport the Euclidean convolution as a whole, for example $\ExpO\bigl(\sum_{i\in\mathcal{N}(q)}W_i\LogO(x_i)+b\bigr)$ (\cref{eq:klein-conv-tangent}). Following HNN++~\citep{Shimizu2021} and HCNN~\citep{Bdeir2024}, Klein-ResNet instead concatenates the receptive field of each location $q$ in the tangent version and applies the Klein FC $\mathcal{F}$ of \cref{thm:klein-fc}, so that the linear part of the convolution is hyperbolic, with each output set by signed distances to Klein hyperplanes (\cref{eq:klein-fc-def}) rather than by a Euclidean affine map in a chart. For inputs $x_i\in\KKd{C}$ it outputs
\begin{equation}
y_q=\mathcal{F}\Bigl(\ExpO\bigl(\Concat_{i\in\mathcal{N}(q)}\LogO(x_i)\bigr)\Bigr)\in\KKd{C'}.
\label{eq:klein-conv}
\end{equation}
Since $\mathcal{F}$ depends on the concatenated vector only through its norm and its inner products with the $C'$ hyperplane normals, which are standard convolutions of $\LogO(x)$, \cref{eq:klein-conv} is computed without forming patches (\cref{prop:fc-conv}). The tangent and spatial versions of the whole convolution are faster at comparable accuracy (\cref{app:ops-ablation}).

\textbf{Klein-ResNet.}
Klein-ResNet replaces every component of the CIFAR ResNet-18 of HCNN~\citep{Bdeir2024} with the Klein layers above and maps each pixel into the ball by $X_q=\sigma^{-1}(\nu x_q)$ with an input scale $\nu>0$ (\cref{tab:resnet-hparams}). At equal curvature and $\nu=1$, the two networks differ under the Lorentz--Klein isometry only in the convolution, the radial part of the activation and, for the reference implementation of HCNN, the spread measure used by batch normalization (\cref{prop:klein-hcnn}). See \cref{app:resnet-exp} for the block structure.

\vspace{-0.5em}
\section{Experiments}
\label{sec:experiments}
\vspace{-0.5em}

We evaluate the Klein layers against Poincar\'{e}, Lorentz, and PV baselines (\cref{app:experiments}). \Crefrange{sec:exp-stability}{sec:exp-bn-ablation} replace single layers in fixed backbones, and \cref{sec:exp-resnet} builds a fully hyperbolic ResNet.

\vspace{-0.6em}
\needspace{12\baselineskip}
\subsection{Numerical Stability}
\label{sec:exp-stability}
\vspace{-0.4em}

\setlength{\intextsep}{6pt}   
\setlength{\columnsep}{8pt}    

\begin{wraptable}{r}{0.44\textwidth}   
\vspace*{-1\baselineskip}                
\centering
\begingroup%
\caption{Curvature sweep (mean$\pm$std, 3 seeds).}
\label{tab:stab-curv}
\setlength{\tabcolsep}{3pt}            
\renewcommand{\arraystretch}{1.0}     
\setlength{\aboverulesep}{1pt}\setlength{\belowrulesep}{1pt}
\footnotesize                                 
\adjustbox{max width=\linewidth}{%
\begin{tabular}{@{}c|>{\columncolor{blue!8}}ccc@{}}
\toprule
$|K|$ & \textbf{Klein} & \textbf{Poincar\'{e}} & \textbf{Lorentz} \\
\midrule
$0.01$ & $72.29\pm0.21$ & \textcolor{red}{\textbf{NaN}} & $72.37\pm0.14$ \\
$0.1$  & $72.90\pm0.15$ & $73.91\pm0.12$ & $72.58\pm0.29$ \\
$0.3$  & $73.32\pm0.19$ & $73.54\pm0.21$ & $73.10\pm0.20$ \\
$1.0$  & $73.11\pm0.15$ & $73.17\pm0.24$ & $73.17\pm0.12$ \\
$10.0$ & $70.88\pm0.16$ & \textcolor{red}{\textbf{NaN}} & \textcolor{red}{\textbf{NaN}} \\
\bottomrule
\end{tabular}}
\endgroup
\end{wraptable}
Across the two sweeps below, the Poincar\'{e} reference diverges in six of ten configurations and the Lorentz reference in two; however, Klein remains stable throughout. We use two ResNet-18 sweeps and locate each failure to one floating-point step. The sweeps use CIFAR-100 for $100$ epochs over three seeds. Each manifold is deployed through its standard reference implementation. Further diagnostics are in \cref{app:numerical-extra} and \cref{app:cka}.

\begin{wraptable}{r}{0.44\textwidth}   %
\vspace*{-\baselineskip}                %
\centering
\begingroup%
\caption{Clipping sweep at $K{=}{-}0.3$ by geodesic radius $\rho$ (3 seeds).}
\label{tab:stab-clip}
\setlength{\tabcolsep}{3pt}            %
\renewcommand{\arraystretch}{1.0}      %
\setlength{\aboverulesep}{1pt}\setlength{\belowrulesep}{1pt}
\footnotesize                                  %
\adjustbox{max width=\linewidth}{%
\begin{tabular}{@{}c|>{\columncolor{blue!8}}ccc@{}}
\toprule
$\rho$ & \textbf{Klein} & \textbf{Poincar\'{e}} & \textbf{Lorentz} \\
\midrule
$2$      & $74.54\pm0.13$ & $73.47\pm0.19$ & $74.68\pm0.56$ \\
$10$     & $73.51\pm0.12$ & \textcolor{red}{\textbf{NaN}} & $72.44\pm0.21$ \\
$\infty$ & $71.69\pm0.15$ & \textcolor{red}{\textbf{NaN}} & \textcolor{red}{\textbf{NaN}} \\
\bottomrule
\end{tabular}}
\endgroup
\end{wraptable}
\textbf{Curvature magnitude.}~We sweep $|K|\in\{0.01,0.1,0.3,1.0,10.0\}$ at fixed $r{=}1.0$, spanning three orders of magnitude. Klein remains finite across all five curvatures with low variation across seeds. Poincar\'{e} diverges at both ends of the range, $|K|{=}0.01$ and $|K|{=}10$, and Lorentz diverges at $|K|{=}10$ after about $30$ epochs.

\textbf{Feature clipping radius.}~We sweep $r\in\{1,2,5,10,10^{9}\}$ at fixed $K{=}{-}0.3$. The $r{=}10^{9}$ row effectively removes clipping. Clipping the tangent vector to norm $r$ places Klein and Lorentz features at geodesic radius $\rho{=}r$ but Poincar\'{e} features at $\rho{=}2r$, since the Poincar\'{e} conformal factor equals $2$ at the origin, so \cref{tab:stab-clip} indexes the sweep by $\rho$ (all radii in \cref{tab:stab-clip-full}). Klein remains finite at every radius, whereas Poincar\'{e} diverges at $\rho{=}4$ in two of three seeds and in every seed from $\rho{=}5$, and Lorentz diverges only without clipping.

\begin{wraptable}{r}{0.44\textwidth}
\vspace*{-1\baselineskip}
\centering
\begingroup%
\caption{Predicted FP32 failing offset and NaN configurations of each head.}
\label{tab:stab-fp32}
\setlength{\tabcolsep}{9pt}
\renewcommand{\arraystretch}{1.0}
\setlength{\aboverulesep}{1pt}\setlength{\belowrulesep}{1pt}
\footnotesize
\adjustbox{max width=\linewidth}{%
\begin{tabular}{@{}l|>{\columncolor{blue!8}}ccc@{}}
\toprule
 & \textbf{Klein} & \textbf{Poincar\'{e}} & \textbf{Lorentz} \\
\midrule
Failing offset & $87$ & ${\le}\,45$ & $8.3$--$9.0$ \\
NaN configs & $0/10$ & $6/10$ & $2/10$ \\
\bottomrule
\end{tabular}}
\endgroup
\end{wraptable}
\textbf{Where the baselines fail.}~For every diverging run, we record the first operation that returns a NaN or an infinity (\cref{app:stability-exp}). It is always one step of the classifier head, reached once the distance of a hyperplane from the origin, its offset, is too large for FP32 in that step. \cref{tab:stab-fp32} predicts this offset from the limits of FP32, and the runs fail at $40$--$45$ for Poincar\'{e} and $8.4$--$9.0$ for Lorentz. The Lorentz head~\citep{Bdeir2024} takes the square root of the difference of two nearly equal large numbers, which rounds to $0$, and the Poincar\'{e} head~\citep{Shimizu2021} squares a large number, which overflows. The Klein head has neither step.

Rewriting this Lorentz step or computing it in FP64 removes the Lorentz failures. The Poincar\'{e} failures remain in FP64, because with its default radius and initialization the offset keeps growing until the output overflows, whereas the Klein head trains with the radius and initialization of either model (\cref{app:stability-exp}). Klein coordinates have their own FP32 limit, at geodesic radius $8.7$ against $17.3$ for Poincar\'{e}~\citep{Mishne2023}, and one of $15$ Klein runs at $K{=}{-}1$ and $r{=}10$ diverges unless the features are projected back into the ball. Lorentz matches Klein under composition, and all three collapse at sufficient depth (\cref{app:ordering,app:capability,app:numerical-extra,app:depth}).

\vspace{-1em}

\subsection{Image Classification}
\label{sec:exp-image}
\vspace{-0.5em}
\textbf{Setup.}
Following~\citet{Guo2022}, we adopt a ResNet-18~\citep{He2016} Euclidean backbone and replace only the final classifier, comparing the two Klein heads against a Euclidean baseline and the six hyperbolic heads of \cref{tab:image-cls} on CIFAR-10/100~\citep{Krizhevsky2009} over five seeds, and all nine heads on ImageNet-1K~\citep{Deng2009} following \citet{Chen2026} (\cref{app:image-exp}).

\textbf{Results.}
Within the geodesic MLR family, Klein MLR reaches the accuracy of the strongest hyperbolic heads on CIFAR-10, CIFAR-100 and ImageNet-1K (\cref{tab:image-cls}), as isometry predicts. Both Klein heads also carry Euclidean-scale parameter counts, half that of PBMLR-P. Klein BMLR attains the highest ImageNet-1K top-1 among all heads.

\begin{table}[ht]
\centering
\caption{Hyperbolic MLR accuracy (\%) on a ResNet-18 backbone. Best in \textbf{bold}, largest \#Params in \textcolor{red!70!black}{red}.}
\label{tab:image-cls}
\footnotesize
\setlength{\tabcolsep}{4pt}
\renewcommand{\arraystretch}{0.9}
\adjustbox{max width=\textwidth}{%
\begin{tabular}{lclcccccc}
\toprule
\multirow{2}{*}{\textbf{Family}} & \multirow{2}{*}{\textbf{Space}} & \multirow{2}{*}{\textbf{Method}}
 & \multicolumn{2}{c}{\textbf{CIFAR-10} (C{=}10)}
 & \multicolumn{2}{c}{\textbf{CIFAR-100} (C{=}100)}
 & \multicolumn{2}{c}{\textbf{ImageNet-1K} (C{=}1000)} \\
\cmidrule(lr){4-5}\cmidrule(lr){6-7}\cmidrule(lr){8-9}
 & & & Acc & \#Params & Acc & \#Params & Top-1 & \#Params \\
\midrule
Euclidean & $\RR^n$ & Euclidean MLR~\citep{He2016}
 & \second{95.14$\pm$0.12} & 5.13K
 & 77.72$\pm$0.15 & 51.30K & 71.89 & 513.00K \\
\midrule
Pseudo & $\PP_K^n$ & PBMLR-P~\citep{Nguyen2025}
   & \second{95.23$\pm$0.08} & \textcolor{red!70!black}{10.24K}
   & 77.78$\pm$0.15 & \textcolor{red!70!black}{102.40K} & 71.57 & \textcolor{red!70!black}{1024K} \\
\midrule
\multirow{4}{*}{Geodesic} & $\PP_K^n$ & PMLR~\citep{Shimizu2021}
   & 95.04$\pm$0.13 & 5.14K
   & 77.19$\pm$0.50 & 51.40K & 71.70 & 514.00K \\
& $\LL_K^n$ & LMLR~\citep{Bdeir2024}
   & 94.98$\pm$0.12 & 5.13K
   & \second{78.03$\pm$0.21} & 51.30K & 72.50 & 513.00K \\
& $\PV_K^n$ & PV MLR~\citep{Chen2026PVNN}
   & \second{95.22$\pm$0.04} & 5.13K
   & \second{78.05$\pm$0.17} & 51.30K & 72.46 & 513.00K \\
\rowcolor{blue!8}
& $\KKn$ & \textbf{Klein MLR}
   & \first{95.29$\pm$0.15} & 5.13K
   & \first{78.25$\pm$0.13} & 51.30K & \first{73.16} & 513.00K \\
\midrule
\multirow{3}{*}{Busemann} & $\PP_K^n$ & BMLR-P~\citep{Chen2026}
   & \second{95.32$\pm$0.14} & 5.14K
   & \second{78.10$\pm$0.35} & 51.40K & 73.30 & 514.00K \\
& $\LL_K^n$ & BMLR-L~\citep{Chen2026}
   & \second{95.25$\pm$0.02} & 5.14K
   & \second{78.07$\pm$0.26} & 51.40K & 73.24 & 514.00K \\
\rowcolor{blue!8}
& $\KKn$ & \textbf{Klein BMLR}
   & \first{95.33$\pm$0.08} & 5.14K
   & \first{78.24$\pm$0.13} & 51.40K & \first{73.41} & 514.00K \\
\bottomrule
\end{tabular}
}
\end{table}

\vspace{-0.8em}
\subsection{Link Prediction}
\label{sec:exp-lp}
\vspace{-0.6em}
\textbf{Setup.}
We compare our Klein FC and Klein BFC against prior hyperbolic FC layers, M\"{o}bius FC~\citep{Ganea2018} (tangent space), Lorentz FC~\citep{Chen2022LGCN} (ambient Minkowski space), PLFC~\citep{Shi2026} and PV FC~\citep{Chen2026PVNN} (geodesic hyperplanes), and BFC-P/BFC-L~\citep{Chen2026} (horospheres). Following~\citet{Chami2019,Chen2026PVNN}, we evaluate on four graph datasets (Disease~\citep{Anderson1991}, Airport~\citep{Zhang2018LinkPred}, PubMed~\citep{Namata2012}, Cora~\citep{Sen2008}) ordered by Gromov $\delta$-hyperbolicity. All methods share a 2-layer FC encoder and a Fermi--Dirac decoder, every FC layer is followed by the batch normalization of its own model and a ReLU, and each method selects its learning rate, from $0.001$ to $0.015$, dropout and weight decay on one grid (\cref{app:graph-exp}).

\textbf{Results.}
Klein FC is the most accurate layer on Airport and is not significantly below the best layer, M\"{o}bius FC, on Disease and Cora, and it is $0.38$ points below M\"{o}bius FC on PubMed. Among the layers not built in the tangent space, Klein FC has the highest accuracy on all four datasets, by $0.40$ to $1.96$ points (\cref{tab:sig-graph}). The two Klein layers have the shortest training time per epoch on every dataset, $1.6$ to $2.1$ times shorter than M\"{o}bius FC (\cref{tab:lp-time}).

\vspace{-2mm}
\subsection{Node Classification}
\vspace{-3mm}
\label{sec:exp-nc}

\textbf{Setup.}
We study node classification on the datasets of \cref{sec:exp-lp}. All methods share a two-FC-plus-MLR architecture, differing only in the hyperbolic model, with the baselines listed in \cref{tab:nc}. Klein-HNN++ pairs Klein FC with Klein MLR and Klein-HBNN++ Klein BFC with Klein BMLR.

\textbf{Results.}
\cref{tab:nc} reports accuracy over five seeds. Under the same protocol (bottom block), Klein-HNN++ is statistically indistinguishable from PVNN on Disease, Airport and PubMed and $6.94$ points higher on Cora ($p=0.001$), while Klein-HBNN++ is $1.38$ points below PVNN on PubMed ($p=0.04$, not significant after Holm correction). 

\vspace{-0.5em}
\subsection{Ablation on MLR and GyroBN}
\label{sec:exp-bn-ablation}
\vspace{-0.5em}

\textbf{MLR runtime.}
\cref{tab:mlr-runtime} records per-iteration wall time for each MLR head on random Euclidean features at batch $1024$, $100$ classes and FP32. The Klein variants are the fastest hyperbolic heads at every input dimension, by a small margin of $1.13$--$1.20\times$ over BMLR-L, with $1.49$--$1.55$\,ms against $1.68$--$2.53$\,ms for the Poincar\'{e}, Lorentz and PV heads.

\textbf{GyroBN gyromean.}
\cref{tab:bn-ablation} records accuracy and fit time for five Fr\'{e}chet truncations, the Einstein midpoint and no BN, all on one Klein FC backbone. The midpoint is the fastest on all four datasets, $1.9\times$ to $14\times$ ahead of the converged Fr\'{e}chet mean, and matches its accuracy on three. On Airport the Fr\'{e}chet mean degrades as it converges, from $98.09\pm0.57$ at one iteration to $85.67\pm7.99$ at convergence, while the midpoint reaches $97.59\pm0.68$ in one pass with no iteration count to choose. Against the unnormalized backbone it adds $6.2$ points on Cora and under one elsewhere.

\begin{table}[!tbp]
\centering
\caption{FC layers on link prediction (ROC AUC \%), 5 seeds, each followed by the batch normalization of its own model and tuned on one grid. Bold is best, underlined is not significantly worse.}
\footnotesize
\label{tab:link-pred}
\renewcommand{\arraystretch}{0.9}
\setlength{\tabcolsep}{6pt}
\adjustbox{max width=\textwidth}{%
\begin{tabular}{cl cccc}
\toprule
\textbf{Space} & \textbf{Method}
 & \textbf{Disease} ($\delta{=}0$)
 & \textbf{Airport} ($\delta{=}1$)
 & \textbf{PubMed} ($\delta{=}3.5$)
 & \textbf{Cora} ($\delta{=}11$) \\
\midrule
\multirow{2}{*}{$\PP_K^n$}
 & M\"{o}bius FC~\citep{Ganea2018} & \first{83.09$\pm$0.43} & \second{95.85$\pm$0.10} & \first{96.96$\pm$0.05} & \first{94.69$\pm$0.21} \\
 & BFC-P~\citep{Chen2026} & 76.38$\pm$1.11 & 94.64$\pm$0.36 & 90.99$\pm$0.53 & 86.28$\pm$1.96 \\
\midrule
\multirow{3}{*}{$\LL_K^n$}
 & Lorentz FC~\citep{Chen2022LGCN} & 64.71$\pm$5.83 & \second{93.90$\pm$1.30} & 96.18$\pm$0.18 & 93.94$\pm$0.29 \\
 & PLFC~\citep{Shi2026} & \second{80.70$\pm$1.48} & 95.02$\pm$0.13 & 95.04$\pm$0.05 & 91.43$\pm$0.82 \\
 & BFC-L~\citep{Chen2026} & 79.95$\pm$0.52 & \second{95.29$\pm$0.38} & 95.14$\pm$0.08 & 91.57$\pm$0.33 \\
\midrule
$\PV_K^n$
 & PV FC~\citep{Chen2026PVNN} & 71.74$\pm$1.92 & 93.65$\pm$0.27 & 95.61$\pm$0.06 & 93.93$\pm$0.18 \\
\midrule
\rowcolor{blue!8}
 & Klein FC & \second{82.66$\pm$0.68} & \first{95.86$\pm$0.09} & 96.58$\pm$0.07 & \second{94.38$\pm$0.26} \\
\rowcolor{blue!8}
\multirow{-2}{*}{$\KKn$} & Klein BFC & 79.77$\pm$0.84 & \second{95.79$\pm$0.08} & 95.70$\pm$0.24 & 92.46$\pm$0.31 \\
\bottomrule
\end{tabular}
}
\end{table}
\begin{table}[!tbp]
\centering
\caption{Gyromean methods inside the Klein GyroBN layer. Fastest fit time (ms/epoch) in {\color{green!60!black}green}, slowest in {\color{red}red}.}
\label{tab:bn-ablation}
\footnotesize
\setlength{\tabcolsep}{4pt}
\renewcommand{\arraystretch}{0.9}
\adjustbox{max width=\textwidth}{%
\begin{tabular}{l | cc | cc | cc | cc}
\toprule
\multirow{2}{*}{\textbf{Method}}
 & \multicolumn{2}{c|}{\textbf{Disease}}
 & \multicolumn{2}{c|}{\textbf{Airport}}
 & \multicolumn{2}{c|}{\textbf{PubMed}}
 & \multicolumn{2}{c}{\textbf{Cora}} \\
\cmidrule(lr){2-3}\cmidrule(lr){4-5}\cmidrule(lr){6-7}\cmidrule(lr){8-9}
& Acc & Fit Time
& Acc & Fit Time
& Acc & Fit Time
& Acc & Fit Time \\
\midrule
Klein-HNN++ (no BN)
  & 81.15$\pm$0.26 & 44.0
  & 97.46$\pm$0.91 & 112.7
  & 73.32$\pm$0.78 & 7.6
  & 51.80$\pm$2.18 & 14.1 \\
\midrule
Fr\'{e}chet $1$ iter
  & 80.96$\pm$0.40 & 101.2
  & \textbf{98.09$\pm$0.57} & 274.0
  & 72.44$\pm$1.22 & 16.4
  & 57.80$\pm$0.86 & 31.7 \\
Fr\'{e}chet $2$ iters
  & 80.77$\pm$0.21 & 121.6
  & 96.83$\pm$0.65 & 335.2
  & 72.72$\pm$1.39 & 19.5
  & 57.86$\pm$0.77 & 37.5 \\
Fr\'{e}chet $5$ iters
  & \textbf{81.24$\pm$0.71} & 184.9
  & 92.73$\pm$3.62 & 546.9
  & 72.14$\pm$0.87 & 28.0
  & \textbf{58.06$\pm$1.46} & 54.9 \\
Fr\'{e}chet $10$ iters
  & 80.77$\pm$0.40 & 290.9
  & 89.40$\pm$3.12 & \textcolor{red}{782.0}
  & 72.90$\pm$0.87 & \textcolor{red}{41.9}
  & 57.16$\pm$0.70 & \textcolor{red}{88.7} \\
Fr\'{e}chet $\infty$
  & 80.77$\pm$0.71 & \textcolor{red}{396.2}
  & 85.67$\pm$7.99 & \textcolor{red}{3116.4}
  & \textbf{73.08$\pm$0.68} & 25.9
  & 57.60$\pm$0.37 & 59.4 \\
\midrule
\rowcolor{blue!8}
\textbf{Einstein midpoint}
  & 80.57$\pm$0.26 & \textbf{\textcolor{green!60!black}{81.1}}
  & 97.59$\pm$0.68 & \textbf{\textcolor{green!60!black}{221.3}}
  & 72.50$\pm$0.64 & \textbf{\textcolor{green!60!black}{13.9}}
  & 58.04$\pm$0.43 & \textbf{\textcolor{green!60!black}{26.2}} \\
\bottomrule
\end{tabular}}
\end{table}

\vspace{-0.6em}
\subsection{Fully Hyperbolic ResNet}
\label{sec:exp-resnet}
\vspace{-0.5em}
\textbf{Setup.}
Following \citet{Bdeir2025Robust}, we compare Klein-ResNet with Poincar\'{e} ResNet, HCNN, ILNN and a Euclidean ResNet-18 on CIFAR-100 and Tiny-ImageNet. Klein-ResNet follows the protocol of \citet{Bdeir2024}, $200$ epochs of Riemannian SGD at batch size $128$; baseline accuracies are from the original papers. Every model is timed on one A100 with \texttt{torch.compile}~\citep{Ansel2024PyTorch2}, using the released code of Poincar\'{e} ResNet and HCNN and our ILNN implementation.
\begin{table}[!htb]
\centering
\caption{ResNet-18 on CIFAR-100 and Tiny-ImageNet. -- marks a run that diverges.}
\label{tab:resnet}
\setlength{\tabcolsep}{4pt}
\renewcommand{\arraystretch}{0.8}
\setlength{\aboverulesep}{1pt}\setlength{\belowrulesep}{1pt}
\footnotesize
\adjustbox{max width=\linewidth}{%
\begin{tabular}{@{}l|ccc|ccc@{}}
\toprule
 & \multicolumn{3}{c|}{\textbf{CIFAR-100}} & \multicolumn{3}{c}{\textbf{Tiny-ImageNet}} \\
\textbf{Method} & \textbf{Acc.} & \textbf{VRAM} & $\bm{t_{\text{epoch}}}$ & \textbf{Acc.} & \textbf{VRAM} & $\bm{t_{\text{epoch}}}$ \\
\midrule
Euclidean~\citep{He2016} & 77.72$\pm$0.15 & 0.8\,GB & 4.9\,s & 65.19$\pm$0.12 & 2.5\,GB & 29.9\,s \\
Poincar\'{e}~\citep{vanSpengler2023} & 76.60$\pm$0.32 & 14.6\,GB & 195.3\,s & 62.01$\pm$0.56 & 57.6\,GB & 949.7\,s \\
ILNN~\citep{Shi2026} & 78.41$\pm$0.23 & 11.2\,GB & 56.5\,s & -- & 44.2\,GB & 372.4\,s \\
HCNN~\citep{Bdeir2024} & 78.07$\pm$0.17 & 7.7\,GB & 62.4\,s & 65.71$\pm$0.13 & 30.3\,GB & 270.3\,s \\
\rowcolor{blue!8}
\textbf{Klein-ResNet} & 78.09$\pm$0.20 & 2.8\,GB & 15.6\,s & 63.03$\pm$0.25 & 10.5\,GB & 110.7\,s \\
\bottomrule
\end{tabular}}
\end{table}

\textbf{Results.}
On CIFAR-100 all five networks reach similar accuracy, and Klein-ResNet is the fastest fully hyperbolic network in \cref{tab:resnet}, training $4.0$ times faster than the reference implementation of HCNN with about a third of its GPU memory. Layer-wise timing traces this saving to the batch normalization and to the convolution, which runs as two standard convolutions (\cref{prop:fc-conv}) where the reference HCNN copies every receptive field into a separate patch (\cref{tab:spd-conv,tab:spd-bn}). With the patch copies and the \texttt{ManifoldParameter} shift removed by hand, HCNN matches the speed of Klein-ResNet, which still uses about a fifth less GPU memory (\cref{tab:spd-swap}, \cref{app:resnet-exp}). ILNN diverges on Tiny-ImageNet in every seed (\cref{app:resnet-exp}).

\vspace{-0.7em}
\section{Conclusion}
\label{sec:conclusion}
\vspace{-0.7em}
Klein-HNN++ derives five closed-form layers from three Klein-native primitives. In the two stability sweeps, the Klein layers train in all ten configurations, reaching $70.88\%$ to $74.54\%$ on CIFAR-100, whereas the reference Poincar\'{e} and Lorentz implementations diverge in six and two of them. The layers match their isometric counterparts in accuracy on image and node classification, and the closed-form Einstein midpoint replaces iterative Fr\'{e}chet means at a fraction of their cost. Klein-ResNet trains $4.0$ times faster than the reference HCNN with a third of its memory, and even a hand-optimized HCNN that matches its speed still needs more memory. Future work extends the layers to attention, graph convolution and hyperbolic Transformers~\citep{Yang2024Hypformer}.

\clearpage

\section*{AI Use Statement}

We used generative AI tools only for language polishing and for translating
mathematical expressions we had already derived into code, not for research
ideas, derivations, or experiments. All AI-assisted output was reviewed by the
authors, and we take responsibility for the final content of this work.

\section*{Ethics Statement}

This work studies layer-level operators for hyperbolic neural networks and
raises no concerns under the ICLR Code of Ethics. All experiments use publicly
available standard benchmarks, and the work involves no human subjects and no
newly released data.

\section*{Reproducibility Statement}

The five proposed layers are given in closed form in \cref{sec:klein-nn} and
proved in \cref{app:proofs}, so the architecture can be reimplemented directly
from the paper text. \cref{app:experiments} reports the full experimental
protocol, including the training recipes, the per-dataset hyperparameters, the
seed counts and dataset splits used throughout, and the hardware, together with
the significance tests behind the reported comparisons and the per-seed numbers
they are computed from.

\newpage
\bibliography{references}

\addtocontents{toc}{\protect\setcounter{tocdepth}{2}}

\clearpage
\begin{center}
  {\LARGE\bfseries Klein Neural Networks++}\\[0.6em]
  {\Large Supplementary Material}
\end{center}
\vspace{1em}

\appendix

\renewcommand{\floatpagefraction}{0.9}
\renewcommand{\bottomfraction}{0.7}

\setcounter{tocdepth}{2}
{\hypersetup{linkcolor=black}%
\renewcommand{\contentsname}{Appendix Contents}%
\renewcommand{\baselinestretch}{0.92}\selectfont
\tableofcontents}%
\clearpage

\section{Notation}
\label{app:notation}

\cref{tab:notation} summarizes the notation used throughout the paper.

\begin{table}[!ht]
\centering
\caption{Summary of notation.}
\label{tab:notation}
\renewcommand{\arraystretch}{1.15}
\begin{tabular}{@{}ll@{}}
\toprule
\textbf{Notation} & \textbf{Description} \\
\midrule
\multicolumn{2}{@{}l}{\emph{Riemannian geometry}} \\
$\mathcal{M}$                              & Riemannian manifold \\
$T_x\mathcal{M}$                           & Tangent space at $x$ \\
$g_x(u,v)$                                 & Riemannian metric at $x$ \\
$d(x,y)$                                   & Geodesic distance \\
$\Expop_x(v)$                              & Exponential map at $x$ \\
$\Logop_x(y)$                              & Logarithmic map at $x$ \\
$\PT_{x\to y}(v)$                          & Parallel transport of $v$ from $x$ to $y$ \\
$\gamma(t)$                                & Unit-speed geodesic ray \\
$B^{\gamma}(x)$                            & Busemann function associated with $\gamma$ \\
$B^{v}(x)$                                 & Busemann function associated with unit direction $v\!\in\!\Sn^{n-1}$ \\
$H^{v}_{\tau}$                             & Horosphere $\{x\mid B^{v}(x){=}\tau\}$ \\
$H_{a,p}$                                  & Hyperbolic hyperplane parameterized by $p$ and tangent normal $a$ \\
\midrule
\multicolumn{2}{@{}l}{\emph{Euclidean}} \\
$\RR^n$                                    & Euclidean space of dimension $n$ \\
$\langle\cdot,\cdot\rangle$, $\|\cdot\|$   & Euclidean inner product and norm \\
$\mathbf{0}$                               & Zero vector in $\RR^n$ \\
$\Sn^{n-1}$                                & Unit sphere in $\RR^n$ \\
$e_k\!\in\!\RR^m$                          & $k$-th standard basis vector \\
\midrule
\multicolumn{2}{@{}l}{\emph{Hyperbolic models (curvature $K\!<\!0$)}} \\
$\KKn$                                     & Klein ball model in $\RR^n$ with curvature $K$ \\
$\PP_K^n$                                  & Poincar\'{e} ball model in $\RR^n$ with curvature $K$ \\
$\LL_K^n$                                  & Lorentz hyperboloid model in $\RR^{1+n}$ with curvature $K$ \\
$\PV_K^n$                                  & Proper-velocity manifold with curvature $K$ \\
$\langle x,y\rangle_{\LL}$                 & Lorentzian inner product $\langle x_s,y_s\rangle - x_t y_t$ \\
$x{=}[x_t,x_s^\top]^\top$                  & Time and spatial decomposition in the Lorentz ambient space \\
\midrule
\multicolumn{2}{@{}l}{\emph{Gyrovector operations}} \\
$\oE,\,\otE$                               & Einstein gyroaddition and scalar gyromultiplication on $\KKn$ \\
$\ominus_{\mathrm{E}}$                     & Einstein inverse \\
$\oM,\,\otM$                               & M\"{o}bius gyroaddition and scalar gyromultiplication on $\PP_K^n$ \\
$\gamma_x{=}1/\sqrt{1+K\|x\|^2}$           & Klein Lorentz factor \\
$\lambda_x^K{=}2/(1+K\|x\|^2)$             & Poincar\'{e} conformal factor \\
$\mathrm{gyr}[x,y]$                        & Gyration automorphism induced by $x,y$ \\
\midrule
\multicolumn{2}{@{}l}{\emph{Layers and learning}} \\
$C$                                        & Number of classes \\
$m,\,n$                                    & Output / input dimension of an FC layer \\
$u_k(x)$                                   & Logit for class $k$ \\
$\phi:\RR\!\to\!\RR$                       & Activation function \\
$\sign(\cdot)$                             & Sign function \\
$\mathrm{softmax}$                         & Softmax operator \\
$\FM(\{x_i\})$                             & Fr\'{e}chet mean of $\{x_i\}$ on $\KKn$ \\
$\mu_{\mathrm{E}},\,\mu_{\mathrm{F}}$      & Einstein midpoint, Fr\'{e}chet mean \\
\bottomrule
\end{tabular}
\end{table}

\section{Extended Related Work}
\label{app:ext-related}

This appendix extends the related-work discussion in \cref{sec:introduction} with additional context on hyperbolic representation learning, batch normalization, graph networks, and genomic applications that we omitted from the main text for space.

\subsection{Hyperbolic Representation Learning}
\label{app:ext-repr}

The use of hyperbolic geometry for machine learning was pioneered by \citet{Nickel2017}, who showed that Poincar\'{e} embeddings can represent WordNet-scale taxonomies with far fewer dimensions than Euclidean embeddings, owing to the exponential volume growth of hyperbolic space. \citet{Nickel2018} extended this approach to the Lorentz model, demonstrating improved optimization stability. \citet{Sala2018} provided a theoretical analysis of the representation tradeoffs, proving that trees with bounded branching factor embed in $\PP^2$ with arbitrarily low distortion. \citet{Khrulkov2020} applied hyperbolic embeddings to computer vision, showing that image features exhibit latent hierarchical structure that is better captured in hyperbolic space. These foundational works establish the motivation for our approach. If data is better represented in hyperbolic space, then operating on it with hyperbolic-native neural network layers should be beneficial.

\subsection{Hyperbolic Batch Normalization and Deep Architectures}

Extending batch normalization to hyperbolic space requires notions of mean, variance, and rescaling on curved manifolds. \citet{Lou2020} proposed differentiating through the Fr\'{e}chet mean, enabling its use as a centering operation in deep networks. \citet{vanSpengler2023} introduced Poincar\'{e} ResNet, which combines the M\"{o}bius midpoint for batch normalization, M\"{o}bius residual connections, and convolutions that apply the Poincar\'{e} FC layer of \citet{Shimizu2021} to the $\beta$-concatenation of each receptive field to build deep hyperbolic residual networks. \citet{Bdeir2024} proposed Lorentz batch normalization based on the Lorentz centroid. HCNN~\citep{Bdeir2024} and ILNN~\citep{Shi2026} build fully hyperbolic Lorentz CNNs whose convolutions apply a hyperbolic FC layer to every unfolded receptive field, which the Klein convolution of \cref{sec:klein-resnet} avoids. Recent work on Klein networks~\citep{Mao2024} establishes the underlying gyrostructure but has not explored the Einstein midpoint as a BN centroid. We develop this connection in \cref{sec:klein-bn}.

\subsection{Hyperbolic Graph Neural Networks}
\label{app:ext-graph}

\citet{Chami2019} proposed Hyperbolic Graph Convolutional Networks (HGCN), combining graph convolutional layers with hyperbolic geometry via tangent-space message passing and hyperbolic MLR classifiers. HGCN demonstrated significant gains over Euclidean GCNs on link prediction for datasets with strong hierarchical structure (e.g., Disease and Airport). Subsequent work has explored Lorentz graph networks~\citep{Chen2022LGCN}. Our link prediction experiments follow the HGCN framework, replacing its Poincar\'{e} and Lorentz FC and MLR layers with Klein FC, BFC, and BMLR layers to enable a controlled comparison of the underlying geometry.

\subsection{Genome and Biological Sequence Classification}
\label{app:ext-genome}

Hyperbolic geometry has recently found applications in computational biology, motivated by the tree-like phylogenetic structure of biological sequences. \citet{Khan2025HGE} introduced the Transposable Element Benchmark (TEB), a suite of genomic classification tasks where hyperbolic models should excel due to the hierarchical taxonomy of transposable element families, demonstrating that Poincar\'{e} and Lorentz models can improve over Euclidean baselines on TEB tasks, particularly for multi-class problems with deep class hierarchies.

\section{Preliminaries}
\label{app:prelim}

This appendix provides the technical background for the main paper. \cref{app:riemannian} reviews Riemannian geometry, \cref{app:gyro} states the gyrovector-space axioms, and \cref{app:hyp-models} collects the Riemannian operators and gyrovector structures of the three canonical hyperbolic models (Klein, Poincar\'{e}, Lorentz).

\subsection{Riemannian Geometries}
\label{app:riemannian}

For an in-depth discussion on Riemannian geometry, one can refer to~\citet{Lee2018}.

A \emph{Riemannian manifold} $(\mathcal{M}, g)$ carries a smoothly varying inner product $g_x: T_x\mathcal{M}\times T_x\mathcal{M}\to\RR$ on each tangent space. The induced norm is $\|v\|_x = \sqrt{g_x(v,v)}$. \emph{Geodesics} are constant-speed curves that locally minimize length. The shortest distance between $x,y\in\mathcal{M}$ is the geodesic distance $\mathrm{d}(x,y)$.

The \emph{exponential map} $\Expop_x: T_x\mathcal{M}\supset\mathcal{V}\to\mathcal{M}$ sends $v$ to $\gamma_{x,v}(1)$, where $\gamma_{x,v}$ is the geodesic with $\gamma(0)=x$, $\dot\gamma(0)=v$. Its local inverse is the \emph{logarithmic map} $\Logop_x$.
\emph{Parallel transport}
$\PT_{x\to y}: T_x\mathcal{M}\to T_y\mathcal{M}$ is the isometric transport of tangent vectors along the geodesic from $x$ to $y$. A smooth map $f:(\mathcal{M},g)\to(\widetilde{\mathcal{M}},\tilde g)$ between Riemannian manifolds is a \emph{Riemannian isometry} if $g_x(u,v)=\tilde g_{f(x)}(d_xf(u),d_xf(v))$ for all $u,v\in T_x\mathcal{M}$, where $d_xf$ is the differential at $x$.

\cref{tab:riemannian-ops} summarizes the correspondence between Euclidean and Riemannian operations.

\begin{table}[!htbp]
\centering
\caption{Euclidean operations and their Riemannian generalizations.}
\label{tab:riemannian-ops}
\renewcommand{\arraystretch}{1.25}
\setlength{\tabcolsep}{12pt}
\begin{tabular}{lcc}
\toprule
\textbf{Operation} & \textbf{Euclidean} & \textbf{Riemannian} \\
\midrule
Straight line & Straight line & Geodesic \\
Subtraction  & $\overrightarrow{xy} = y-x$
             & $\overrightarrow{xy} = \Logop_x(y)$ \\
Addition     & $y = x + \overrightarrow{xy}$
             & $y = \Expop_x(\overrightarrow{xy})$ \\
Parallel move & $v\to v$ & $\PT_{x\to y}(v)$ \\
\bottomrule
\end{tabular}
\end{table}

\subsection{Busemann Function}
\label{app:busemann}

The Busemann function and its level sets (horospheres) are introduced in \cref{sec:preliminary}, \cref{eq:busemann-def}. Here we collect the closed-form Busemann functions on the three hyperbolic models. For a unit direction $v\in\Sn^{n-1}$ at the origin,
\begin{align}
\KKn:\;\; B^v_{\KK}(x) &= \frac{1}{\sqrt{-K}}\log\!\left(
  \frac{1-\sqrt{-K}\langle x,v\rangle}{\sqrt{1+K\|x\|^2}}\right),
\label{eq:busemann-klein-summary}\\[2pt]
\PP_K^n:\;\; B^v_{\PP}(x) &= \frac{1}{\sqrt{-K}}\log\!\left(
  \frac{\|v-\sqrt{-K}\,x\|^2}{1+K\|x\|^2}\right),
\label{eq:busemann-poincare}\\[2pt]
\LL_K^n:\;\; B^v_{\LL}(x) &= \frac{1}{\sqrt{-K}}\log\!\left(
  \sqrt{-K}\,(x_t - \langle x_s,v\rangle)\right).
\label{eq:busemann-lorentz}
\end{align}
The Klein expression is derived as \cref{thm:klein-busemann} in the main paper. The Poincar\'{e} and Lorentz expressions follow~\citet{Chen2026}\,[Prop.~9]. The three forms are mutually consistent under the Klein--Poincar\'{e} and Klein--Lorentz isometries of \crefrange{eq:klein-poincare-map}{eq:klein-lorentz-map}, verified explicitly in \cref{app:busemann-verify}.

\textbf{Horosphere geometry.}
In each model the level sets of $B^v$ are horospheres $H^v_\tau=\{x\mid B^v(x)=\tau\}$. On $\KKn$ and $\PP_K^n$ the horospheres are ellipsoids tangent internally to the boundary sphere. On $\LL_K^n$ they are intersections of the upper hyperboloid with affine planes parallel to the asymptotic null cone. As $K\to 0^-$, all three Busemann functions converge to $-\langle x,v\rangle$ and the horospheres degenerate to the Euclidean hyperplanes $H^v_\tau=\{x\in\RR^n\mid\langle x,v\rangle=\tau\}$.

\textbf{Horospheres as complementary decision boundaries.}
\cref{sec:klein-mlr} and \cref{sec:klein-bmlr} provide geodesic and horospherical
decision boundaries, respectively. The Busemann head complements the geodesic formulation
through the intrinsic hyperbolic counterpart of the Euclidean inner product.

\textbf{Euclidean-to-hyperbolic correspondence.} \cref{tab:eucl-hyp-dictionary} collects the correspondence.
The Euclidean MLR logit is $\langle a_k,x\rangle+b_k$; replacing the inner product by
its intrinsic counterpart, the Busemann function of \cref{eq:busemann-def}, yields
\cref{eq:klein-bmlr} and changes nothing else. This is why \cref{eq:klein-bmlr-expanded}
recovers $\alpha_k\langle v_k,x\rangle+b_k$ exactly as $K\to 0^-$ (\cref{app:proof-bmlr}):
the Busemann head is the deformation of the Euclidean softmax that leaves the flat case
intact.

\begin{table}[!htbp]
\centering
\caption{Euclidean notions and their intrinsic hyperbolic counterparts, following
\citet{Chen2026}\,[Tab.~1].}
\label{tab:eucl-hyp-dictionary}
\begin{tabular}{@{}lll@{}}
\toprule
\textbf{Euclidean} & \textbf{Hyperbolic} & \textbf{In this paper} \\
\midrule
straight line & geodesic & Klein: Euclidean line segment \\
parallel lines & asymptotic geodesic rays & \cref{sec:preliminary} \\
inner product $\langle x,v\rangle$ & Busemann function $-B^v(x)$ &
  \crefrange{eq:busemann-def}{eq:busemann-euclidean} \\
hyperplane $\{\langle x,v\rangle{=}\tau\}$ & horosphere $\{B^v(x){=}\tau\}$ &
  \cref{thm:klein-busemann} \\
\emph{parallel} hyperplanes, fixed $v$ & \emph{equidistant} horospheres, fixed $v$ &
  \cref{sec:preliminary} \\
\bottomrule
\end{tabular}
\end{table}

\textbf{Metric parallelism.} In $\RR^n$ the level
sets of a linear functional are equidistant, which is what makes the reading
``logit $=$ signed margin'' meaningful. Passing to $\mathbb{H}^n$, exactly one of the two families
retains this. The level sets of a single Busemann function satisfy
$d(H^v_{\tau_1},H^v_{\tau_2})=|\tau_2-\tau_1|$ exactly, and the point-to-horosphere distance
is the true geodesic distance $|B^v(x)-\tau|$~\citep[Thm.~3.3, Cor.~3.4]{Chen2026}. Hence for
Klein BMLR every level set of the logit is again a horosphere of the same family, and
$|u_k(x)|=\alpha_k\,d(x,H_{v_k,\alpha_k,b_k})$ with no approximation. By contrast, two
distinct geodesic hyperplanes of $\mathbb{H}^n$ are never equidistant: they either intersect or
diverge. The level sets $\{x: d(x,H)=c\}$ of a geodesic head's logit are equidistant
hypersurfaces, which are not themselves totally geodesic, so only the decision boundary
$c=0$ lies in the intended family. The ``curved'' appearance of a horosphere in the Klein
chart is therefore a property of the chart, not a loss of flatness.

\begin{figure}[!htbp]
\centering
\includegraphics[width=\textwidth]{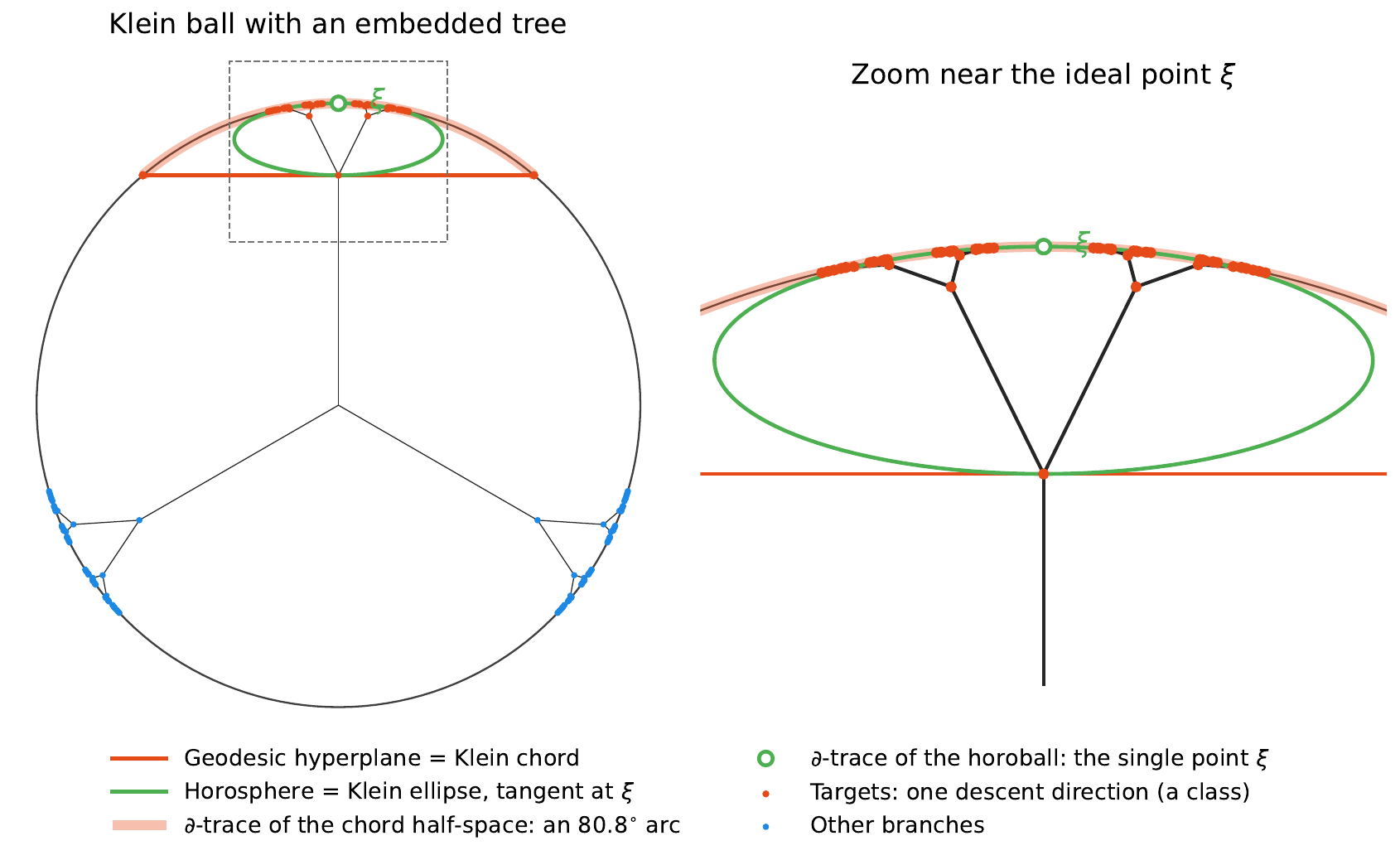}
\caption{Geodesic and horospherical boundaries at the same depth in the Klein ball.}
\label{fig:horosphere-vs-chord}
\end{figure}

\textbf{Ideal-boundary geometry.} A geodesic hyperplane meets the ideal
boundary in an $(n{-}2)$-sphere, so each of its half-spaces contains an entire boundary cap;
a horoball meets the ideal boundary at exactly one ideal point $\xi$. In hierarchical data a
class is a direction of descent, that is an ideal point, so the horosphere family is indexed
by the same object as the labels. \cref{fig:horosphere-vs-chord} illustrates this: both
surfaces cut the ray towards $\xi$ at the same point, hence at the same depth, yet the
chord's half-space already contains an $80.8^\circ$ arc of the ideal boundary while the
horoball touches it only at $\xi$.

In \cref{fig:horosphere-vs-chord}, both boundaries intersect the ray towards $\xi$ at the
same hyperbolic depth. The geodesic half-space spans a boundary arc, whereas the horoball
selects the single ideal direction $\xi$ and encloses its descendants.

\subsection{Gyro-Structures}
\label{app:gyro}

This subsection reviews the gyrovector-space framework of~\citet{Ungar2022}, which generalizes the vector-space structure to constant-curvature manifolds and underpins the Klein, Poincar\'{e} and Lorentz operators of \cref{app:hyp-models}. Gyrogroups generalize groups. Unlike groups, the binary operation is non-associative but satisfies a left gyroassociative law characterized by a family of automorphisms (gyrations).

\begin{definition}[Gyrogroups~\citep{Ungar2022}]
\label[definition]{def:gyrogroup}
A nonempty set $G$ with binary operation
$\oplus: G\times G\to G$ forms a \emph{gyrogroup} $(G,\oplus)$
if its binary operation satisfies the following axioms for any
$x,y,z\in G$.

\noindent\textbf{(G1)} There is at least one element $e\in G$
called a \emph{left identity} such that $e\oplus x=x$.

\noindent\textbf{(G2)} There is an element $\ominus x\in G$
called a \emph{left inverse} of $x$ such that
$\ominus x\oplus x=e$.

\noindent\textbf{(G3)} There is an automorphism
$\mathrm{gyr}[x,y]:G\to G$ for each $x,y\in G$ such that
\[
x\oplus(y\oplus z)=(x\oplus y)\oplus\mathrm{gyr}[x,y]z
\qquad(\text{Left Gyroassociative Law}).
\]

\noindent\textbf{(G4)} Left reduction law
$\mathrm{gyr}[x,y]=\mathrm{gyr}[x\oplus y,y]$.

\noindent The automorphism $\mathrm{gyr}[x,y]$ is called the
\emph{gyration} generated by $x,y$.
\end{definition}

\begin{definition}[Gyrocommutative gyrogroups~\citep{Ungar2022}]
A gyrogroup $(G,\oplus)$ is \emph{gyrocommutative} if it
satisfies $x\oplus y=\mathrm{gyr}[x,y](y\oplus x)$.
\end{definition}

A gyrocommutative gyrogroup equipped with a scalar gyromultiplication satisfying further regularity axioms is a gyrovector space.

\begin{definition}[Gyrovector spaces~\citep{Ungar2022}]
\label[definition]{def:gyrovector}
A gyrocommutative gyrogroup $(G,\oplus)$ equipped with scalar
gyromultiplication $\otimes:\RR\times G\to G$ is called a
\emph{gyrovector space} if it satisfies the following axioms
for any $s,t\in\RR$ and $x,y,z\in G$.

\noindent\textbf{(V1)} Identity scalar multiplication
$1\otimes x=x$.

\noindent\textbf{(V2)} Scalar distributive law
$(s+t)\otimes x=s\otimes x\oplus t\otimes x$.

\noindent\textbf{(V3)} Scalar associative law
$(st)\otimes x=s\otimes(t\otimes x)$.

\noindent\textbf{(V4)} Gyroautomorphism property
$\mathrm{gyr}[x,y](t\otimes z)=t\otimes\mathrm{gyr}[x,y]z$.

\noindent\textbf{(V5)} Identity gyroautomorphism
$\mathrm{gyr}[s\otimes x,t\otimes x]=\mathbb{I}$, where
$\mathbb{I}$ is the identity map.
\end{definition}

\begin{definition}[Real inner-product gyrovector spaces~\citep{Ungar2022}]
\label[definition]{def:rip-gyro}
Let $(G,\oplus,\otimes)$ be a gyrovector space with
$G\subseteq\RR^n$ inheriting the Euclidean inner product
$\langle\cdot,\cdot\rangle$ and norm $\|\cdot\|$.
$(G,\oplus,\otimes,\langle\cdot,\cdot\rangle)$ is a
\emph{real inner-product gyrovector space} if it further
satisfies the following axioms.

\noindent\textbf{(V6)} Inner product gyroinvariance
$\langle\mathrm{gyr}[x,y]u,\mathrm{gyr}[x,y]v\rangle=\langle u,v\rangle$
for all $u,v\in G$.

\noindent\textbf{(V7)} Scaling preserves direction
$\dfrac{|s|\otimes x}{\|s\otimes x\|}=\dfrac{x}{\|x\|}$
for $x\neq\mathbf{0}$ and $s\neq 0$.

\noindent\textbf{(V8)} Homogeneity
$\|s\otimes x\|=|s|\otimes\|x\|$.

\noindent\textbf{(V9)} Gyrotriangle inequality
$\|x\oplus y\|\leq\|x\|\oplus\|y\|$.
\end{definition}

\begin{definition}[Gyrovector space isomorphism~\citep{Ungar2022}]
\label[definition]{def:gyro-iso}
A bijection
$\phi:(G_1,\oplus_1,\otimes_1)\to(G_2,\oplus_2,\otimes_2)$
between two real inner-product gyrovector spaces is a
\emph{gyrovector space isomorphism} if for all $x,y\in G_1$ and
$t\in\RR$,
$\phi(x\oplus_1 y)=\phi(x)\oplus_2\phi(y)$,
$\phi(t\otimes_1 x)=t\otimes_2\phi(x)$, and the inner-product
ratio is preserved,
$\frac{\langle\phi(x),\phi(y)\rangle}{\|\phi(x)\|\,\|\phi(y)\|}
=\frac{\langle x,y\rangle}{\|x\|\,\|y\|}$.
\end{definition}

\begin{proposition}[Isomorphism preserves gyrations and identity
{\citep{Ungar2022}}]
\label[proposition]{prop:iso-gyrations}
If $\phi:G_1\to G_2$ is a gyrovector space isomorphism with
gyrations $\mathrm{gyr}_1$ and $\mathrm{gyr}_2$, then for all
$x,y,z\in G_1$,
$\phi(\mathrm{gyr}_1[x,y]z)=\mathrm{gyr}_2[\phi(x),\phi(y)]\phi(z)$,
$\phi(e_1)=e_2$, and $\phi(\ominus_1 x)=\ominus_2\phi(x)$,
where $e_1,e_2$ are the gyro-identities of $G_1,G_2$.
\end{proposition}

The Einstein, M\"{o}bius and Lorentz gyrovector spaces are all real inner-product gyrovector spaces, with closed-form operations summarized in \cref{app:klein-geom}.

\subsection{Three Hyperbolic Models}
\label{app:hyp-models}

\subsubsection{Riemannian Operators}
\label{app:isometries}

The Klein closed-form Riemannian operators in this appendix (distance, exponential and logarithmic maps, parallel transport) follow~\citet{Chen2025gyrobn}, who derive them under arbitrary $K<0$ via the Klein--Poincar\'{e} isometry. We restate them here for completeness and consistent notation with our Klein layer derivations.

There are three standard models of $n$-dimensional hyperbolic space with sectional curvature $K<0$.
\begin{itemize}
\item \textbf{Klein model.} $\KKn=\{x\in\RR^n\mid\|x\|^2<-1/K\}$ with metric $g_x^{\KK}$ defined in \cref{eq:klein-metric}.
\item \textbf{Poincar\'{e} ball.} $\PP_K^n=\{x\in\RR^n\mid\|x\|^2<-1/K\}$ with conformal metric $g_x^{\PP}=(\lambda_x^K)^2\langle\cdot,\cdot\rangle$ where $\lambda_x^K=2/(1+K\|x\|^2)$.
\item \textbf{Lorentz model (hyperboloid).} $\LL_K^n=\{x=(x_t,x_s)\in\RR^{1+n}\mid \langle x,x\rangle_{\LL}=1/K,\;x_t>0\}$ with Lorentz inner product $\langle x,y\rangle_{\LL}=-x_t y_t+\langle x_s,y_s\rangle$.
\end{itemize}
\cref{tab:riem-ops-3model} summarizes the manifold, metric, distance, exponential and logarithmic maps, and parallel transport on the three models side-by-side.

\begin{table}[!ht]
\centering
\caption{Riemannian operators on the three hyperbolic models
($K<0$).}
\label{tab:riem-ops-3model}
\renewcommand{\arraystretch}{1.7}
\setlength{\tabcolsep}{5pt}
\resizebox{\linewidth}{!}{%
\begin{tabular}{lccc}
\toprule
\textbf{Operator}
 & \textbf{Klein} $\KKn$~\citep{Chen2025gyrobn}
 & \textbf{Poincar\'{e}} $\PP_K^n$~\citep{Ganea2018}
 & \textbf{Lorentz} $\LL_K^n$~\citep{Nickel2018} \\
\midrule
Manifold
 & $\{x\in\RR^n\mid\|x\|^2<-1/K\}$
 & $\{x\in\RR^n\mid\|x\|^2<-1/K\}$
 & $\{x\in\RR^{1+n}\mid\langle x,x\rangle_{\LL}=1/K,\,x_t>0\}$
\\
$g_x(u,v)$
 & $\dfrac{\langle u,v\rangle}{1+K\|x\|^2}-\dfrac{K\langle x,u\rangle\langle x,v\rangle}{(1+K\|x\|^2)^2}$
 & $(\lambda_x^K)^2\langle u,v\rangle$
 & $\langle u_s,v_s\rangle-u_t v_t$
\\
$d(x,y)$
 & $\dfrac{1}{\sqrt{-K}}\cosh^{-1}\!\!\left(\dfrac{1+K\langle x,y\rangle}{\sqrt{(1+K\|x\|^2)(1+K\|y\|^2)}}\right)$
 & $\dfrac{2}{\sqrt{-K}}\tanh^{-1}\!\bigl(\sqrt{-K}\|{-}x\oM y\|\bigr)$
 & $\dfrac{1}{\sqrt{-K}}\cosh^{-1}\!\bigl(K\langle x,y\rangle_{\LL}\bigr)$
\\
$\Expop_x v$
 & $x+\dfrac{\sinh(\|v\|_x)\hat v}{\cosh(\|v\|_x)+\gamma_x^2\langle x,\hat v\rangle\sinh(\|v\|_x)}$
 & $x\oM\!\left(\tanh\!\left(\dfrac{\sqrt{-K}\lambda_x^K\|v\|}{2}\right)\dfrac{v}{\sqrt{-K}\|v\|}\right)$
 & $\cosh(\alpha)x+\dfrac{\sinh(\alpha)}{\alpha}v,\;\alpha=\sqrt{-K}\|v\|_{\LL}$
\\
$\Logop_x y$
 & $\PT^{\KK}_{\mathbf{0}\to x}\!\bigl(\Logop_{\mathbf{0}}^{\KK}((-x)\oE y)\bigr)$
 & $\dfrac{2}{\sqrt{-K}\lambda_x^K}\tanh^{-1}\!\bigl(\sqrt{-K}\|{-}x\oM y\|\bigr)\dfrac{-x\oM y}{\|{-}x\oM y\|}$
 & $\dfrac{\cosh^{-1}(\beta)}{\sqrt{\beta^2-1}}(y-\beta x),\,\beta=K\langle x,y\rangle_{\LL}$
\\
$\PT_{x\to y}(v)$
 & $d\phi_{\PP\to\KK}\circ\PT^{\PP}_{\phi(x)\to\phi(y)}\circ d\phi_{\KK\to\PP}(v)$
 & $\dfrac{\lambda_x^K}{\lambda_y^K}\mathrm{gyr}[y,-x]v$
 & $v-\dfrac{K\langle y,v\rangle_{\LL}}{1+K\langle x,y\rangle_{\LL}}(x+y)$
\\
\bottomrule
\end{tabular}}
\end{table}

The three models are connected by Riemannian isometries given in \cref{eq:klein-poincare-map,eq:klein-lorentz-map}.
\begin{align}
  \phi_{\KK\to\PP}(x) &= \frac{x}{1+\sqrt{1+K\|x\|^2}},
  &\phi_{\PP\to\KK}(y) &= \frac{2y}{1-K\|y\|^2},
  \label{eq:klein-poincare-map}\\[4pt]
  \phi_{\KK\to\LL}(x) &= \Bigl(\frac{\gamma_x}{\sqrt{-K}},\;
    \gamma_x x\Bigr),
  &\phi_{\LL\to\KK}(x_t,x_s) &= \frac{x_s}{\sqrt{-K}\,x_t},
  \label{eq:klein-lorentz-map}
\end{align}
where $\gamma_x=1/\sqrt{1+K\|x\|^2}$ is the Lorentz factor. These isometries preserve the geodesic distance and induce correspondences between the gyrovector operations, namely Einstein $\oE$ for Klein, M\"{o}bius $\oM$ for Poincar\'{e}, and Lorentz $\oplus_{\LL}$ for the hyperboloid. See \cref{app:klein-geom}.

\textbf{Notation.}
For $x\in\KKn$ we denote the Lorentz factor by $\gamma_x=1/\sqrt{1+K\|x\|^2}$. For $x\in\PP_K^n$ we denote the conformal factor by $\lambda_x^K=2/(1+K\|x\|^2)$. For $x\in\LL_K^n$ we write $x=(x_t,x_s)$ with $x_t\in\RR$ the time coordinate and $x_s\in\RR^n$ the spatial component. Throughout we assume $K<0$, so $\sqrt{-K}\in(0,\infty)$.

\textbf{Klein closed-form distance.}
The Klein geodesic distance admits the closed-form $\arccosh$ expression
\begin{equation}
d^{\KK}(x,y)
 = \frac{1}{\sqrt{-K}}\arccosh\!\left(
  \frac{1+K\langle x,y\rangle}
       {\sqrt{(1+K\|x\|^2)(1+K\|y\|^2)}}\right),
\label{eq:klein-dist-full}
\end{equation}
which matches the geodesic distance on $\PP_K^n$ and $\LL_K^n$ under the isometries~\eqref{eq:klein-poincare-map}--\eqref{eq:klein-lorentz-map}.

\textbf{Klein exp/log maps at a general point.}\label{app:klein-explog-general}
For $p\in\KKn$ and $v\in T_p\KKn$ with Klein norm $\|v\|_p=\|v\|_{\KK}$, the exponential map is~\citep{Mao2024}
\begin{equation}
\Expop_p^{\KK}(v) = p + \frac{
  \sinh(\|v\|_p)\,\hat v
}{
  \cosh(\|v\|_p) + \gamma_p^2\langle p,\hat v\rangle\sinh(\|v\|_p)
},
\qquad \hat v=v/\|v\|_p.
\label{eq:exp-general}
\end{equation}
Since Klein geodesics are straight line segments, $\Expop_p^{\KK}(v)$ is a point on the chord through $p$ in direction $v$, with a nonlinear reparameterization that accounts for the varying Klein metric.

\textbf{Klein parallel transport.}\label{app:klein-pt}
Parallel transport from the origin to $p\in\KKn$ admits the closed form
\begin{equation}
\PT_{\mathbf{0}\to p}^{\KK}(v)
 = \sqrt{1+K\|p\|^2}\left(v+\frac{K\langle p,v\rangle}{1+\sqrt{1+K\|p\|^2}}\,p\right),
\label{eq:pt-0-to-p}
\end{equation}
given for $K=-1$ by \citet{Mao2024}. It scales the component of $v$ along $p$ by $1+K\|p\|^2$ and the component orthogonal to $p$ by $\sqrt{1+K\|p\|^2}$, which preserves the Klein norm.
At the origin the Klein metric reduces to the Euclidean inner product, so $T_{\mathbf{0}}\KKn\cong\RR^n$. The general $\PT^{\KK}_{x\to y}$ is obtained either from \cref{eq:pt-0-to-p} via left-gyrotranslation or from a Klein--Poincar\'{e} conjugation. See \cref{tab:riem-ops-3model}.

\begin{proposition}[Gyrovector--Riemannian equivalence
{\citep{Mao2024,Ungar2022}}]
\label[proposition]{prop:gyro-riemannian}
The Einstein operations and Riemannian operations on $\KKn$ are
related by
\begin{align}
  t\otE x
    &= \Expop_{\mathbf{0}}^{\KK}\!\left(t\cdot\Logop_{\mathbf{0}}^{\KK}(x)\right),
  \label{eq:scalar-riemannian}\\[2pt]
  \PT_{\mathbf{0}\to p}^{\KK}(v)
    &= \Logop_p^{\KK}\!\left(p\oE\Expop_{\mathbf{0}}^{\KK}(v)\right).
  \label{eq:pt-riemannian}
\end{align}
\end{proposition}
These identities let us implement Klein layers using either gyrovector or Riemannian operations interchangeably.

\subsubsection{Gyrovector Spaces}
\label{app:klein-geom}

\textbf{Klein space and the Einstein gyrovector space.}
The Klein space carries the \emph{Einstein gyrovector space} $(\KKn,\oE,\otE)$~\citep[Ch.~6]{Ungar2022}, with $\oE$ and $\otE$ defined in \crefrange{eq:einstein-add}{eq:einstein-scalar} of the main paper. It satisfies the four gyrogroup axioms (G1--G4) of \cref{def:gyrogroup}, where the gyro-identity is $\mathbf{0}$ (so $\mathbf{0}\oE x=x$), the gyro-inverse is the Euclidean negation $\ominus_{\mathrm{E}}x=-x$ (so $(-p)\oE x$ is the centered version of $x$ relative to $p$), and the Einstein gyration $\mathrm{gyr}_{\mathrm{E}}[x,y]$ takes a closed polynomial form in $\gamma_x,\gamma_y$. Einstein addition obeys the polynomial $\gamma$-identity $\gamma_{x\oE y}=\gamma_x\gamma_y(1-K\langle x,y\rangle)$ and recovers Euclidean addition as $K\to 0^-$. When $K=-1$ and $n=3$, $\oE$ coincides with the relativistic velocity composition law of special relativity, with $\mathrm{gyr}_{\mathrm{E}}[x,y]$ realizing the Thomas precession~\citep{Ungar2022}.

\textbf{Poincar\'{e} ball and the M\"{o}bius gyrovector space.}
The Poincar\'{e} ball carries the \emph{M\"{o}bius gyrovector space} $(\PP_K^n,\oM,\otM)$~\citep[Ch.~6]{Ungar2022} with gyroaddition
\begin{equation}
  x\oM y = \frac{
    (1-2K\langle x,y\rangle-K\|y\|^2)x + (1+K\|x\|^2)y
  }{
    1-2K\langle x,y\rangle+K^2\|x\|^2\|y\|^2
  },
  \label{eq:mobius-add}
\end{equation}
and scalar gyromultiplication coinciding with $\otE$ in coordinates, $t\otM x=\frac{1}{\sqrt{-K}}\tanh\!\bigl(t\arctanh(\sqrt{-K}\|x\|)\bigr)\frac{x}{\|x\|}$. The M\"{o}bius gyration formula is given in \cref{tab:riem-ops-3model}. The Klein and Poincar\'{e} gyrovector spaces are isomorphic via $\phi_{\KK\to\PP}$ in \cref{eq:klein-poincare-map}, so identities transfer between the two.

\textbf{Lorentz gyrovector space.}
The Lorentz hyperboloid carries the \emph{Lorentz gyrovector space} $(\LL_K^n,\oplus_{\LL},\otimes_{\LL})$, derived in~\citet{Chen2025gyrobn}, with gyro-identity $\bar{\mathbf{0}}=(1/\sqrt{-K},\mathbf{0})$ and gyro-inverse $\ominus_{\LL}(x_t,x_s)=(x_t,-x_s)$. The closed-form Lorentz gyroaddition and scalar gyromultiplication are piecewise expressions involving the time and spatial components, given in~\citet{Chen2025gyrobn} Props.~24--25. We omit the explicit formulas as the Lorentz gyrovector structure is not used by our Klein layers.

\textbf{Einstein $\oE$ vs.\ M\"{o}bius $\oM$ structural comparison.}\label{app:einstein-mobius} Although the Klein and Poincar\'{e} gyrovector spaces are isomorphic, their addition formulas differ in three ways that make Klein preferable for deep-learning pipelines.
\begin{itemize}[leftmargin=1.5em, itemsep=1pt]
\item \textbf{Linearity in the second argument.} The numerator of $x\oE y$ is linear in $y$, while the numerator of $x\oM y$ is quadratic in $y$ and couples $\|y\|^2$ with $x$. The denominator of $\oE$ is also linear in $\langle x,y\rangle$, whereas $\oM$'s denominator is quartic, containing $\|x\|^2\|y\|^2$. This reduces both the FLOP count and the depth of the autograd graph at every addition site.

\item \textbf{Polynomial $\gamma$-composition law.} Klein $\gamma$-values compose polynomially under Einstein addition, $\gamma_{x\oE y}=\gamma_x\gamma_y(1-K\langle x,y\rangle)$, whereas the M\"{o}bius counterpart $\gamma_{x\oM y}=\gamma_x\gamma_y\sqrt{1-2K\langle x,y\rangle+K^2\|x\|^2\|y\|^2}$ adds a square root on top of the inputs $\gamma_x,\gamma_y$. The two formulas share the cost of computing $\gamma_x$ itself (one square root per point), but the Klein composition keeps the autograd graph polynomial at every subsequent Busemann or distance evaluation, whereas the M\"{o}bius composition adds one $\sqrt{\cdot}$ node per addition site.

\item \textbf{Straight geodesics (Beltrami--Klein theorem).} Geodesics of $\KKn$ are Euclidean line segments in coordinates, so Klein geodesic hyperplanes coincide with the Euclidean hyperplanes cut by the open ball. Poincar\'{e} geodesics are circular arcs, so a geodesic MLR on $\PP_K^n$ requires a M\"{o}bius action to bend each learned hyperplane.
\end{itemize}
By contrast, the \emph{scalar multiplication} formulas $t\otE x$ and $t\otM x$ coincide in coordinates because scalar gyromultiplication is intrinsically a rescaling along the unique geodesic through the origin, and that geodesic shares the same $\tanh$ form in both coordinates. The same identity holds for $\Expop_{\mathbf{0}}$ and $\Logop_{\mathbf{0}}$. As a consequence, every Klein layer that depends only on origin-based exp/log or scalar multiplication (Klein FC, MLR, BMLR, BFC) inherits its derivation directly from the corresponding M\"{o}bius construction by left-gyrotranslation. The gyrogroup-aggregation layer (Klein GyroBN, which sums points under $\oE$) uses the closed-form Einstein midpoint, which is the M\"{o}bius gyromidpoint and the Lorentzian centroid written in Klein coordinates~\citep[Ch.~6]{Ungar2022}. \citet{Gulcehre2019} already use this midpoint in closed form in hyperbolic attention, and the midpoint batch normalization of Poincar\'{e} ResNet~\citep{vanSpengler2023} and the Lorentz batch normalization of HCNN~\citep{Bdeir2024} also compute their centre in closed form, so a closed-form centre is not specific to the Klein model.

\phantomsection\label{app:intrinsic}Following \citet{Chen2026,Shi2026}, the quantities of the manifold that define an intrinsic layer include the geodesic distance to a hyperplane and the Busemann function, as opposed to a Euclidean operation in a tangent space or in the ambient space. The layers of \cref{sec:klein-nn} are intrinsic in this sense, and their closed forms are specific to the Klein model.

Each Klein layer is the closed form of a prior construction. The MLR is the signed distance to a hyperplane~\citep{Ganea2018,Shimizu2021} (\cref{thm:klein-hyp,thm:klein-mlr}), the BMLR the Busemann logit~\citep{Chen2026} (\cref{thm:klein-busemann}), the FC and BFC the fully connected layer written as signed distances~\citep{Shimizu2021,Chen2026} (\cref{thm:klein-fc}), and GyroBN the gyrogroup batch normalization of \citet{Chen2025gyrobn}, with the Einstein midpoint as its centre (\cref{sec:klein-bn}).

\textbf{Summary of Klein operations.}\label{app:klein-summary}
\cref{tab:klein-ops} summarizes the Klein operations used in the neural network layers of \cref{sec:klein-nn}. Every operation has a closed-form expression and reduces to its Euclidean counterpart as $K\to 0^-$.

\begin{table}[!htbp]
\centering
\caption{Klein operations and their roles in neural network layers.}
\label{tab:klein-ops}
\renewcommand{\arraystretch}{1.3}
\setlength{\tabcolsep}{14pt}
\begin{tabular}{lll}
\toprule
\textbf{Operation} & \textbf{Formula} & \textbf{Used in} \\
\midrule
Einstein addition $\oE$
  & \cref{eq:einstein-add}
  & BN shift \\
Einstein scalar mult $\otE$
  & \cref{eq:einstein-scalar}
  & BN rescaling \\
$\Expop_{\mathbf{0}}^{\KK}$ / $\Logop_{\mathbf{0}}^{\KK}$
  & Special case of \cref{eq:exp-general} at $p{=}\mathbf{0}$
  & FC tangent lift \\
Geodesic distance
  & \cref{eq:klein-dist-full}
  & MLR, FC \\
Busemann function
  & \cref{eq:klein-busemann}
  & BMLR, BFC \\
Einstein midpoint
  & \cref{eq:einstein-midpoint}
  & BN centering \\
\bottomrule
\end{tabular}
\end{table}

\section{Experimental Details}
\label{app:experiments}

\textbf{Common protocols.}
\label{app:common}

As we use trivialization tricks in our MLR, FC, and GyroBN layers, all parameters in Klein neural networks lie in Euclidean space. The optimizer is selected per experiment to match the baseline protocol; see each experiment's setup for details.
All experiments are conducted on NVIDIA A100 GPUs.
\label{app:hardware}

\textbf{Interpretation of cross-model comparisons.}
\label{app:comparison-principle}

The Poincar\'{e}, Lorentz, and Klein models are isometric realizations of the same hyperbolic space, so coordinate-equivalent constructions have the same representational capacity. We therefore interpret the experiments along three separate axes: predictive performance under the statistical protocol below, computational cost, and the range of settings over which training remains numerically viable.

\textbf{Significance testing protocol.}
\label{app:significance}

For \cref{tab:image-cls,tab:link-pred,tab:nc}, we apply unpaired two-sample Welch
$t$-tests~\citep{Welch1947} to the reported mean and standard deviation over five seeds.
We control multiplicity with the Holm step-down procedure~\citep{Holm1979}, treating the
following two questions as separate test families.

\textbf{Comparison with the column best.} In each column, the numerically best result
is shown in \first{bold}, and results that are not significantly worse under the
Holm-adjusted Welch test ($p{\geq}0.05$) are \second{underlined}. All remaining results are
significantly below the column best.

\textbf{Comparison with the strongest baseline.} Each Klein layer is compared with the
strongest baseline in the same column. Asterisks indicate a significant positive difference
after Holm adjustment: $^{*}p{<}0.05$, $^{**}p{<}0.01$, and $^{***}p{<}0.001$.

\newpage
\subsection{Numerical Stability}
\label{app:stability-exp}
\paragraph{Localization of the non-finite runs.}
We rerun each configuration of \cref{tab:stab-curv,tab:stab-clip} that becomes
non-finite with a probe that records the first operation whose inputs are finite
and whose output is not, together with the inputs and parameters of the
classifier head at that step, and recompute the head term by term in FP32 and in
FP64. In all reproduced failures, three of the Lorentz head and six of the
Poincar\'{e} head, the failing operation lies in the head. In the Lorentz MLR of \citet{Bdeir2024} it is
$\beta_k=\sqrt{\cosh^2(\sqrt{-K}a_k)\|z_k\|^2-\sinh^2(\sqrt{-K}a_k)\|z_k\|^2}$,
which equals $\|z_k\|$ but rounds to $0$ once $\sqrt{-K}|a_k|$ reaches $8.4$ to
$9.0$, so that the logit becomes $0\cdot\infty$. In the Poincar\'{e}
MLR~\citep{Shimizu2021} it is $u^2$ inside
$\operatorname{arsinh}(u)=\log(u+\sqrt{1+u^2})$, which overflows in five of the
six runs once $2\sqrt{-K}|r_k|$ reaches $40$ to $45$, and in the sixth it is
$\cosh(2\sqrt{-K}r_k)$ itself, at $2\sqrt{-K}|r_k|=99$.

\paragraph{Rewriting the failing step.}
To test whether the failing step is the whole cause, we retrain each failing
configuration with that step rewritten, $\beta_k$ computed as $\|z_k\|$ and
$\operatorname{arsinh}$ evaluated by \texttt{torch.asinh}, or with the head in
FP64, three seeds each. For Lorentz at $K{=}{-}10$ either change removes every
failure, with $70.69$ and $70.46$ against $70.62$ for Klein in the same batch.
Without clipping the Lorentz runs stay finite once the features are bounded at
the projection radius of the Poincar\'{e} implementation, $6.21/\sqrt{-K}$ in
geodesic distance, with $72.36\pm0.32$ against $72.82\pm0.19$ for Klein. The
Poincar\'{e} head still fails at $r\ge5$ with either change, and at
$K{=}{-}0.01$ the FP64 head stays finite but does not learn ($2.8\%$).
Recording its parameters at every step at $K{=}{-}0.3$ and $r{=}5$ shows that
the offset $2\sqrt{-K}|r_k|$ grows from $0$ to $41$ within ten steps and that
the logits overflow after $11$ steps in FP32 and after $59$ in FP64. Swapping
the feature radius and the initialization between the two heads for $400$
steps, three seeds each, locates this growth in the Poincar\'{e} head. With the
geodesic radius $10$ and mean weight norm $1.83$ of the reference Poincar\'{e}
setup, the Poincar\'{e} head becomes non-finite in three of three runs within
$21$ steps, while the Klein head stays finite with logits below $30$. With the
Klein radius and initialization, $5$ and $0.58$, the Poincar\'{e} head trains,
and matching only one of the two leaves its logits in the thousands or
non-finite. On Tiny-ImageNet the same rewrites give $64.00\pm0.26$ for Lorentz
at $r{=}10$ and leave Poincar\'{e} non-finite in every run at geodesic radius
$5$ and above.

\paragraph{Failure points predicted by FP32.}
FP32 represents magnitudes up to $3.4\times10^{38}\approx e^{88.7}$ with a
24-bit significand. Each head contains a term that grows like $e^{s}$, where $s$
is the geodesic distance of the hyperplane from the origin in units of
$1/\sqrt{-K}$, namely $a=\sqrt{-K}|a_k|$ for the Lorentz head,
$x=2\sqrt{-K}|r_k|$ for the Poincar\'{e} head and $\rho=\sqrt{-K}|r_k|$ for the
Klein head. The two terms under the square root in $\beta_k$ differ by
$\|z_k\|^2$ and round to the same value once $\cosh^2 a$ reaches $2^{22}$ to
$2^{24}$, that is $a\approx8.3$ to $9.0$. The Poincar\'{e} head squares a term
of size about $e^{x}/2$, which overflows at $x\le45$. The Klein logit
$\gamma_x(\cosh\rho\,t-\sinh\rho)$ with $|t|<1$ overflows only once
$\gamma_x e^{\rho}$ exceeds $e^{88.7}$, at $\rho\approx87$ for $\gamma_x\le12$, or
at $\rho\approx72$ when FP32 rounds a feature onto the boundary and $\gamma_x$
is clamped at $3.2\times10^{7}$. No Klein run of the two sweeps becomes
non-finite.

\paragraph{The limit of the Klein coordinates.}
A point at geodesic distance $d$ from the origin has Klein norm
$\tanh(\sqrt{-K}d)/\sqrt{-K}$ and Poincar\'{e} norm
$\tanh(\sqrt{-K}d/2)/\sqrt{-K}$, so following \citet{Mishne2023} FP32 rounds it
onto the Klein boundary once $\sqrt{-K}d\gtrsim8.7$, half the radius
$17.3$ of the Poincar\'{e} ball. In a clipping sweep at $K{=}{-}1$, $r\in\{1,2,5,10,10^{9}\}$
and three seeds, one of the $15$ Klein runs becomes non-finite, at $r{=}10$,
where $60$ to $72\%$ of the features lie on the boundary in FP32 and the
argument of the Klein logit overflows at $\rho=73$ to $107$. Projecting the
features back into the ball, $\|x\|\le1/\sqrt{-K}-10^{-5}$, after
$\Expop_{\mathbf{0}}$ removes these failures, with no non-finite value in nine
runs at $(K,r)\in\{(-1,10),(-1,10^{9}),(-0.3,10^{9})\}$ and accuracies of
$71.02\pm0.49$, $71.41\pm1.18$ and $72.59\pm0.17$.

\label{app:numerical-extra}
\vspace{-0.6em}

The failures of the reference implementations in the stability sweeps lie in one
FP32 step of each classifier head, a cancellation in the Lorentz MLR and an overflow
in the Poincar\'{e} MLR (\cref{app:stability-exp}). The diagnostics below examine
the coordinate maps and operators on their own.

This appendix replicates the stability sweeps of
\cref{sec:exp-stability} on CIFAR-10 and Tiny-ImageNet, and
reports four operator-level diagnostics. The operator-level
tests use $256$ unit-norm tangent vectors at the origin in
$\mathbb{R}^{64}$ with $K{=}{-}1$. Each manifold uses its
standard reference implementation as in
\cref{sec:exp-stability}.

\textbf{CIFAR-100 clipping radius at every geodesic radius.} \cref{tab:stab-clip-full} lists the clipping sweep of \cref{tab:stab-clip} at every geodesic radius. Since Poincar\'{e} features lie at $\rho{=}2r$, we also run Poincar\'{e} at $r{=}0.5$ and $2.5$, so that every model has an entry at $\rho\in\{1,2,5,10,\infty\}$. Poincar\'{e} trains at $\rho{=}1$ and becomes non-finite in every seed from $\rho{=}5$.

\begin{table}[!htbp]
\centering
\caption{CIFAR-100 clipping sweep at $K{=}{-}0.3$ at every geodesic radius $\rho$ (3 seeds). Poincar\'{e} clips at $r{=}\rho/2$ and the other models at $r{=}\rho$.}
\label{tab:stab-clip-full}
\footnotesize
\begin{tabular}{@{}c|>{\columncolor{blue!8}}ccc@{}}
\toprule
$\rho$ & \textbf{Klein} & \textbf{Poincar\'{e}} & \textbf{Lorentz} \\
\midrule
$1$      & $73.41\pm0.33$ & $73.20\pm0.17$ & $72.84\pm0.13$ \\
$2$      & $74.54\pm0.13$ & $73.47\pm0.19$ & $74.68\pm0.56$ \\
$5$      & $74.16\pm0.03$ & \textcolor{red}{\textbf{NaN}} & $73.88\pm0.13$ \\
$10$     & $73.51\pm0.12$ & \textcolor{red}{\textbf{NaN}} & $72.44\pm0.21$ \\
$\infty$ & $71.69\pm0.15$ & \textcolor{red}{\textbf{NaN}} & \textcolor{red}{\textbf{NaN}} \\
\bottomrule
\end{tabular}
\end{table}

\textbf{CIFAR-10 clipping radius.} We rerun the
clipping-radius sweep of \cref{sec:exp-stability} on CIFAR-10
under the same $100$-epoch protocol at fixed $K{=}{-}0.3$.
\cref{tab:stab-clip-c10} reports CIFAR-10 top-1 accuracy.
Klein and Lorentz train at every radius, and Poincar\'{e} becomes non-finite at every $r{\ge}5$.

\begin{table}[!htbp]
\centering
\caption{CIFAR-10 clipping sweep (single seed). The Lorentz column is rerun at the labeled curvature.}
\label{tab:stab-clip-c10}
\setlength{\tabcolsep}{10pt}
\renewcommand{\arraystretch}{1.0}
\small
\begin{tabular}{@{}c|ccc@{}}
\toprule
$r$ & \textbf{Klein} & \textbf{Poincar\'{e}} & \textbf{Lorentz} \\
\midrule
$0.1$    & 57.37 & 93.65 & 66.63 \\
$0.5$    & 94.58 & 94.66 & 94.29 \\
$1.0$    & 94.72 & 94.55 & 94.67 \\
$2.0$    & 94.59 & 94.53 & 94.89 \\
$5.0$    & 94.70 & \textcolor{red}{\textbf{NaN}} & 94.81 \\
$10.0$   & 94.24 & \textcolor{red}{\textbf{NaN}} & 94.66 \\
$10^{9}$ & 93.97 & \textcolor{red}{\textbf{NaN}} & 94.28 \\
\bottomrule
\end{tabular}
\end{table}

\textbf{CIFAR-10 curvature.} The curvature sweep at fixed
$r{=}1.0$ on CIFAR-10 in \cref{tab:stab-curv-c10} reproduces
the CIFAR-100 picture, with Klein staying inside a $0.4$-point band
across all five curvatures, while Poincar\'{e} becomes non-finite at
$|K|{=}10$ and Lorentz trains at every curvature. The accuracy
ceiling on CIFAR-10 is high enough that the within-band
variation is below seed noise.

\begin{table}[!htbp]
\centering
\caption{CIFAR-10 curvature sweep (single seed). The Lorentz column is rerun at the labeled curvature.}
\label{tab:stab-curv-c10}
\setlength{\tabcolsep}{10pt}
\renewcommand{\arraystretch}{1.0}
\small
\begin{tabular}{@{}c|ccc@{}}
\toprule
$|K|$ & \textbf{Klein} & \textbf{Poincar\'{e}} & \textbf{Lorentz} \\
\midrule
$0.01$ & 94.59 & 94.74 & 94.59 \\
$0.1$  & 94.72 & 94.82 & 94.50 \\
$0.3$  & 94.69 & 94.71 & 94.57 \\
$1.0$  & 94.65 & 94.42 & 94.76 \\
$10.0$ & 94.46 & \textcolor{red}{\textbf{NaN}} & 94.25 \\
\bottomrule
\end{tabular}
\end{table}

\textbf{Tiny-ImageNet clipping radius.} \cref{tab:stab-clip-tin} repeats the clipping sweep on Tiny-ImageNet, where chance is $0.5\%$. The Klein entries at $r{\ge}5$ are rerun with five seeds on the full training set, since the runs first reported there iterated over only 31 to 116 of the 782 training batches per epoch. Klein learns at every radius from $r{=}1$ on and drops at $r{\le}0.5$. Without clipping one of the five seeds becomes non-finite in its first epoch, and projecting the features back into the ball after $\exp_{\mathbf{0}}$ removes it ($63.74\pm0.23$). Poincar\'{e} becomes non-finite at every $r{\ge}5$. The Lorentz column was run at $K{=}{-}1/0.3$ instead of $K{=}{-}0.3$ and is not comparable with the other two.

\begin{table}[!htbp]
\centering
\caption{Tiny-ImageNet clipping sweep (top-1 \%). Single seed, except Klein at $r{\ge}5$, five seeds on the full training set (at $r{=}10^{9}$ the four that stay finite). -- marks a run that used only part of the training set.}
\label{tab:stab-clip-tin}
\setlength{\tabcolsep}{10pt}
\renewcommand{\arraystretch}{1.0}
\small
\begin{tabular}{@{}c|ccc@{}}
\toprule
$r$ & \textbf{Klein} & \textbf{Poincar\'{e}} & \textbf{Lorentz} ($K{=}{-}1/0.3$) \\
\midrule
$0.1$    & 5.98 & -- & 6.64 \\
$0.5$    & 21.69 & -- & 36.92 \\
$1.0$    & 61.01 & 65.43 & 64.49 \\
$2.0$    & 64.87 & 65.65 & 65.30 \\
$5.0$    & $65.50\pm0.15$ & \textcolor{red}{\textbf{NaN}} & \textcolor{red}{\textbf{NaN}} \\
$10.0$   & $64.09\pm0.33$ & \textcolor{red}{\textbf{NaN}} & \textcolor{red}{\textbf{NaN}} \\
$10^{9}$ & $63.32\pm0.56$ & \textcolor{red}{\textbf{NaN}} & \textcolor{red}{\textbf{NaN}} \\
\bottomrule
\end{tabular}
\end{table}

\textbf{Boundary saturation.}
We compute $x=\exp_0(r\hat v)$ for $r\in\{1,5,10,50,100,500,1000\}$
and report the fraction of samples within $10^{-6}$ of the
manifold's projection boundary in \cref{tab:num-saturation}.
All three manifolds remain finite thanks to their built-in
projection. Klein activates its $1{-}10^{-5}$ clamp at
$r{=}10$, Poincar\'{e}'s more conservative $1{-}4\times10^{-3}$
threshold never triggers in this range, and the Lorentz time
component stays below $10^{6}$.

\begin{table}[!htbp]
\centering
\caption{Boundary saturation.}
\label{tab:num-saturation}
\setlength{\tabcolsep}{4pt}
\renewcommand{\arraystretch}{1.05}
\small
\begin{tabular}{@{}l|ccccccc@{}}
\toprule
\textbf{Saturation rate (\%)} & $r{=}1$ & $5$ & $10$ & $50$ & $100$ & $500$ & $1000$ \\
\midrule
Klein        & 0 & 0 & 100 & 100 & 100 & 100 & 100 \\
Poincar\'{e} & 0 & 0 & 0   & 0   & 0   & 0   & 0   \\
Lorentz      & 0 & 0 & 0   & 0   & 0   & 0   & 0   \\
\bottomrule
\end{tabular}
\end{table}

\textbf{Round-trip error.}
For tangent vectors of norm $\|v\|\in\{1,2,5,10,20,50\}$ we
report the mean error $\|\log_0(\exp_0(v))-v\|$ in
\cref{tab:num-roundtrip}. Klein occupies a middle ground:
at $\|v\|{=}5$ in FP32 it is four orders of magnitude more
accurate than Poincar\'{e} but three orders worse than Lorentz.
Switching to FP64 widens Lorentz's lead but barely helps
Poincar\'{e}, indicating that Poincar\'{e}'s FP32 error is
algorithmic precision loss rather than a representation issue.

\begin{table}[!htbp]
\centering
\caption{Log/exp round-trip error.}
\label{tab:num-roundtrip}
\setlength{\tabcolsep}{4pt}
\renewcommand{\arraystretch}{1.05}
\small
\resizebox{\textwidth}{!}{%
\begin{tabular}{@{}l|cccccc|cccccc@{}}
\toprule
& \multicolumn{6}{c|}{\textbf{FP32}} & \multicolumn{6}{c}{\textbf{FP64}} \\
\textbf{Manifold} & $\|v\|{=}1$ & $2$ & $5$ & $10$ & $20$ & $50$ & $\|v\|{=}1$ & $2$ & $5$ & $10$ & $20$ & $50$ \\
\midrule
Klein   & 4.2e\textminus 8 & 6.6e\textminus 7 & 2.1e\textminus 4 & 3.9 & 13.9 & 43.9 & 1.7e\textminus 16 & 1.7e\textminus 15 & 5.1e\textminus 13 & 3.9 & 13.9 & 43.9 \\
Poincar\'{e} & 2.0e\textminus 7 & 1.6e\textminus 6 & 1.9 & 6.9 & 16.9 & 46.9 & 1.7e\textminus 7 & 9.3e\textminus 7 & 3.3e\textminus 4 & 3.9 & 13.9 & 43.9 \\
Lorentz & \first{8.9e\textminus 8} & \first{1.0e\textminus 7} & \first{3.0e\textminus 7} & \first{5.9e\textminus 7} & \first{10.0} & \first{40.0} & \first{1.3e\textminus 16} & \first{1.8e\textminus 16} & \first{5.7e\textminus 16} & \first{1.3e\textminus 15} & \first{10.0} & \first{40.0} \\
\bottomrule
\end{tabular}}
\end{table}

\textbf{Composition drift.}
We iterate $y_k=\exp_0(\log_0(y_{k-1}))$ with
$y_0=\exp_0(v)$, $\|v\|{=}3$, and report the relative drift
$\|y_k-y_0\|/\|y_0\|$ in \cref{tab:num-drift}. Klein and
Lorentz are effectively idempotent under composition: their
FP64 drift stays near machine epsilon through $k{=}100$.
Poincar\'{e} drifts linearly in $k$, reaching $6.1\times 10^{-6}$
at $k{=}100$ in FP64. Klein and Lorentz are therefore
preferable when the manifold operator is restacked across
many layers.

\begin{table}[!htbp]
\centering
\caption{Composition drift.}
\label{tab:num-drift}
\setlength{\tabcolsep}{4pt}
\renewcommand{\arraystretch}{1.05}
\small
\begin{tabular}{@{}l|ccccc|ccccc@{}}
\toprule
& \multicolumn{5}{c|}{\textbf{FP32}} & \multicolumn{5}{c}{\textbf{FP64}} \\
\textbf{Manifold} & $k{=}1$ & $5$ & $10$ & $50$ & $100$ & $k{=}1$ & $5$ & $10$ & $50$ & $100$ \\
\midrule
Klein        & 5.8e\textminus 8 & 6.7e\textminus 8 & 6.6e\textminus 8 & 6.6e\textminus 8 & 6.6e\textminus 8 & \first{1.0e\textminus 16} & \first{1.4e\textminus 16} & \first{1.4e\textminus 16} & \first{1.4e\textminus 16} & \first{1.4e\textminus 16} \\
Poincar\'{e} & 7.4e\textminus 8 & 2.8e\textminus 7 & 5.4e\textminus 7 & 2.3e\textminus 6 & 4.3e\textminus 6 & 6.1e\textminus 8 & 3.1e\textminus 7 & 6.1e\textminus 7 & 3.1e\textminus 6 & 6.1e\textminus 6 \\
Lorentz      & 1.1e\textminus 7 & 1.4e\textminus 7 & 1.5e\textminus 7 & 1.5e\textminus 7 & 1.5e\textminus 7 & 1.6e\textminus 16 & 2.4e\textminus 16 & 2.4e\textminus 16 & 2.4e\textminus 16 & 2.4e\textminus 16 \\
\bottomrule
\end{tabular}
\end{table}

\textbf{FP32 vs.\ FP64 training.} We re-train each
manifold's classification head at the default configuration
in FP32 and FP64. As shown in \cref{tab:stress-dtype}, the
accuracy changes by at most $0.33$ points between the two precisions.
Away from the default radius, where \cref{tab:stab-curv,tab:stab-clip}
record the non-finite runs, FP64 removes the failures of the Lorentz head but not
those of the Poincar\'{e} head (\cref{app:stability-exp}).

\begin{table}[!htbp]
\centering
\caption{FP32 vs.\ FP64.}
\label{tab:stress-dtype}
\setlength{\tabcolsep}{8pt}
\renewcommand{\arraystretch}{1.0}
\small
\begin{tabular}{@{}l|cc@{}}
\toprule
\textbf{Manifold} & FP32 & FP64 \\
\midrule
Klein        & 77.11 & 77.10 \\
Poincar\'{e} & 76.70 & 76.79 \\
Lorentz      & 76.85 & 77.18 \\
\bottomrule
\end{tabular}
\end{table}

\textbf{Caveats and reproducibility.}
Each manifold uses its standard projection threshold (Klein
$1{-}10^{-5}$, Poincar\'{e} $1{-}4\times10^{-3}$, Lorentz no
boundary cap), so the NaN events are intrinsic to these
implementations as deployed in the literature. The
$K{=}{-}0.3$ used in the clipping sweeps follows the
image-classification setup of \citet{Chen2026PVNN}, and the curvature
sweeps span three orders of magnitude under the same
pattern, ruling out single-$K$ bias. Cells are single-seed (seed~$42$) unless the caption states the number of
seeds.

\subsection{Image Classification}
\label{app:image-exp}

\subsubsection{Datasets}

CIFAR-10 and CIFAR-100~\citep{Krizhevsky2009} datasets contain 60K $32\times32$ colored images from 10 and 100 classes, respectively. We use the dataset split implemented in PyTorch, with 50K training and 10K testing images. Following \citet{Bdeir2024}, we use data augmentation that includes random cropping with padding of 4 pixels and random horizontal flipping.

\subsubsection{Implementation Details}

We implement the experiments using the official code of \citet{Bdeir2024}. All models share a ResNet-18 backbone followed by a hyperbolic MLR classifier. The output embedding is mapped to the target manifold via the exponential map at the identity $e$, that is, $\Expop_e(x)$, with $e=\mathbf{0}$ for the Klein and Poincar\'{e} models. The training hyperparameters follow the ResNet-18 protocol of~\citet{Bdeir2024} and are listed in \cref{tab:image-cls-hp}. Feature clipping at norm $r$ is applied before $\Expop^{\KK}_{\mathbf{0}}$ to prevent overflow near the Klein ball boundary. For Klein heads, the curvature $K$ and clipping radius $r$ are selected per dataset on the grid of~\citet{Chen2026} App.~E.1.2 with $K\in\{-1,-0.5,-0.3\}$ and $r\in\{1,2,4\}$. The Poincar\'{e} and Lorentz baselines retain their original \citet{Chen2026} settings selected on the same grid, so all methods receive equal hyperparameter budget. We report 5-seed mean$\pm$std on CIFAR-10/100. ImageNet-1K entries are single runs at seed 42.

\begin{table}[!htbp]
\centering
\caption{Training hyperparameters for the ResNet-18 image classification protocol.}
\label{tab:image-cls-hp}
\renewcommand{\arraystretch}{0.95}
\small
\begin{tabular}{@{}l c@{}}
\toprule
\textbf{Setting} & \textbf{Value} \\
\midrule
Optimizer    & SGD \\
Initial learning rate & $0.1$ \\
Momentum     & $0.9$ \\
Weight decay & $5\times 10^{-4}$ \\
Batch size   & $128$ \\
Epochs       & $200$ \\
LR decay     & factor $\gamma=0.2$ at epochs $60$, $120$, $160$ \\
Curvature    & $K{\in}\{-1,-0.5,-0.3\}$, selected per dataset for Klein heads \\
\bottomrule
\end{tabular}
\end{table}

\subsection{Fully Hyperbolic ResNet}
\label{app:resnet-exp}

\textbf{Training protocol.}
Klein-ResNet follows the CIFAR-100 configuration of \citet{Bdeir2024}, also used for our reproductions of ResNet-18 and HCNN, with $200$ epochs, batch size $128$, Riemannian SGD with Nesterov momentum $0.9$, learning rate $0.1$ decayed by a factor $0.2$ at epochs $60$, $120$ and $160$, weight decay $5\times10^{-4}$, random cropping and horizontal flipping, and embedding dimension $512$. As in \citet{Bdeir2024}, the reported accuracy is the best test accuracy during training. Klein-ResNet uses the convolution of \cref{eq:klein-conv}, the pointwise ReLU, the residual connection of \cref{eq:klein-res}, the Klein GyroBN with the Einstein midpoint and the Klein MLR.

\textbf{Architecture.}
The input lift $X_q=\nu\otS\sigma^{-1}(x_q)=\sigma^{-1}(\nu x_q)$ rescales the input lift $\sigma^{-1}(x_q)$ of HCNN, so that $\|\sigma(X_q)\|=\nu\|x_q\|$. A $3\times3$ Klein convolution, the Klein GyroBN $\mathcal{B}$ with the Einstein midpoint and ReLU act on the lifted image, four stages of two residual blocks follow, and the Einstein midpoint over all locations feeds the Klein MLR. A block computes $h=\ReLU(\mathcal{B}(\mathcal{C}_1x))$ and $y=\ReLU\bigl(\mathcal{B}(\mathcal{C}_2h)\oS s(x)\bigr)$, where $\mathcal{C}_1,\mathcal{C}_2$ are $3\times3$ Klein convolutions, $\mathcal{C}_1$ carries the stride, and the shortcut $s$ is the identity, or a $1\times1$ Klein convolution followed by $\mathcal{B}$ when the block changes the resolution or the width.

\textbf{Hyperparameters.}
\cref{tab:resnet-hparams} lists the hyperparameters of Klein-ResNet on CIFAR-100 and Tiny-ImageNet. The reported models are trained on the full training set with seeds $0$ to $4$.

\begin{table}[h]
\centering
\small
\caption{Hyperparameters of Klein-ResNet on CIFAR-100 and Tiny-ImageNet.}
\label{tab:resnet-hparams}
\begin{tabular}{lcc}
\toprule
Hyperparameter & CIFAR-100 & Tiny-ImageNet \\
\midrule
Epochs & $200$ & $200$ \\
Batch size & $128$ & $128$ \\
Learning rate & $0.1$ & $0.1$ \\
Learning-rate decay epochs & $60$, $120$, $160$ & $60$, $120$, $160$ \\
Learning-rate decay factor & $0.2$ & $0.2$ \\
Weight decay & $5\times10^{-4}$ & $5\times10^{-4}$ \\
Optimizer & \multicolumn{2}{c}{Riemannian SGD, Nesterov momentum $0.9$} \\
Floating-point precision & $32$ bit & $32$ bit \\
GPU & $1\times$ NVIDIA A100 & $1\times$ NVIDIA A100 \\
Curvature $K$ & $-0.3$ & $-0.3$ \\
Input scale $\nu$ & $0.125$ & $0.25$ \\
Embedding dimension & $512$ & $512$ \\
\bottomrule
\end{tabular}
\end{table}

\textbf{Baseline accuracies.}
\cref{tab:resnet} lists the accuracies reported in the original papers, so that each baseline is represented under its own implementation. In our pipeline and under the same rule, ResNet-18 reaches $77.81\pm0.35$ over ten seeds and HCNN $77.76\pm0.19$ over eight seeds, against $77.72$ and $78.07$ reported by \citet{Bdeir2024}. Poincar\'{e} ResNet with the added fourth stage reaches $75.90\pm0.11$ over two seeds against $76.60$. On Tiny-ImageNet HCNN reaches $62.11\pm0.23$ over five seeds against $65.71$ reported, and the official ILNN code becomes non-finite in its GyroBN layer in all five seeds after completing $64$ to $84$ of the $200$ epochs, with a best test accuracy of $37.50\pm1.46$ before that point.

\textbf{Significance.}
Following \cref{app:significance}, we apply Welch $t$-tests to the mean and standard deviation of Klein-ResNet over five seeds and of each baseline as reported, which covers five runs for ResNet-18, Poincar\'{e} ResNet and HCNN~\citep{Bdeir2024} and five seeds for ILNN~\citep{Shi2026}, and control multiplicity over the four comparisons with the Holm procedure. \cref{tab:sig-resnet} reports the tests. Against our own reproductions in the same pipeline, Klein-ResNet is $0.28$ points above ResNet-18 ($77.81\pm0.35$ over ten seeds, $p{=}0.070$) and $0.33$ points above HCNN ($77.76\pm0.19$ over eight seeds, $p{=}0.017$), with two-sided Welch $t$-tests and no multiplicity correction.

\begin{table}[h]
\centering
\small
\caption{Welch $t$-tests for \cref{tab:resnet}. $\Delta$ is Klein-ResNet minus the baseline, so a positive value favors Klein-ResNet. Holm correction over the four tests.}
\label{tab:sig-resnet}
\begin{tabular}{lcccc}
\toprule
Baseline & $n$ & $\Delta$ & $t$ & Holm $p$ \\
\midrule
ResNet-18 & 5 & $+0.37$ & $3.33$ & $0.034$ \\
Poincar\'{e} ResNet & 5 & $+1.49$ & $8.86$ & $<0.001$ \\
ILNN & 5 & $-0.32$ & $-2.36$ & $0.093$ \\
HCNN & 5 & $+0.02$ & $0.17$ & $0.87$ \\
\bottomrule
\end{tabular}
\end{table}

\textbf{Timing.}
Each model is timed in a separate process on one NVIDIA A100-SXM 64\,GB with the same data pipeline. It trains for four epochs under \texttt{torch.compile}~\citep{Ansel2024PyTorch2} with static shapes. $t_{\text{epoch}}$ is the median duration of epochs two to four, since the first contains compilation, and VRAM is the peak allocated memory over the same epochs. HCNN and ILNN use the HyperbolicCV implementation, while Poincar\'{e} ResNet uses its official code with the Riemannian SGD of its paper. The official Poincar\'{e} ResNet has three stages, and we add a fourth stage with $512$ channels and stride $2$ to match ResNet-18, which gives $11.2$M parameters. Under \texttt{torch.compile} the ILNN loss diverges, which does not change the measured computation. On CIFAR-100, Klein-ResNet is timed at $K=-0.3$ and $\nu=0.5$ in a later session on the same hardware, in which the tangent variant takes $12.1$\,s per epoch, close to its $12.3$\,s in the session of the baselines. On Tiny-ImageNet, the baselines are timed together in one session and Klein-ResNet at $K=-0.3$ and $\nu=0.25$ in a separate session on the same hardware.

\textbf{Cost by component.}
We time single layers and whole networks in separate processes on the same GPU to trace the timing of \cref{tab:resnet} to individual components. \cref{tab:spd-fc,tab:spd-conv,tab:spd-bn,tab:spd-act} report one fully connected layer, one $3\times3$ convolution, one batch normalization and one ReLU of each network under \texttt{torch.compile} from small to large width, and \cref{tab:spd-eager} repeats one width of each layer in eager mode. Each layer receives random points at curvature $K=-1$ with gradients enabled, and the time covers the forward and the backward pass. \cref{tab:spd-swap} replaces the batch-normalization shift and the convolution of HCNN one at a time inside the whole ResNet-18, with Klein-ResNet at $K=-0.3$ and $\nu=0.125$, and \cref{tab:spd-batch} reports the largest batch size that fits in memory. Trained on CIFAR-100 with five seeds, the HCNN with both replacements becomes non-finite in every seed between epochs 10 and 12, and trains to the end once its features are bounded at the radius Klein-ResNet uses. HCNN uses its official implementation in HyperbolicCV, ILNN our implementation in the same codebase and Poincar\'{e} ResNet its official code, as in the timing above, except that the batch-normalization entries of ILNN in \cref{tab:spd-bn,tab:spd-eager} come from the official ILNN code, and \cref{tab:kres-cost} gives the operation counts of the convolutions.

Each single-layer time is the mean $\pm$ standard deviation over $5$ independent jobs, each the median of $30$ steps, and peak memory is the mean over the same jobs. In \cref{tab:spd-bn}, Poincar\'{e} ResNet uses its midpoint batch normalization, ILNN the batch normalization of its official code, HCNN its Lorentz batch normalization with the shift stored as a \texttt{ManifoldParameter} or, in the second HCNN row, as a plain parameter, and Klein-ResNet the Klein GyroBN with the Einstein midpoint; the ILNN row and the plain-shift HCNN row come from a separate set of $5$ jobs with the same script, in which Klein takes $2.02$\,ms and $578$\,MB at $C{=}512$. In \cref{tab:spd-act}, ILNN uses the Lorentz ReLU code of HCNN and Poincar\'{e} ResNet applies \texttt{nn.ReLU} to tangent vectors at the origin. In \cref{tab:spd-swap}, one training step is the forward pass, the cross-entropy and the backward pass at batch size $128$ on random inputs of the given image size, without the optimizer, and its time is the mean $\pm$ standard deviation over $5$ jobs; Euclidean $\beta$ makes the BN shift an ordinary parameter instead of a \texttt{ManifoldParameter}, and unfold-free conv replaces the convolution by the tangent-space convolution, which applies a standard convolution without unfolding patches. \cref{tab:spd-batch} tests batch sizes $32, 64, \dots, 1024$ in a single run.

\begin{table}[tbp]
\centering
\caption{Forward$+$backward time and extra peak memory of one FC layer on $4096$ random points with input and output dimension $d$, under \texttt{torch.compile} on one A100. Bold is best among the hyperbolic networks.}
\label{tab:spd-fc}
\footnotesize
\setlength{\tabcolsep}{4pt}
\adjustbox{max width=\linewidth}{%
\begin{tabular}{lrrrr@{\hspace{1.2em}}rrrr}
\toprule
 & \multicolumn{4}{c}{Time (ms) $\downarrow$} & \multicolumn{4}{c}{Peak memory (MB) $\downarrow$} \\
\cmidrule(lr){2-5} \cmidrule(lr){6-9}
Method & $d{=}16$ & $d{=}64$ & $d{=}256$ & $d{=}1024$ & $d{=}16$ & $d{=}64$ & $d{=}256$ & $d{=}1024$ \\
\midrule
Euclidean & $0.47\pm0.01$ & $0.48\pm0.01$ & $0.56\pm0.01$ & $1.56\pm0.08$ & 1.3 & 2.0 & 8.3 & 36.1 \\
\midrule
Poincar\'{e} & $3.28\pm0.05$ & $3.34\pm0.18$ & $3.49\pm0.04$ & $4.38\pm0.12$ & 2.4 & 9.1 & 36.2 & 145 \\
ILNN & $2.51\pm0.03$ & $2.53\pm0.02$ & $2.63\pm0.03$ & $3.40\pm0.05$ & 3.9 & 18.2 & 68.5 & 279 \\
HCNN & $\bm{0.60\pm0.01}$ & $\bm{0.56\pm0.01}$ & $\bm{0.56\pm0.01}$ & $\bm{1.70\pm0.01}$ & \textbf{1.3} & \textbf{3.8} & \textbf{12.1} & \textbf{49.0} \\
\rowcolor{blue!8}Klein & $0.76\pm0.01$ & $0.78\pm0.01$ & $0.88\pm0.01$ & $1.83\pm0.01$ & 1.6 & 6.0 & 24.1 & 96.1 \\
\bottomrule
\end{tabular}}
\end{table}

\begin{table}[tbp]
\centering
\caption{Forward$+$backward time and extra peak memory of one $3{\times}3$ convolution with $C$ input and output channels on $128$ random $16{\times}16$ feature maps, as in \cref{tab:spd-fc}. Bold is best among the hyperbolic networks.}
\label{tab:spd-conv}
\footnotesize
\setlength{\tabcolsep}{4pt}
\adjustbox{max width=\linewidth}{%
\begin{tabular}{lrrrr@{\hspace{1.2em}}rrrr}
\toprule
 & \multicolumn{4}{c}{Time (ms) $\downarrow$} & \multicolumn{4}{c}{Peak memory (MB) $\downarrow$} \\
\cmidrule(lr){2-5} \cmidrule(lr){6-9}
Method & $C{=}16$ & $C{=}64$ & $C{=}256$ & $C{=}512$ & $C{=}16$ & $C{=}64$ & $C{=}256$ & $C{=}512$ \\
\midrule
Euclidean & $0.59\pm0.00$ & $0.60\pm0.00$ & $1.72\pm0.01$ & $4.64\pm0.02$ & 8.8 & 34.1 & 197 & 430 \\
\midrule
Poincar\'{e} & $4.31\pm0.41$ & $6.50\pm0.39$ & $22.6\pm0.2$ & $53.5\pm0.2$ & 109 & 433 & 1731 & 3466 \\
ILNN & $0.85\pm0.01$ & $1.29\pm0.00$ & $8.91\pm0.01$ & $28.4\pm0.0$ & 39.0 & 155 & 619 & 1242 \\
HCNN & $\bm{0.74\pm0.01}$ & $1.39\pm0.01$ & $9.27\pm0.01$ & $29.4\pm0.0$ & 58.5 & 227 & 905 & 1807 \\
\rowcolor{blue!8}Klein & $1.03\pm0.01$ & $\bm{1.03\pm0.01}$ & $\bm{2.73\pm0.01}$ & $\bm{6.59\pm0.03}$ & \textbf{15.4} & \textbf{57.0} & \textbf{261} & \textbf{557} \\
\bottomrule
\end{tabular}}
\end{table}

\begin{table}[tbp]
\centering
\caption{Forward$+$backward time and extra peak memory of one batch normalization in training mode on the inputs of \cref{tab:spd-conv}, each network with its own layer (\cref{app:resnet-exp}). Bold is best among the hyperbolic networks.}
\label{tab:spd-bn}
\footnotesize
\setlength{\tabcolsep}{4pt}
\adjustbox{max width=\linewidth}{%
\begin{tabular}{lrrrr@{\hspace{1.2em}}rrrr}
\toprule
 & \multicolumn{4}{c}{Time (ms) $\downarrow$} & \multicolumn{4}{c}{Peak memory (MB) $\downarrow$} \\
\cmidrule(lr){2-5} \cmidrule(lr){6-9}
Method & $C{=}16$ & $C{=}64$ & $C{=}256$ & $C{=}512$ & $C{=}16$ & $C{=}64$ & $C{=}256$ & $C{=}512$ \\
\midrule
Euclidean & $0.53\pm0.02$ & $0.56\pm0.05$ & $0.50\pm0.01$ & $0.65\pm0.02$ & 4.0 & 16.0 & 64.0 & 128 \\
\midrule
Poincar\'{e} & $139\pm9$ & $141\pm13$ & $141\pm4$ & $153\pm10$ & 40.2 & 148 & 595 & 1158 \\
ILNN & $2.43\pm0.02$ & $2.46\pm0.03$ & $2.82\pm0.02$ & $4.35\pm0.06$ & 19.2 & 67.2 & \textbf{259} & \textbf{516} \\
HCNN & $6.48\pm0.24$ & $6.66\pm0.44$ & $6.80\pm0.76$ & $7.47\pm0.77$ & 22.9 & 83.0 & 323 & 643 \\
HCNN, plain shift & $1.77\pm0.02$ & $1.79\pm0.01$ & $2.06\pm0.06$ & $3.45\pm0.09$ & 23.0 & 83.1 & 323 & 644 \\
\rowcolor{blue!8}Klein & $\bm{1.13\pm0.05}$ & $\bm{1.15\pm0.02}$ & $\bm{1.42\pm0.05}$ & $\bm{2.01\pm0.01}$ & \textbf{11.6} & \textbf{41.6} & 290 & 578 \\
\bottomrule
\end{tabular}}
\end{table}

\begin{table}[tbp]
\centering
\caption{Forward$+$backward time and extra peak memory of one ReLU on the inputs of \cref{tab:spd-conv}, as in \cref{tab:spd-fc}. Bold is best among the hyperbolic networks.}
\label{tab:spd-act}
\footnotesize
\setlength{\tabcolsep}{4pt}
\adjustbox{max width=\linewidth}{%
\begin{tabular}{lrrrr@{\hspace{1.2em}}rrrr}
\toprule
 & \multicolumn{4}{c}{Time (ms) $\downarrow$} & \multicolumn{4}{c}{Peak memory (MB) $\downarrow$} \\
\cmidrule(lr){2-5} \cmidrule(lr){6-9}
Method & $C{=}16$ & $C{=}64$ & $C{=}256$ & $C{=}512$ & $C{=}16$ & $C{=}64$ & $C{=}256$ & $C{=}512$ \\
\midrule
Euclidean & $0.37\pm0.01$ & $0.36\pm0.01$ & $0.37\pm0.01$ & $0.44\pm0.00$ & 4.5 & 18.0 & 72.0 & 144 \\
\midrule
Poincar\'{e} & $0.37\pm0.01$ & $\bm{0.36\pm0.01}$ & $\bm{0.36\pm0.01}$ & $\bm{0.44\pm0.00}$ & 4.5 & 18.0 & 72.0 & 144 \\
ILNN & $0.40\pm0.01$ & $0.42\pm0.01$ & $0.47\pm0.02$ & $0.62\pm0.00$ & \textbf{4.4} & \textbf{16.5} & \textbf{64.5} & \textbf{129} \\
HCNN & $0.39\pm0.01$ & $0.43\pm0.04$ & $0.47\pm0.01$ & $0.64\pm0.03$ & \textbf{4.4} & \textbf{16.5} & \textbf{64.5} & \textbf{129} \\
\rowcolor{blue!8}Klein & $\bm{0.36\pm0.01}$ & $0.37\pm0.00$ & $\bm{0.36\pm0.01}$ & $\bm{0.44\pm0.00}$ & 4.5 & 18.0 & 72.0 & 144 \\
\bottomrule
\end{tabular}}
\end{table}

\begin{table}[tbp]
\centering
\caption{Forward$+$backward time (ms) of single layers in eager mode and under \texttt{torch.compile}, on the inputs of Tables~\ref{tab:spd-fc}--\ref{tab:spd-act}; the ratio row is eager over compiled time. Bold is best among the hyperbolic networks.}
\label{tab:spd-eager}
\footnotesize
\setlength{\tabcolsep}{4pt}
\adjustbox{max width=\linewidth}{%
\begin{tabular}{llrrrr>{\columncolor{blue!8}}r}
\toprule
Layer & Mode & Euclidean & Poincar\'{e} & ILNN & HCNN & Klein \\
\midrule
\multirow{3}{*}{FC ($d{=}256$)} & eager & $0.27\pm0.01$ & $4.60\pm0.29$ & $8.38\pm0.29$ & $\bm{0.58\pm0.01}$ & $1.79\pm0.02$ \\
 & compile & $0.56\pm0.01$ & $3.49\pm0.04$ & $2.63\pm0.03$ & $\bm{0.56\pm0.01}$ & $0.88\pm0.01$ \\
 & ratio & 0.5$\times$ & 1.3$\times$ & 3.2$\times$ & 1.0$\times$ & 2.0$\times$ \\
\midrule
\multirow{3}{*}{Conv ($C{=}256$)} & eager & $1.44\pm0.00$ & $29.6\pm0.1$ & $19.1\pm0.0$ & $14.1\pm0.0$ & $\bm{6.80\pm0.01}$ \\
 & compile & $1.72\pm0.01$ & $22.6\pm0.2$ & $8.91\pm0.01$ & $9.27\pm0.01$ & $\bm{2.73\pm0.01}$ \\
 & ratio & 0.8$\times$ & 1.3$\times$ & 2.1$\times$ & 1.5$\times$ & 2.5$\times$ \\
\midrule
\multirow{3}{*}{BN ($C{=}256$)} & eager & $0.38\pm0.00$ & $24.5\pm0.2$ & $12.0\pm0.3$ & $13.1\pm0.1$ & $\bm{9.29\pm0.01}$ \\
 & compile & $0.50\pm0.01$ & $141\pm4$ & $2.82\pm0.02$ & $6.80\pm0.76$ & $\bm{1.42\pm0.05}$ \\
 & ratio & 0.8$\times$ & 0.2$\times$ & 4.2$\times$ & 1.9$\times$ & 6.6$\times$ \\
\midrule
\multirow{3}{*}{Activation ($C{=}256$)} & eager & $0.16\pm0.00$ & $\bm{0.16\pm0.00}$ & $0.69\pm0.01$ & $0.69\pm0.00$ & $\bm{0.16\pm0.00}$ \\
 & compile & $0.37\pm0.01$ & $\bm{0.36\pm0.01}$ & $0.47\pm0.02$ & $0.47\pm0.01$ & $\bm{0.36\pm0.01}$ \\
 & ratio & 0.4$\times$ & 0.4$\times$ & 1.5$\times$ & 1.5$\times$ & 0.4$\times$ \\
\bottomrule
\end{tabular}}
\end{table}

\begin{table}[tbp]
\centering
\caption{Whole-network cost of one training step under \texttt{torch.compile} when HCNN components are replaced one at a time (\cref{app:resnet-exp}). Bold is best among the hyperbolic networks.}
\label{tab:spd-swap}
\footnotesize
\setlength{\tabcolsep}{4pt}
\adjustbox{max width=\linewidth}{%
\begin{tabular}{lccrrrrr}
\toprule
 & & & \multicolumn{3}{c}{CIFAR ($32{\times}32$)} & \multicolumn{2}{c}{Tiny-ImageNet ($64{\times}64$)} \\
\cmidrule(lr){4-6} \cmidrule(lr){7-8}
Network & Euclidean $\beta$ & Unfold-free conv & ms/step $\downarrow$ & GB $\downarrow$ & graph breaks $\downarrow$ & ms/step $\downarrow$ & GB $\downarrow$ \\
\midrule
Euclidean ResNet-18 &  &  & $11.6\pm0.0$ & 0.68 & 0 & $36.8\pm0.2$ & 2.42 \\
\midrule
HCNN &  &  & $158\pm3$ & 7.61 & 50 & $319\pm1$ & 30.23 \\
HCNN & \checkmark &  & $72.7\pm0.2$ & 6.92 & 2 & $257\pm0$ & 27.45 \\
HCNN &  & \checkmark & $149\pm6$ & 4.22 & 50 & $198\pm6$ & 16.54 \\
HCNN & \checkmark & \checkmark & $45.4\pm3.1$ & 3.42 & 2 & $\bm{119\pm0}$ & 13.46 \\
\midrule
\rowcolor{blue!8}Klein-ResNet &  &  & $\bm{41.3\pm1.9}$ & \textbf{2.68} & \textbf{1} & $122\pm0$ & \textbf{10.41} \\
\bottomrule
\end{tabular}}
\end{table}

\begin{table}[tbp]
\centering
\caption{Largest batch size that trains on one 64\,GB A100, with the peak memory in GB at that batch in parentheses. Bold is best among the hyperbolic networks.}
\label{tab:spd-batch}
\footnotesize
\adjustbox{max width=\linewidth}{%
\begin{tabular}{lrrrr}
\toprule
 & \multicolumn{2}{c}{\texttt{torch.compile}} & \multicolumn{2}{c}{eager} \\
\cmidrule(lr){2-3} \cmidrule(lr){4-5}
Method & CIFAR $\uparrow$ & Tiny-ImageNet $\uparrow$ & CIFAR $\uparrow$ & Tiny-ImageNet $\uparrow$ \\
\midrule
Euclidean & 1024 (4.7) & 1024 (18.4) & 1024 (4.7) & 1024 (18.4) \\
\midrule
Poincar\'{e} & 256 (28.9) & 128 (57.5) & 256 (32.6) & 64 (32.5) \\
ILNN & 512 (44.3) & 128 (44.1) & 256 (40.4) & 64 (40.4) \\
HCNN & \textbf{1024} (61.0) & 256 (60.4) & 512 (31.9) & 128 (31.9) \\
\rowcolor{blue!8}Klein-ResNet & \textbf{1024} (20.7) & \textbf{512} (41.3) & \textbf{1024} (53.7) & \textbf{256} (53.7) \\
\bottomrule
\end{tabular}}
\end{table}

\subsection{Graph Learning}
\label{app:graph-exp}

\subsubsection{Datasets}

\textbf{Disease}~represents a SIR disease transmission tree (2 classes, 1044 nodes, 1043 edges). \textbf{Airport}~is a transductive dataset of airline routes (4 classes, 3188 nodes, 18631 edges). \textbf{PubMed}~is a citation network in medicine (3 classes, 19717 nodes, 44338 edges). \textbf{Cora}~is a citation network in machine learning (7 classes, 2708 nodes, 5429 edges).

\subsubsection{Implementation Details}

We follow the link-prediction protocol of HGCN~\citep{Chami2019} (data splits, Fermi--Dirac decoder, training and evaluation schedule). Each encoder stacks two hyperbolic FC layers of width $128$ with no graph aggregation, followed by a Fermi--Dirac decoder. In \cref{tab:link-pred} every FC layer is followed by the batch normalization of its own model and then $\sigma=\mathrm{ReLU}$, with GyroBN~\citep{Chen2025gyrobn} for M\"{o}bius FC and BFC-P, the Lorentzian-centroid BN of~\citet{Bdeir2024} for Lorentz FC, PLFC and BFC-L, the GyroBN of PVNN~\citep{Chen2026PVNN} for PV FC and Klein GyroBN for Klein FC and Klein BFC, at curvature $K=-1$ and float32. Training stops at the first non-finite link score and keeps the best validation epoch before it. The boundary comparison of \cref{tab:geo-vs-horo} uses no batch normalization. For reference, \citet{Chen2025gyrobn} report HNN, whose layers are M\"{o}bius FC, at $75.1$, $90.8$, $94.9$ and $89.0$ on Disease, Airport, PubMed and Cora without batch normalization and at $81.2$, $95.4$, $95.8$ and $94.3$ with GyroBN.

M\"{o}bius FC applies $\Expop_{\mathbf{0}}\circ\sigma\circ\Logop_{\mathbf{0}}$ after the batch normalization as in HNN~\citep{Ganea2018}, Lorentz FC and PLFC apply $\sigma$ to the space coordinates after it as in HBNN~\citep{Chen2026}, and Klein FC, PV FC and the three BFC layers apply no activation inside the layer and the same $\Expop_{\mathbf{0}}\circ\sigma\circ\Logop_{\mathbf{0}}$ after the batch normalization. Without batch normalization (\cref{tab:geo-vs-horo}), Klein FC and Klein BFC apply $\sigma$ to the logit before the $\sinh$ projection onto the manifold, as in PVNN~\citep{Chen2026PVNN} and HBNN.

For the three layers on the Lorentz model, the decoder computes the distance as $d(x,y)=2\,\mathrm{asinh}\big(\sqrt{\langle x-y,x-y\rangle_{\LL}}/2\big)$, which equals the $\mathrm{arccosh}$ form of HGCN. That form clamps $-\langle x,y\rangle_{\LL}$ at $1+10^{-7}$, so in float32 every distance below about $5\times10^{-4}$ takes one value with zero gradient.

\begin{table}[!tbp]
\centering
\caption{Node classification accuracy (\%), 5 seeds. Top block as reported by \citet{Chen2026PVNN}; bottom block rerun by us with one protocol (\cref{app:klein-nc-tuning}). Bold is best, underlined is not significantly worse (\cref{app:significance}).}
\label{tab:nc}
\footnotesize
\renewcommand{\arraystretch}{0.7}
\setlength{\tabcolsep}{6pt}
\adjustbox{max width=\textwidth}{%
\begin{tabular}{c l cccc}
\toprule
\textbf{Space} & \textbf{Method}
 & \textbf{Disease} ($\delta{=}0$)
 & \textbf{Airport} ($\delta{=}1$)
 & \textbf{PubMed} ($\delta{=}3.5$)
 & \textbf{Cora} ($\delta{=}11$) \\
\midrule
$\PP_K^n$ & HNN++~\citep{Shimizu2021}
   & \first{80.57$\pm$0.23} & 88.40$\pm$0.17 & \first{73.68$\pm$0.39} & 52.06$\pm$0.90 \\
$\LL_K^n$ & LNN~\citep{Bdeir2024}
   & 79.90$\pm$0.01 & 75.20$\pm$1.08 & 68.82$\pm$0.88 & 53.34$\pm$1.65 \\
$\KKn$ & Klein HNN~\citep{Mao2024}
   & 79.41$\pm$0.55 & 92.10$\pm$0.97 & 69.36$\pm$0.76 & 52.26$\pm$1.99 \\
\midrule
$\LL_K^n$ & LNN++~\citep{Shi2026}
   & 79.90$\pm$0.00 & 90.00$\pm$0.97 & \second{69.44$\pm$5.48} & 46.62$\pm$3.37 \\
$\PV_K^n$ & PVNN~\citep{Chen2026PVNN}
   & 79.90$\pm$0.00 & \first{97.55$\pm$0.95} & \second{72.88$\pm$0.58} & 48.86$\pm$2.43 \\
\rowcolor{blue!8}
$\KKn$ & \textbf{Klein-HNN++}
   & 79.71$\pm$0.26 & \second{96.96$\pm$0.77} & 72.06$\pm$0.72 & \first{55.80$\pm$1.59} \\
\rowcolor{blue!8}
$\KKn$ & \textbf{Klein-HBNN++}
   & \second{79.43$\pm$1.07} & \second{96.68$\pm$1.09} & 71.50$\pm$1.04 & 51.06$\pm$1.01 \\
\bottomrule
\end{tabular}
}
\vspace{-1em}
\end{table}

All models are trained with Adam on an NVIDIA A100 GPU for up to $5000$ epochs, with early stopping on the validation ROC (patience $300$, at least $200$ epochs). Every layer uses one grid, learning rate $\{0.001, 0.0025, 0.005, 0.01, 0.015\}$ $\times$ dropout $\{0, 0.2, 0.4\}$ $\times$ weight decay $\{0, 10^{-3}\}$, selected on the mean validation ROC of seeds $100$ and $101$; the selected setting is tested with seeds $0$ to $4$, reporting the test ROC at the epoch of best validation ROC. \cref{tab:graph-hp-lp} lists the selected settings. The two smallest learning rates extend a first grid of $\{0.005, 0.01, 0.015\}$, on which every setting selected for the two Klein layers lay at the smallest learning rate. We added them for every method and repeated the selection over all $30$ settings, which moves $22$ of the $32$ method--dataset cells to a new learning rate, M\"{o}bius FC on PubMed and Cora among them. On the first grid, Klein FC reaches $81.42$, $95.74$, $95.75$ and $94.30$ and M\"{o}bius FC $83.09$, $95.85$, $96.80$ and $94.48$ on Disease, Airport, PubMed and Cora.

\begin{table}[!htbp]
\centering
\caption{Selected link-prediction settings (learning rate / dropout / weight decay) of the layers in \cref{tab:link-pred}, with the settings they share below.}
\label{tab:graph-hp-lp}
\renewcommand{\arraystretch}{0.95}
\small
\begin{tabular}{@{}l cccc@{}}
\toprule
\textbf{Method} & \textbf{Disease} & \textbf{Airport} & \textbf{PubMed} & \textbf{Cora} \\
\midrule
M\"{o}bius FC & $0.005/0/0$ & $0.005/0/10^{-3}$ & $0.001/0/0$ & $0.0025/0.2/10^{-3}$ \\
BFC-P & $0.001/0.2/10^{-3}$ & $0.001/0.2/10^{-3}$ & $0.001/0.2/10^{-3}$ & $0.001/0.2/10^{-3}$ \\
Lorentz FC & $0.0025/0/10^{-3}$ & $0.001/0/10^{-3}$ & $0.001/0.2/0$ & $0.01/0.4/0$ \\
PLFC & $0.01/0/0$ & $0.015/0/10^{-3}$ & $0.001/0/0$ & $0.015/0/10^{-3}$ \\
BFC-L & $0.0025/0.2/10^{-3}$ & $0.0025/0/10^{-3}$ & $0.005/0.4/10^{-3}$ & $0.01/0/10^{-3}$ \\
PV FC & $0.001/0.2/0$ & $0.005/0.2/0$ & $0.001/0.2/10^{-3}$ & $0.0025/0.2/0$ \\
Klein FC & $0.001/0.2/0$ & $0.0025/0/10^{-3}$ & $0.001/0/10^{-3}$ & $0.001/0/10^{-3}$ \\
Klein BFC & $0.005/0.2/10^{-3}$ & $0.001/0/10^{-3}$ & $0.0025/0.4/10^{-3}$ & $0.0025/0.2/10^{-3}$ \\
\midrule
Batch normalization & \multicolumn{4}{c}{that of the model of each layer} \\
Non-finite step & \multicolumn{4}{c}{stop, keep the best validation epoch} \\
Optimizer  & \multicolumn{4}{c}{Adam} \\
Curvature $K$ & \multicolumn{4}{c}{$-1$} \\
Hidden dimension & \multicolumn{4}{c}{$128$} \\
Number of FC layers & \multicolumn{4}{c}{$2$} \\
Maximum epochs & \multicolumn{4}{c}{$5000$} \\
Early stopping & \multicolumn{4}{c}{patience $300$, at least $200$ epochs} \\
Precision  & \multicolumn{4}{c}{float32} \\
Selection seeds & \multicolumn{4}{c}{$\{100, 101\}$} \\
Test seeds & \multicolumn{4}{c}{$\{0, 1, 2, 3, 4\}$} \\
\bottomrule
\end{tabular}
\end{table}

\begin{table}[!htbp]
\centering
\caption{Training time per epoch (ms) of the layers of \cref{tab:link-pred}, each with the batch normalization of its own model, mean of the last 5 of 100 epochs with one fixed setting (learning rate $0.01$, no dropout or weight decay), one run at a time on one A100. BFC-P becomes non-finite within the 100 epochs and stops early, so its time is not reported.}
\label{tab:lp-time}
\small
\begin{tabular}{@{}l cccc@{}}
\toprule
\textbf{Method} & \textbf{Disease} & \textbf{Airport} & \textbf{PubMed} & \textbf{Cora} \\
\midrule
M\"{o}bius FC & 53.5 & 68.3 & 128.5 & 47.9 \\
BFC-P & -- & -- & -- & -- \\
Lorentz FC & 29.0 & 58.3 & 105.5 & 34.2 \\
PLFC & 33.9 & 63.3 & 110.7 & 40.5 \\
BFC-L & 37.3 & 65.5 & 115.9 & 41.8 \\
PV FC & 930.3 & 952.4 & 1010.2 & 929.4 \\
Klein FC & 26.7 & 44.1 & 72.3 & 30.0 \\
Klein BFC & \textbf{26.0} & \textbf{43.5} & \textbf{72.1} & \textbf{29.0} \\
\bottomrule
\end{tabular}
\end{table}

\subsubsection{NC Backbone Architectures}
\label{app:nc-backbones}

All compared NC methods share the same template. Input features are lifted to the target manifold via $\Expop_{\mathbf{0}}$, passed through a 2-layer hyperbolic FC encoder (no graph aggregation), and classified by a hyperbolic MLR head. The methods differ in (i) the manifold and (ii) the FC and MLR formulation, summarized below.

\textbf{HNN++}~\citep{Shimizu2021} uses the Poincar\'{e} FC and Poincar\'{e} MLR on the Poincar\'{e} ball with curvature $-1$, with M\"{o}bius gyro-addition for biases and an in-layer sigmoid activation.

\textbf{LNN}~\citep{Bdeir2024} uses the Lorentz FC encoder and the Lorentz MLR classifier of~\citet{Bdeir2024} on the hyperboloid model, with tangent-space activation $\Logop_{\mathbf{0}}\!\to\!\sigma\!\to\!\Expop_{\mathbf{0}}$.

\textbf{LNN++}~\citep{Shi2026} replaces the LNN encoder with the point-to-hyperplane PLFC layer, which casts each output coordinate as the signed Lorentzian distance to a learnable hyperplane, and retains LMLR as the classifier. We use the official ILNN implementation, swapping only the encoder.

\textbf{PVNN}~\citep{Chen2026PVNN} uses the unbounded PV space $\PV_K^n=\RR^n$ with the proper-velocity metric, paired with the PV FC layer and the PV MLR head. We follow the authors' default hyperparameters.

\textbf{Klein HNN}~\citep{Mao2024} builds a Klein linear layer as the Einstein version $\Expop_{\mathbf{0}}\circ f\circ\Logop_{\mathbf{0}}$ of a Euclidean linear map and classifies with a Euclidean MLR. Its row in \cref{tab:nc} lists the values reported by \citet{Chen2026PVNN}. Trained in this pipeline back to back with Klein-HNN++ on one CPU node over five seeds, it takes $50.45$, $129.89$, $6.73$ and $14.10$\,ms per training epoch on Disease, Airport, PubMed and Cora, against $41.90$, $107.70$, $5.40$ and $11.61$\,ms for Klein-HNN++, a ratio of $0.80$ to $0.83$.

\textbf{Klein-HNN++} is our improved Klein NN, where the closed-form Klein FC encoder of \cref{thm:klein-fc} is stacked twice and paired with the Klein MLR head of \cref{thm:klein-mlr}.

\textbf{Klein-HBNN++} keeps the same pipeline but uses the horospheric Klein BFC encoder and the Klein BMLR head, with KleinHypAct activation between FC layers.

\subsubsection{PV FC and PLFC Initialization}
\label{app:pv-init}

For the PV FC layer~\citep{Chen2026PVNN} and the PLFC layer~\citep{Shi2026}, we replace the default LeCun initialization (std~$0.02$) with Xavier initialization for the linear weights. All other defaults are preserved.

\subsubsection{Klein NC Hyperparameters and Tuning Protocol}
\label{app:klein-nc-tuning}

We adopt the feature-only node-classification pipeline of \citet{Chen2026PVNN}, in which two FC layers are followed by an MLR head. Klein-HNN++ uses Klein FC with Klein MLR and Klein-HBNN++ uses Klein BFC with Klein BMLR. The HNN++, LNN and Klein HNN rows of \cref{tab:nc} are the values reported by \citet{Chen2026PVNN}. We run the bottom block of \cref{tab:nc}, PVNN, LNN++ (the official PLFC layer with the LMLR head), Klein-HNN++ and Klein-HBNN++, in this pipeline with one protocol. The test split is that of \citet{Chen2026PVNN}. On Disease and Airport, 20\% of the training nodes, stratified by class, form the validation split, and on Cora and PubMed the 500 standard Planetoid validation nodes do. Every run trains with \textsc{RiemannianAdam} for $2000$ epochs and keeps the checkpoint with the highest validation accuracy. For each method and dataset the learning rate in $\{0.005, 0.01, 0.05\}$, the curvature $c=-K$ in $\{0.1, 0.3, 0.5, 1, 1.5\}$ and the dropout rate in $\{0, 0.1, 0.2, 0.4\}$ are selected on the mean validation accuracy of seeds $100$ and $101$, and \cref{tab:nc} reports the test accuracy over seeds $0$ to $4$. \cref{tab:klein-nc-hp} lists the selected values.

\begin{table}[!htbp]
\centering
\caption{Selected hyperparameters (learning rate / curvature $c$ / dropout) of the methods rerun in \cref{tab:nc}, with the settings they share below.}
\label{tab:klein-nc-hp}
\renewcommand{\arraystretch}{0.95}
\small
\begin{tabular}{@{}l cccc@{}}
\toprule
\textbf{Method} & \textbf{Disease} & \textbf{Airport} & \textbf{PubMed} & \textbf{Cora} \\
\midrule
PVNN & $0.01/0.5/0$ & $0.005/0.1/0$ & $0.05/0.3/0$ & $0.01/0.3/0$ \\
LNN++ & $0.01/1.5/0.1$ & $0.01/0.5/0$ & $0.01/1.0/0$ & $0.005/0.5/0.1$ \\
Klein-HNN++ & $0.01/0.3/0.1$ & $0.005/1.0/0$ & $0.005/1.5/0$ & $0.01/0.1/0.2$ \\
Klein-HBNN++ & $0.05/0.3/0$ & $0.01/0.3/0$ & $0.05/0.1/0.4$ & $0.05/0.3/0.2$ \\
\midrule
Optimizer   & \multicolumn{4}{c}{\textsc{RiemannianAdam}} \\
Maximum epochs & \multicolumn{4}{c}{$2000$} \\
Batch size  & \multicolumn{4}{c}{$128$} \\
Hidden dimension & \multicolumn{4}{c}{$16$} \\
Weight decay & \multicolumn{4}{c}{$5\times 10^{-4}$} \\
Selection seeds & \multicolumn{4}{c}{$\{100, 101\}$} \\
Test seeds & \multicolumn{4}{c}{$\{0, 1, 2, 3, 4\}$} \\
\bottomrule
\end{tabular}
\end{table}

\subsection{Computational and Geometric Analysis}

\subsubsection{Decision Boundaries of Hyperbolic MLR Heads}
\label{app:boundary-geometry}

\begin{figure}[!htbp]
\centering
\includegraphics[width=\textwidth]{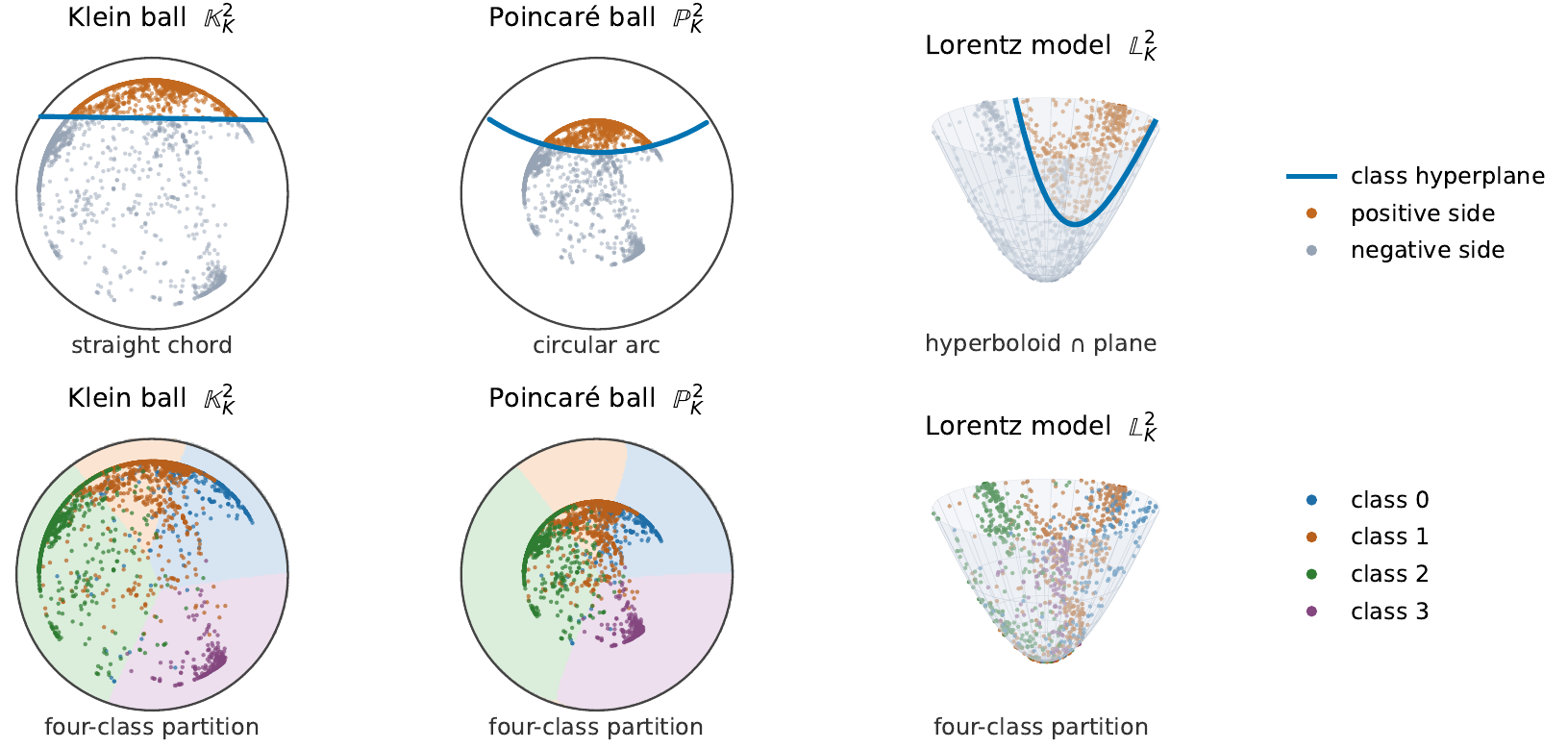}
\caption{A Klein MLR trained on Airport, seen in the three models ($K{=}{-}1$). The stack is that of \cref{sec:exp-nc} with the last layer narrowed to two for plotting, reaching $76.7\%$ on the $3188$ nodes. Top, one class hyperplane; bottom, the four-class partition.}
\label{fig:boundary-models}
\end{figure}

\cref{tab:mlr-comparison} lists what each head computes; \cref{fig:boundary-models} shows what its boundary is.
The head is the two-FC-plus-MLR Klein stack of \cref{sec:exp-nc} trained on Airport with the
last layer set to width two so it can be drawn, reaching $76.7\%$ test accuracy on the $3188$
nodes. The top row shows $H=\{x\mid\langle x,z\rangle=r\}$ for one class, the surface whose
signed distance is that class logit, with points coloured by the side they fall on. The
bottom row shows the partition the same head induces, every pixel classified in Klein
coordinates, so the two panels are one classifier seen through two coordinate systems and
agree on every point. We draw a class hyperplane rather than a boundary between classes
because the latter meet at the origin, where a hyperplane is a diameter in both balls; the
one drawn lies $0.555$ out, the furthest of the four.

Two independent properties of a boundary drive the cost of a head. It is \emph{totally
geodesic} when it is itself a hyperbolic subspace, which is what makes the signed distance to
it a geodesic distance; horospheres are not. Separately it may or may not be \emph{flat} in
the coordinates the head optimizes over, and only the Klein model gives both at once. The
measured bend of $H$ is exactly zero in $\KKn$, since Klein geodesics are Euclidean segments
(\cref{app:einstein-mobius}), against $0.232$ of the ball radius in $\PP_K^n$, where it is a
circular arc meeting the boundary orthogonally. In $\LL_K^n$ it is the intersection of the
hyperboloid with a linear hyperplane through the ambient origin, flat in $\RR^{1+n}$ at the
cost of carrying the time coordinate. The bottom row shows the same contrast at the level of
the partition, whose cells are bounded by straight chords in $\KKn$ and by arcs in $\PP_K^n$.

A curved boundary must be anchored, so each Poincar\'{e} geodesic head carries a
manifold-valued $p_k\in\PP_K^n$ per class, which is what \cref{tab:mlr-comparison} records as
lost compact parameters and an order of magnitude in FLOPs. The Klein heads need only the
Euclidean pair $(z,r)$ drawn here, so their per-class computation is one batched matrix
product, and the geodesic and horospherical Klein heads end up costing the same
(\cref{tab:boundary-cost}).

\subsubsection{Computational Cost of Geodesic and Horospherical Heads}
\label{app:boundary-cost}

Klein MLR and Klein BMLR have nearly identical FLOP counts and runtimes, indicating that the
choice between geodesic and horospherical boundaries introduces no material computational
difference within the Klein model (\cref{tab:boundary-cost}). The larger cost of competing heads
arises from their manifold-valued class parameters, which require per-class operations instead of
a single batched matrix product. Thus, runtime reflects both arithmetic complexity and the
batchability of the parameterization.

\begin{table}[!htbp]
\centering
\caption{Computational cost of geodesic and horospherical classification heads.}
\label{tab:boundary-cost}
\footnotesize
\setlength{\tabcolsep}{4pt}
\adjustbox{max width=\textwidth}{%
\begin{tabular}{llccc}
\toprule
Head & Boundary & Manifold-valued param & FLOPs/class & ms/iter, $n{=}50$ \\
\midrule
Poincar\'{e} MLR \citep{Ganea2018}      & geodesic hyperplane & $p_k\in\PP_K^n$ & $C(19n{+}27)$ & --- \\
Pseudo-Busemann MLR \citep{Nguyen2025}  & horosphere          & $p_k\in\PP_K^n$ & $C(18n{+}30)$ & $217.98$ \\
Klein MLR (\cref{sec:klein-mlr})        & geodesic hyperplane & none            & $C(2n{+}14)$  & $1.498$ \\
Klein BMLR (\cref{sec:klein-bmlr})      & horosphere          & none            & $C(2n{+}11)$  & $1.504$ \\
\bottomrule
\end{tabular}}
\end{table}

\subsubsection{Comparison of Geodesic and Horospherical Decoding}
\label{app:geo-vs-horo}

Klein FC and Klein BFC both apply the activation to the logit inside the layer and share every
other setting (two layers, a Fermi--Dirac decoder, no batch normalization, one
hyperparameter grid and the same seeds), so on the graphs they differ
only in the boundary type. Klein MLR and Klein BMLR on CIFAR-100 share backbone, schedule and
seeds (\cref{tab:geo-vs-horo}). The two boundaries are within one point of each other on every
dataset. The geodesic boundary is higher on PubMed and Cora with $p<0.05$, and the differences
on Disease, Airport and CIFAR-100 are not significant.

\begin{table}[!htbp]
\centering
\caption{Geodesic and horospherical boundaries in the Klein model, differing only in the boundary type.
Test ROC (graphs) and top-1 (CIFAR-100) over 5 seeds; $\Delta$ is horosphere minus geodesic, $p$ the Welch $p$.}
\label{tab:geo-vs-horo}
\footnotesize
\begin{tabular}{@{}lcccc@{}}
\toprule
Setting & Geodesic & Horosphere & $\Delta$ & $p$ \\
\midrule
Disease, $\delta{=}0$   & $73.69\pm4.85$ & $74.11\pm1.39$ & $+0.43$ & $0.87$ \\
Airport, $\delta{=}1$   & $95.18\pm0.33$ & $94.78\pm0.26$ & $-0.40$ & $0.09$ \\
PubMed, $\delta{=}3.5$  & $96.44\pm0.07$ & $96.19\pm0.13$ & $-0.25$ & $0.014$ \\
Cora, $\delta{=}11$     & $94.34\pm0.20$ & $93.41\pm0.35$ & $-0.93$ & $0.003$ \\
CIFAR-100               & $78.25\pm0.13$ & $78.24\pm0.13$ & $-0.01$ & -- \\
\bottomrule
\end{tabular}
\end{table}

\subsection{MLR Head Runtime}
\label{app:ablations}

We benchmark the hyperbolic MLR heads on random Euclidean features at batch size $1024$, $100$ classes and FP32, averaging a forward$+$backward pass over five runs after warm-up (\cref{app:image-exp}). \cref{tab:mlr-runtime} reports the mean per-iteration runtime. PBMLR-P is the slowest due to its high complexity and lack of batch efficiency, whereas the Klein variants are the fastest at every input dimension, by a small margin of $1.13$--$1.20\times$ over BMLR-L.

\begin{table}[!ht]
\centering
\caption{MLR head runtime (ms / iter).}
\label{tab:mlr-runtime}
\setlength{\tabcolsep}{6pt}
\renewcommand{\arraystretch}{1}
\setlength{\aboverulesep}{0pt}\setlength{\belowrulesep}{0pt}
\footnotesize
\begin{tabular}{l|cccc}
\toprule
\textbf{Method} & $n{=}50$ & $n{=}100$ & $n{=}200$ & $n{=}500$ \\
\midrule
Euclidean MLR & 0.263 & 0.273 & 0.265 & 0.269 \\
\midrule
PMLR     & 2.001 & 1.999 & 1.962 & 2.046 \\
PBMLR-P  & \textcolor{red!70!black}{217.98} & \textcolor{red!70!black}{217.86} & \textcolor{red!70!black}{222.30} & \textcolor{red!70!black}{221.86} \\
BMLR-P   & 1.940 & 1.957 & 1.932 & 2.041 \\
\midrule
LMLR     & 2.528 & 2.481 & 2.512 & 2.510 \\
BMLR-L   & 1.696 & 1.684 & 1.695 & 1.801 \\
\midrule
PV MLR   & 2.231 & 2.201 & 2.206 & 2.272 \\
\midrule
\rowcolor{blue!8}
\textbf{Klein MLR}  & \textcolor{green!60!black}{1.498} & 1.527 & 1.538 & \textcolor{green!60!black}{1.502} \\
\rowcolor{blue!8}
\textbf{Klein BMLR} & 1.504 & \textcolor{green!60!black}{1.486} & \textcolor{green!60!black}{1.504} & 1.551 \\
\bottomrule
\end{tabular}
\end{table}

\subsection{Statistical and Robustness Analysis}

\subsubsection{Significance Tests for Accuracy Comparisons}
\label{app:sig-tables}

\cref{tab:sig-cifar,tab:sig-graph} report the statistical tests underlying the markings in
\cref{tab:image-cls,tab:link-pred,tab:nc}, following the protocol in \cref{app:significance}. On each
image dataset, the better Klein head is compared with both the Euclidean baseline and the strongest
hyperbolic baseline; for graph tasks, each Klein model is compared with the strongest other method
of the column run with the same protocol. On the graphs, Klein-HNN++ is significantly above PVNN on
Cora node classification, Klein FC is significantly below M\"{o}bius FC only on PubMed link
prediction, Klein BFC is significantly below M\"{o}bius FC on Disease, PubMed and Cora, and no other
difference is significant, including the margins of Klein FC over the strongest layer not built in the
tangent space.

\begin{table}[!htbp]
\centering
\caption{CIFAR, 5 seeds per group. $\Delta$ is Klein minus the comparator, so a positive value
favors Klein. Holm correction over the four tests shown.}
\label{tab:sig-cifar}
\footnotesize
\begin{tabular}{@{}lccc@{}}
\toprule
Comparison & $\Delta$ & 95\% CI & Holm $p$ \\
\midrule
Klein MLR $-$ Euclidean, CIFAR-100 & $+0.53$ & $[+0.32,+0.74]$ & $0.0014$ \\
Klein BMLR $-$ Euclidean, CIFAR-10 & $+0.19$ & $[+0.04,+0.34]$ & $0.065$ \\
Klein MLR $-$ BMLR-P, CIFAR-100    & $+0.15$ & $[-0.28,+0.58]$ & $0.82$ \\
Klein BMLR $-$ BMLR-P, CIFAR-10    & $+0.01$ & $[-0.16,+0.18]$ & $0.89$ \\
\bottomrule
\end{tabular}
\end{table}

\begin{table}[!htbp]
\centering
\caption{Graphs, 5 seeds. Klein minus the strongest other method of each column under the same
protocol, M\"{o}bius FC for link prediction (LP), the strongest layer not built in the tangent space in the row
marked non-tangent, and PVNN for node classification (NC). Holm-adjusted over the table, $^{*}p{<}0.05$, $^{***}p{<}0.001$.}
\label{tab:sig-graph}
\footnotesize
\begin{tabular}{@{}llcccc@{}}
\toprule
Task & Klein layer & Disease ($\delta{=}0$) & Airport ($\delta{=}1$) &
PubMed ($\delta{=}3.5$) & Cora ($\delta{=}11$) \\
\midrule
LP & Klein FC                & $-0.43$         & $+0.01$      & $-0.38^{***}$ & $-0.31$ \\
LP & Klein BFC               & $-3.32^{*}$     & $-0.06$      & $-1.26^{*}$  & $-2.23^{***}$ \\
LP & Klein FC, non-tangent   & $+1.96$         & $+0.57$      & $+0.40$      & $+0.44$ \\
NC & Klein-HNN++             & $-0.19$         & $-0.59$      & $-0.82$      & $+6.94^{*}$ \\
NC & Klein-HBNN++            & $-0.47$         & $-0.87$      & $-1.38$      & $+2.20$ \\
\bottomrule
\end{tabular}
\end{table}

\subsubsection{Numerical Differences Across Realizations}
\label{app:ordering}

The three realizations are isometric and therefore have the same representational capacity, while
their numerical implementations remain coordinate dependent. \cref{tab:deviation-ordering} orders
the settings by their deviation from the default configuration. In every setting Klein and the
strongest baseline that trains differ by at most $0.28$ points, and no difference is significant. At
$|K|{=}0.01$ and $r{=}5$ the Poincar\'{e} head becomes non-finite, while Klein and Lorentz train.

\begin{table}[!htbp]
\centering
\caption{CIFAR-100 top-1 (\%), ordered by how far the setting is pushed from the default
$K{=}{-}1$, $r{=}1$. No margin here is significant.}
\label{tab:deviation-ordering}
\footnotesize
\begin{tabular}{@{}lccc@{}}
\toprule
Setting & Klein & Best baseline & Margin \\
\midrule
curvature flattened to $|K|{=}0.01$ & $72.29$ & $72.37$ & $-0.08$ \\
clip loosened to $r{=}5$            & $74.16$ & $73.88$ & $+0.28$ \\
each method at its own tuned $K$ and $r$ & $78.25$ & $78.10$ & $+0.15$ \\
default $K{=}{-}1$, $r{=}1$ for all three & $78.07$ & $78.26$ & $-0.19$ \\
\bottomrule
\end{tabular}
\end{table}

\subsubsection{Training Viability under Challenging Settings}
\label{app:capability}

\begin{table}[!htbp]
\centering
\caption{Settings at which a baseline returns no usable model, under the 200-epoch protocol of
the manuscript. ``NaN'' means the training loss became non-finite in the first epoch, ``collapse''
means the encoder settled at chance.}
\label{tab:capability}
\footnotesize
\begin{tabular}{@{}lccc@{}}
\toprule
Setting & Poincar\'{e} & Lorentz & Klein \\
\midrule
feature radius $r\ge5$                       & NaN & finite through $r{=}10$ & finite at every radius \\
curvature $|K|{=}0.01$                       & NaN & trains & $76.45$ \\
BFC encoder at depth 4                       & collapse on Airport & collapse on both & trains on both \\
$100$ composed $\Logop_{\mathbf{0}}\circ\Expop_{\mathbf{0}}$ & drifts to $6.1{\times}10^{-6}$ & machine epsilon & machine epsilon \\
batch center at every step                   & iterative Fr\'{e}chet & iterative Fr\'{e}chet & single $O(mn)$ pass \\
\bottomrule
\end{tabular}
\end{table}

\subsubsection{Effect of Encoder Depth}
\label{app:depth}

\cref{tab:depth} evaluates BFC encoders with depths from $2$ to $6$ on Disease and Airport over five
seeds. On Airport, Klein remains trainable through depth $6$, whereas Poincar\'{e} becomes unstable
at depth $2$ and both baselines collapse by depth $3$; consequently, Klein is the only realization
that remains trainable at depth $4$ on both datasets. On Disease, Poincar\'{e} remains trainable one
layer deeper than Klein, while Lorentz collapses from depth $3$; at depth $8$, all three
realizations collapse on both datasets.

\begin{table}[!htbp]
\centering
\caption{Link-prediction ROC-AUC (\%) against encoder depth, BFC encoder, 5 seeds.
``collapse'' means AUC at chance, ``collapse $k$/5'' that $k$ of the 5 seeds collapsed.}
\label{tab:depth}
\footnotesize
\begin{tabular}{@{}llccc@{}}
\toprule
Dataset & Depth & Klein & Poincar\'{e} & Lorentz \\
\midrule
Disease & 2 & $73.89\pm0.93$ & $72.46\pm2.08$ & $74.71\pm0.68$ \\
Disease & 3 & $78.00\pm0.92$ & $77.20\pm1.70$ & collapse \\
Disease & 4 & $77.65\pm1.31$ & $77.47\pm0.83$ & collapse \\
Disease & 5 & collapse 4/5   & $76.17\pm0.63$ & collapse \\
Disease & 6 & collapse       & collapse 4/5   & collapse \\
\midrule
Airport & 2 & $93.81\pm0.36$ & $60.47\pm14.94$ (collapse 3/5) & $93.82\pm0.24$ \\
Airport & 3 & $95.14\pm0.56$ & collapse 4/5 & collapse \\
Airport & 4 & $94.18\pm0.26$ & collapse & collapse \\
Airport & 5 & $91.47\pm2.74$ & collapse & collapse \\
Airport & 6 & $83.52\pm4.52$ & collapse & collapse \\
\bottomrule
\end{tabular}
\end{table}

\subsection{Learned Representations}
\label{app:cka}

\subsubsection{Class Geometry}
\label{app:class-geometry}

For each CIFAR-100 class, we compute the mean penultimate feature and measure the angles between all
$4950$ pairs of class directions. Because these angles are invariant to global rotation and
rescaling, they characterize the geometry of the learned representations independently of the
coordinate system. The angular spread differs by $0.10^{\circ}$ between the two Klein seeds and by
the same amount between Klein and Poincar\'{e}, while all models have mean angles near
$90^{\circ}$; Lorentz exhibits a larger spread of $7.73^{\circ}$.

\begin{table}[!htbp]
\centering
\caption{Class geometry on CIFAR-100. Nearest-direction agreement is the fraction of samples in
the outer half of the representation whose nearest class direction matches the ground-truth class;
chance agreement is $0.01$.}
\label{tab:class-geometry}
\footnotesize
\begin{tabular}{@{}lcccc@{}}
\toprule
Run & Mean angle & Spread of angles & Nearest-direction agreement & Accuracy \\
\midrule
Klein, seed 0     & $90.54^{\circ}$ & $4.52^{\circ}$ & $0.928$ & $78.07$ \\
Klein, seed 1     & $90.54^{\circ}$ & $4.62^{\circ}$ & $0.921$ & $78.11$ \\
Poincar\'{e}, seed 0 & $90.55^{\circ}$ & $4.42^{\circ}$ & $0.937$ & $78.26$ \\
Lorentz, seed 0   & $90.46^{\circ}$ & $7.73^{\circ}$ & $0.904$ & $77.24$ \\
\bottomrule
\end{tabular}
\end{table}

\subsubsection{Representation Comparison Across Isometric Formulations}

Under matched CIFAR-100 conditions ($K{=}{-}1$, $r{=}1$, and an identical training protocol),
we compare the penultimate representations of the Klein, Poincar\'{e}, and Lorentz MLR models
using linear CKA~\citep{Kornblith2019}. Same-model cross-seed CKA provides an empirical reference
for the variation induced by training stochasticity. This analysis tests whether the three
isometric formulations learn comparable representations in practice.

\begin{table}[h]
\centering
\caption{Representation similarity measured by linear CKA on CIFAR-100 penultimate features at
$K{=}{-}1$.}
\label{tab:cka}
\small
\begin{tabular}{@{}lcl@{}}
\toprule
Comparison & CKA & Interpretation \\
\midrule
Same-model cross-seed & $0.86$ / $0.86$ / $0.89$ & empirical cross-seed reference \\
Klein vs.\ Poincar\'{e} & $0.86$ & comparable to the cross-seed reference \\
Klein vs.\ Lorentz & $0.76$ & below the cross-seed reference \\
Poincar\'{e} vs.\ Lorentz & $0.75$ & below the cross-seed reference \\
\bottomrule
\end{tabular}
\end{table}

\begin{figure}[h]
\centering
\includegraphics[width=\textwidth]{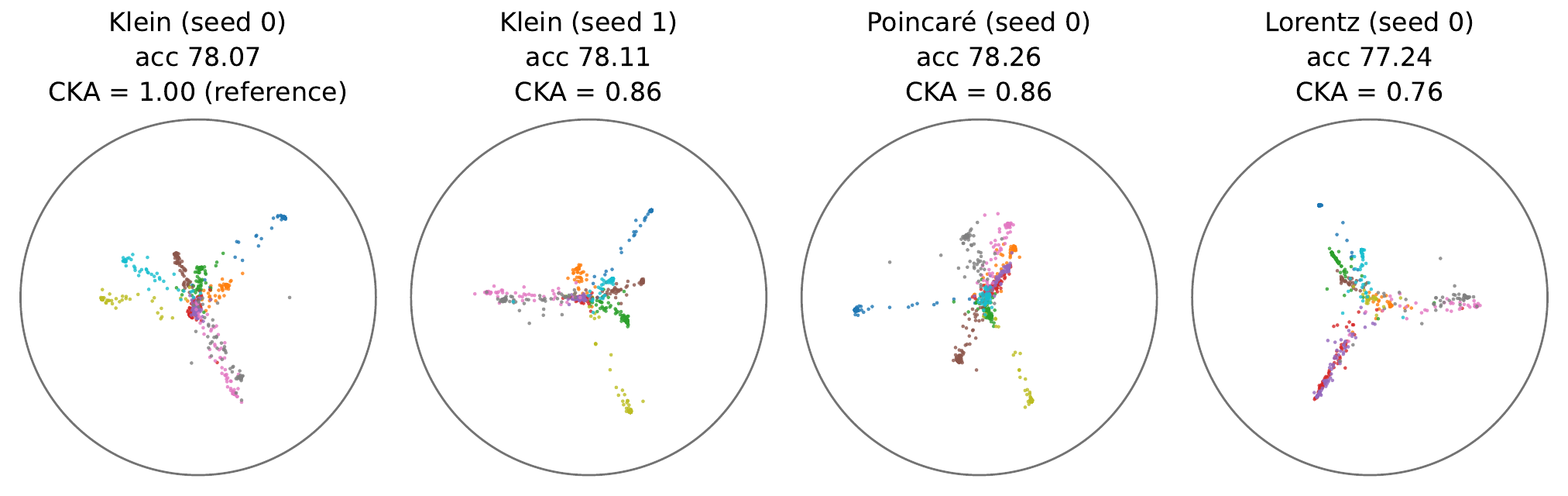}
\caption{PCA visualizations of the learned representations for the three formulations.}
\label{fig:cka-qualitative}
\end{figure}

\cref{fig:cka-qualitative} visualizes the clipped penultimate features of the first ten
CIFAR-100 classes at matched $K{=}{-}1$. Each panel is reduced independently by PCA and rendered
in a Poincar\'{e} disk; the visualization therefore supports a qualitative comparison of
within-panel class organization rather than direct pointwise alignment across formulations. All
three panels exhibit a dense central region with radially separated class clusters.

Klein and Poincar\'{e} attain a CKA of $0.86$, comparable to the same-model cross-seed reference.
Comparisons involving Lorentz yield lower CKA values of $0.75$--$0.76$. Together with their similar
classification accuracy, these results indicate substantial representational similarity across
the three formulations, while leaving room for coordinate-dependent differences in training
dynamics.

\section{Additional Details of Klein Neural Network Layers}
\label{app:klein-nn-details}

This appendix provides derivations, implementation details, and computational analyses for the five Klein layers introduced in \cref{sec:klein-nn}. The convolution and residual connection of \cref{sec:klein-resnet} are treated in \cref{app:klein-resnet}.

\textbf{Overview of the layer derivations.}
\label{app:layer-derivations}

The five Klein layers require closed-form expressions for their underlying geometric primitives in
Klein coordinates (\cref{tab:layer-derivations}). Klein MLR follows the signed-margin reformulation
of Euclidean softmax and uses the point-to-hyperplane distance established in
\cref{thm:klein-hyp}. Klein BMLR replaces the Euclidean inner product with the Klein Busemann
function derived in \cref{thm:klein-busemann}, yielding a horospherical classifier directly in
Klein coordinates.

\begin{table}[!htbp]
\centering
\caption{Closed-form geometric primitives used by the five Klein layers.}
\label{tab:layer-derivations}
\footnotesize
\begin{tabular}{@{}ll@{}}
\toprule
Layer & Required primitive \\
\midrule
Klein MLR, Klein FC   & closed-form signed distance from a point to a Klein hyperplane (\cref{thm:klein-hyp}) \\
Klein BMLR, Klein BFC & closed-form Klein Busemann function (\cref{thm:klein-busemann}) \\
Klein GyroBN          & closed-form Einstein midpoint for single-pass batch centering \\
\bottomrule
\end{tabular}
\end{table}

\subsection{Klein MLR and BMLR}

\textbf{Klein Busemann derivations.}
\label{app:busemann-verify}

\textbf{Derivation via Klein--Lorentz isometry.}
The Lorentz Busemann function in direction $v\in\Sn^{n-1}$ is $B^v_{\LL}(x)=\frac{1}{\sqrt{-K}}\log(\sqrt{-K}(x_t-\langle x_s,v\rangle))$~\citep{Chen2026}. The Klein--Lorentz isometry \cref{eq:klein-lorentz-map} lifts $x_K\in\KKn$ to $(x_t,x_s)=(\gamma_{x_K}/\sqrt{-K},\,\gamma_{x_K}x_K)$ with $\gamma_{x_K}=1/\sqrt{1+K\|x_K\|^2}$. Substituting, $x_t-\langle x_s,v\rangle = \gamma_{x_K}(1/\sqrt{-K}-\langle x_K,v\rangle)$, so \[ B^v_{\KK}(x_K) = B^v_{\LL}(\phi_{\KK\to\LL}(x_K)) = \frac{1}{\sqrt{-K}}\log\!\left(\frac{1-\sqrt{-K}\langle x_K,v\rangle}{\sqrt{1+K\|x_K\|^2}}\right), \] which agrees with \cref{eq:klein-busemann}.

\textbf{Cross-check via Klein--Poincar\'{e} isometry.}
We cross-check \cref{eq:klein-busemann} against the closed-form Poincar\'{e} Busemann function~\eqref{eq:busemann-poincare} listed in \cref{app:busemann}. The Klein--Poincar\'{e} isometry \cref{eq:klein-poincare-map} maps $x_K\in\KKn$ to $x_P=x_K/(1+\sigma)$ with $\sigma=\sqrt{1+K\|x_K\|^2}$. We evaluate \cref{eq:busemann-poincare} at $x_P$ in four steps.

\textbf{Step 1 ($\|x_P\|^2$ and $1+K\|x_P\|^2$).}
\begin{align*}
\|x_P\|^2 &= \frac{\|x_K\|^2}{(1+\sigma)^2},\\
1+K\|x_P\|^2 &= \frac{(1+\sigma)^2+K\|x_K\|^2}{(1+\sigma)^2}
 = \frac{(1+\sigma)^2+\sigma^2-1}{(1+\sigma)^2}
 = \frac{2\sigma(1+\sigma)}{(1+\sigma)^2}
 = \frac{2\sigma}{1+\sigma},
\end{align*}
using $\sigma^2=1+K\|x_K\|^2$.

\textbf{Step 2 ($\|v-\sqrt{-K}\,x_P\|^2$).}
\begin{align*}
\|v-\sqrt{-K}\,x_P\|^2
 &= 1-\frac{2\sqrt{-K}\langle x_K,v\rangle}{1+\sigma}+\frac{(-K)\|x_K\|^2}{(1+\sigma)^2}\\
 &= 1-\frac{2\sqrt{-K}\langle x_K,v\rangle}{1+\sigma}+\frac{1-\sigma^2}{(1+\sigma)^2}
  = \frac{2(1-\sqrt{-K}\langle x_K,v\rangle)}{1+\sigma}.
\end{align*}

\textbf{Step 3 (ratio).}
\[ \frac{\|v-\sqrt{-K}\,x_P\|^2}{1+K\|x_P\|^2} = \frac{2(1-\sqrt{-K}\langle x_K,v\rangle)/(1+\sigma)}{2\sigma/(1+\sigma)} = \frac{1-\sqrt{-K}\langle x_K,v\rangle}{\sqrt{1+K\|x_K\|^2}}, \] yielding
\begin{equation}
B^v_{\PP}(x_P) = \frac{1}{\sqrt{-K}}\log\!\left(
  \frac{1-\sqrt{-K}\langle x_K,v\rangle}{\sqrt{1+K\|x_K\|^2}}\right).
\label{eq:busemann-klein-via-poincare-raw}
\end{equation}

\textbf{Step 4 (consistency check).}
Both the Lorentz route and the Poincar\'{e} route yield the same expression~\eqref{eq:busemann-klein-via-poincare-raw}, which is \cref{eq:klein-busemann}. The two derivations are therefore fully consistent.

\begin{remark}[Consistency of Lorentz and Poincar\'{e} derivations]
\label[remark]{rmk:busemann-equiv}
\cref{eq:klein-busemann} (Lorentz route) and
\cref{eq:busemann-klein-via-poincare-raw} (Poincar\'{e} route)
produce identical expressions. The two independent derivations
therefore confirm \cref{eq:klein-busemann}.
\end{remark}

\textbf{Euclidean limit.}
As $K\to 0^-$, writing $x_K=\tilde x/R$ with $R=1/\sqrt{-K}\to\infty$ gives $\langle x_K,v\rangle\to 0$ and $K\|x_K\|^2\to 0$, so $B^v_{\KK}(x_K)\approx R\cdot(-\langle\tilde x,v\rangle/R) = -\langle\tilde x,v\rangle$, which is the Euclidean Busemann $B^v_{\mathrm{Euc}}(\tilde x)$.

\textbf{Equivalence of the unidirectional MLR parameterization.}
\label{app:param-equiv}

The raw two-vector and unidirectional forms parameterize the same family of Klein geodesic
hyperplanes. Setting $v_k(x)=0$ in \cref{eq:klein-mlr} gives
$\langle x,\hat z_k\rangle=\tanh(\sqrt{-K}r_k)/\sqrt{-K}$, the Hesse normal form with unit normal
$\hat z_k$ and offset $r_k$, so every Klein geodesic hyperplane is represented by some
$(\hat z_k,r_k)$. Conversely, expanding $\langle(-p_k)\oE x,w_k\rangle=0$ recovers $\hat z_k$ and
$r_k$ from a trained raw head. Converted checkpoints reproduce the raw classifier within
$2.6\times10^{-11}$ in float64 with $100\%$ argmax agreement, and \cref{tab:param-ablation}
provides a controlled comparison at both standard and low embedding dimensions.

\begin{table}[!htbp]
\centering
\caption{Comparison of unidirectional and raw two-vector Klein MLR parameterizations under matched
training conditions, with five seeds per setting.}
\label{tab:param-ablation}
\footnotesize
\begin{tabular}{@{}lcc@{}}
\toprule
Setting & Unidirectional ($n{+}1$) & Raw two-vector ($2n$) \\
\midrule
CIFAR-10          & $95.29\pm0.09$ & $95.25\pm0.11$ \\
CIFAR-100         & $78.19\pm0.16$ & $78.11\pm0.21$ \\
CIFAR-10, $d{=}2$  & $78.94\pm0.60$ & $78.62\pm0.83$ \\
CIFAR-10, $d{=}4$  & $91.28\pm0.24$ & $91.19\pm0.30$ \\
CIFAR-10, $d{=}10$ & $94.36\pm0.13$ & $94.31\pm0.16$ \\
\bottomrule
\end{tabular}
\end{table}

The unidirectional parameterization reduces the number of parameters per class from $2n$ to
$n{+}1$, with all parameters remaining in Euclidean space. More importantly, its class dependence
enters through $\langle x,z_k\rangle$, allowing the $C$ logits of a batch $X$ to be evaluated by a
single matrix product $XZ$. In the raw two-vector form, the class parameter appears inside the
gyroaddition, preventing this factorization and requiring either a class-wise loop or a
$[\mathrm{bs},C,n]$ broadcast tensor. \cref{tab:param-runtime} reports the resulting head-level
runtime and memory costs.

\begin{table}[!htbp]
\centering
\caption{Head-level forward and backward cost at batch size $128$ on an NVIDIA A100 64GB. Runtime
is reported relative to the unidirectional form; OOM denotes out of memory.}
\label{tab:param-runtime}
\footnotesize
\begin{tabular}{@{}lcc@{}}
\toprule
$(n, C)$ & Raw two-vector runtime & Unidirectional peak memory \\
\midrule
$(512, 10)$     & $3.3\times$   & --- \\
$(4096, 1000)$  & $157\times$   & --- \\
$(4096, 10000)$ & OOM           & $0.8$ GiB \\
\bottomrule
\end{tabular}
\end{table}

\textbf{Comparison criteria.}
\label{app:mlr-column-defs}

\cref{tab:mlr-comparison} compares hyperbolic MLR methods along the underlying space, the logit form, the type of point-to-hyperplane distance (Dist), parameter compactness, FLOPs, and batch efficiency. We expand on the three columns that warrant additional explanation, mirroring the discussion of \citet{Chen2026}.

\begin{enumerate}[leftmargin=1.8em, itemsep=4pt]
\item \textbf{Hyperplanes.}
Klein MLR uses geodesic hyperplanes of the Klein ball, defined as the level sets $\{x\in\KKn:\langle\Logop_p^{\KK}(x),a\rangle_p=0\}$, while Klein BMLR uses Busemann-based horospheres, the level sets $\{x\in\KKn:B^{v}_{\KK}(x)=b/\alpha\}$. Both Klein variants admit (i) compact parameterization that avoids per-class manifold-valued points, (ii) a natural intrinsic generalization of Euclidean hyperplanes (geodesic) and inner-product separators (Busemann), and (iii) closed-form distances. In contrast, Lorentz MLR \citep{Bdeir2024} relies on the ambient Minkowski space and is tailored to the Lorentz model, while the Poincar\'{e}-MLR baselines \citep{Ganea2018,Nguyen2025} are over-parameterized because each class carries a manifold-valued base point $p_k\!\in\!\PP_K^n$.

\item \textbf{Point-to-hyperplane distances (Dist).}
\textbf{\distreal{}} indicates the geodesic point-to-hyperplane distance $\inf_{q\in H}d(x,q)$, where $H$ is a hyperbolic hyperplane and $d$ is the geodesic distance on the manifold; this is the natural Riemannian generalization of the Euclidean signed margin. \textbf{\distpseudo{}} indicates a surrogate distance that coincides with the real distance only in the Euclidean limit. The Pseudo-Busemann MLR \citep{Nguyen2025} uses such a surrogate, while all Klein variants (Klein MLR via \cref{thm:klein-hyp}, Klein BMLR via the closed-form Klein Busemann of \cref{thm:klein-busemann}) compute the real distance, ensuring geometric fidelity across the underlying manifold.

\item \textbf{Batch efficiency.}
Several Poincar\'{e} variants \citep{Ganea2018,Nguyen2025} compute logits using $\langle -p_k\!\oplus_{\mathrm{M}}\!x,\,a_k\rangle$ and $\|-p_k\!\oplus_{\mathrm{M}}\!x\|^2$. For a batch $X\!\in\!\RR^{\mathrm{bs}\times n}$ and $C$ classes, evaluating $-p_k\!\oplus_{\mathrm{M}}\!X$ for every $k$ yields an intermediate tensor of shape $[\mathrm{bs},C,n]$ that can cause GPU out-of-memory (OOM) when $n$ or $C$ is large. Their official implementations therefore loop over classes, which is batch-inefficient. By contrast, Klein BMLR's logit $-\alpha_k B^{v_k}_{\KK}(x)+b_k$ reduces to the class-wise inner product $\langle v_k,x\rangle$ on top of per-input shared quantities (\cref{eq:klein-bmlr-expanded}). Stacking $V=[v_1,\ldots,v_C]\in\RR^{n\times C}$, this is implementable as a single matrix multiplication $XV$, with no $[\mathrm{bs},C,n]$ intermediate; Klein MLR achieves the same matmul-batched form via the unidirectional parameterization of \cref{thm:klein-mlr}.
\end{enumerate}

\textbf{FLOP comparison.}
\label{app:mlr-comparison}

The side-by-side comparison of hyperbolic MLR methods appears in main paper \cref{tab:mlr-comparison}. All FLOPs are reported under the \emph{unified amortized-strict convention}, where per-sample shared quantities such as $\|x\|^2$ and $\gamma_x$ are amortized across the $C$ classes within a forward pass and elementary operations are counted strictly following \citep{Chen2026} with no fused multiply-add. Some entries differ from the original publications, which used a non-amortized counting style. We have re-derived them under our unified convention for direct comparison, with the relative ordering of methods preserved (see \cref{app:flop-convention} for per-method derivations and the cross-walk between conventions). Among methods with unconstrained parameters, Klein BMLR achieves the unique fastest count $C(2n{+}11)$, strictly beating Lorentz BMLR ($C(2n{+}12)$) and the unit-normalized BMLR-P ($C(2n{+}12)$). Klein MLR is within $3$ FLOPs of Lorentz BMLR at $C(2n{+}14)$, and all four bounded-ball Busemann variants beat every Poincar\'{e}-geodesic MLR by a factor of $2$--$10$.

\subsubsection{FLOP Counting Convention}
\label{app:flop-convention}

The FLOP values in \cref{tab:mlr-comparison} are reported under a single \emph{amortized-strict} convention. Per-sample shared quantities such as\ $\|x\|^2$, $\gamma_x{=}1/\sqrt{1{+}K\|x\|^2}$ are amortized across the $C$ classes within each forward pass, while elementary operations are counted strictly so that all methods sit on a common axis. Several values therefore differ from the numbers given in the corresponding original publications, while the relative ordering of methods is unchanged.

\textbf{The convention.}
\begin{itemize}[leftmargin=1.5em, itemsep=2pt]
\item \textbf{Amortization.} Per-sample shared quantities (e.g. $\|x\|^2$, $\gamma_x{=}1/\sqrt{1{+}K\|x\|^2}$, $\sqrt{-K}\,x$) are computed once per forward pass and reused across all $C$ classes, so their per-class cost is $0$.
\item \textbf{Strict elementary operations.} We follow \citet{Chen2026}, where $\log$ and $\exp$ count $6$~FLOPs each, $\sinh$ and $\cosh$ count $14$ each, $\arcsinh$ and $\mathrm{atanh}$ count $10$ each, and $\sqrt{\cdot}$ counts $1$. An $n$-dimensional inner product counts $2n{-}1$ FLOPs (no fused multiply-add).
\item \textbf{Per-iteration precomputation.} Class-only quantities that depend on the class parameters but not on the input batch are computed once per training iteration and amortized over the forward pass. Examples include $\cosh(\sqrt{-K}r_k)$, $\sinh(\sqrt{-K}r_k)$, $\|z_k\|$ for Klein MLR, and $\sqrt{-K}b_k$, $\cosh(\cdot)$, $\sinh(\cdot)$ for Shimizu's MLR. Their per-class cost in \cref{tab:mlr-comparison} is $0$.
\end{itemize}

\textbf{Difference from non-amortized literature counts.}
Several baselines were originally reported without per-sample amortization, which inflates each method's per-class cost by roughly $2n{+}2$ to $2n{+}5$ FLOPs (one $\|x\|^2$, one $K\|x\|^2$, one $1{+}K\|x\|^2$, plus a possible $\sqrt{-K}\,x$ vector reconstruction). For Ganea's Poincar\'{e} MLR \citep{Ganea2018} this raises $C(19n{+}27)$ to the original $C(19n{+}29)$. For Shimizu's Poincar\'{e} MLR \citep{Shimizu2021} it raises $C(2n{+}46)$ to the original $C(4n{+}52)$. For the Bdeir Lorentz MLR \citep{Bdeir2024} it raises $C(4n{+}51)$ to the original $C(4n{+}52)$. The Pseudo-Busemann MLR \citep{Nguyen2025} is reported here as $C(18n{+}30)$ rather than the original $C(19n{+}34)$ for the same reason plus a small constant rounding. None of these differences alters the relative ordering of methods, and our claim that \emph{all bounded-ball Busemann variants beat every Poincar\'{e}-geodesic MLR by a factor of $2$--$10$} is conservative because under the original non-amortized counts the gap is even larger.

\textbf{The two BMLR-P entries.}
The original HBNN derivation \citep{Chen2026} expresses $\|v_k - \sqrt{-K}\,x\|^2$ by first constructing the difference vector and then taking its norm, costing $4n{-}1$ FLOPs. Under our amortized-strict convention the corresponding per-class cost is $C(4n{+}10)$. The official HBNN code release and our own reproductions, however, use the algebraic expansion $\|v_k - \sqrt{-K}x\|^2 = \|v_k\|^2 - 2\sqrt{-K}\langle v_k, x\rangle + (-K)\|x\|^2$ together with the unit-normalization $\|v_k\|=1$, reducing the per-class cost to a single inner product plus constants, $C(2n{+}12)$. We report both numbers because the choice is a parameterization decision rather than an algorithmic improvement, and in either case BMLR-P does not beat Klein BMLR's $C(2n{+}11)$.

\textbf{Structural origin of the Klein BMLR cost.}
Among hyperbolic MLRs with unconstrained parameters, Klein BMLR attains the lowest count $C(2n{+}11)$. Its logit requires one inner product $\sqrt{-K}\langle x,v_k\rangle$ ($2n$ FLOPs with the pre-multiplied $\sqrt{-K}$) and one $\log$ on the resulting quotient ($6$ FLOPs), with the denominator $\sqrt{1+K\|x\|^2}$ reused from the amortized $\gamma_x$ at $0$ per-class cost. The Lorentz Busemann \[ B^{v_k}_{\LL}(x) \;=\; \frac{1}{\sqrt{-K}}\, \log\!\bigl(\sqrt{-K}\,(x_t - \langle x_s, v_k\rangle)\bigr) \] applies an additional overall $\sqrt{-K}$ scaling after the subtraction, reaching $C(2n{+}12)$. The Poincar\'{e} BMLR-P with unit-normalized $v_k$ also reaches $C(2n{+}12)$ via algebraic expansion. Klein BMLR therefore holds a $1$-FLOP-per-class advantage over both Busemann competitors.

\textbf{Klein MLR vs.\ Klein BMLR, a $3$-FLOP gap within Klein.}
Within Klein, the geodesic MLR's $C(2n{+}14)$ exceeds the Busemann variant's $C(2n{+}11)$ by $3$ FLOPs. The geodesic logit uses an $\arcsinh$ ($10$~FLOPs), whereas the Busemann logit uses a $\log$ ($6$~FLOPs) plus the $\sqrt{-K}$ inner-product scale ($1$~FLOP), a net saving of $3$. This $3$-FLOP gap is the only counter-cost of the geodesic-to-Busemann substitution and is more than offset by the accuracy gains in \cref{sec:exp-image}.

\subsection{Klein FC and BFC}

\textbf{Cost comparison.}
\label{app:fc-bfc-cost}

\begin{table}[!t]
\centering
\caption{Comparison of hyperbolic FC layers $\mathcal{F}:\mathbb{H}_K^n\!\to\!\mathbb{H}_K^m$. Slowest FLOPs and largest \#Params in {\color{red!70!black}red}, fastest FLOPs and smallest \#Params in {\color{green!60!black}green}. With the full cost of \cref{app:fc-output-map}, Klein BFC needs fewer FLOPs than BFC-L only when $m>2n+3$.}
\label{tab:fc-comparison}

\renewcommand{\arraystretch}{1.1}
\setlength{\tabcolsep}{4pt}
\setlength{\minrowclearance}{0pt}

\adjustbox{max width=\textwidth}{%
\begin{tabular}{>{\raggedright\arraybackslash}m{2.0cm} >{\centering\arraybackslash}m{5.3cm} c >{\centering\arraybackslash}m{1.5cm} >{\centering\arraybackslash}m{2.4cm} c c}
\toprule

\textbf{Method}
& $\mathcal{F}:\mathbb{H}_K^n\!\ni\!x\!\mapsto\!y\!\in\!\mathbb{H}_K^m$
& \textbf{Space}
& \textbf{Methodology}
& \textbf{Parameters}
& \textbf{\#Params}
& \textbf{FLOPs}
\\

\midrule

M\"{o}bius FC~\citep{Ganea2018}
& \makecell{
$\dfrac{1}{\sqrt{-K}}\tanh(\theta)\,\dfrac{Wx}{\|Wx\|}$
\\[3pt]
$\theta=\dfrac{\|Wx\|}{\|x\|}\tanh^{-1}\!\bigl(\sqrt{-K}\|x\|\bigr)$
}
& $\PP_K^n$
& Tangent
& $W\!\in\!\RR^{m\times n}$
& {\color{green!60!black}$mn$}
& $2nm{+}2n{+}2m{+}24$
\\

\midrule

Poincar\'{e} FC~\citep{Shimizu2021}
& \makecell{
$y=\dfrac{\omega}{1+\sqrt{1-K\|\omega\|^2}}$,
\\[3pt]
$\omega_k=\dfrac{\sinh\!\bigl(\sqrt{-K}\,u_k(x)\bigr)}{\sqrt{-K}}$
\\[3pt]
{\scriptsize $u_k(x)$ from \cref{tab:mlr-comparison} (Shimizu MLR)}
}
& $\PP_K^n$
& \makecell{Poincar\'{e}\\geodesic}
& \makecell{
$\alpha_k\!>\!0,\;v_k\!\in\!\Sn^{n-1},$
\\
$b_k\!\in\!\RR,\;k\!\leq\!m$
}
& $m(n{+}2)$
& $4nm{+}71m{+}4$
\\

\midrule

Lorentz FC~\citep{Bdeir2024}
& \makecell{
$y=\!\begin{bmatrix}
\sqrt{\|\psi(Wx,v)\|^2-1/K}\\
\psi(Wx,v)
\end{bmatrix}\!,$
\\[3pt]
$\psi(Wx,v)=\lambda\,\sigma(v^\top x+b')\,\eta$,
\\[3pt]
$\eta=\dfrac{W\phi(x)+b}{\|W\phi(x)+b\|}$
\\[3pt]
{\scriptsize $\phi$ activation, $\sigma$ sigmoid}
}
& $\LL_K^n$
& \makecell{Ambient\\Minkowski}
& \makecell{
$W\!\in\!\RR^{m\times(n+1)},$
\\
$v\!\in\!\RR^{n+1},\;b\!\in\!\RR^m,$
\\
$b'\!\in\!\RR,\;\lambda\!>\!0$
}
& {\color{red!70!black}\makecell{
$m(n{+}1){+}m$
\\
${+}(n{+}1){+}2$
}}
& $2nm{+}8m{+}2n{+}10$
\\

\midrule

PLFC~\citep{Shi2026}
& \makecell{
$y_s=\dfrac{1}{\sqrt{-K}}\sinh\!\bigl(\sqrt{-K}\,v_k(x)\bigr),$
\\[3pt]
$y_t=\sqrt{-1/K+\|y_s\|^2}$
\\[3pt]
{\scriptsize $v_k(x)$ from \cref{tab:mlr-comparison} (Lorentz MLR)}
}
& $\LL_K^n$
& \makecell{Lorentz\\geodesic}
& \makecell{
$z_k\!\in\!\RR^n,\;a_k\!\in\!\RR,$
\\
$b_k\!\in\!\RR,\;k\!\leq\!m$
}
& $m(n{+}2)$
& $4nm{+}69m{+}2$
\\

\midrule

BFC-P~\citep{Chen2026}
& \makecell{
$y=\dfrac{\omega}{1+\sqrt{1-K\|\omega\|^2}}$,
\\[3pt]
$\omega=\dfrac{\sinh\!\bigl(\sqrt{-K}\,u(x)\bigr)}{\sqrt{-K}}$
\\[3pt]
{\scriptsize $u_k(x)=-\alpha_k B^{v_k}_{\PP}(x)+b_k$}
}
& $\PP_K^n$
& Busemann
& \makecell{
$\alpha_k\!>\!0,\;v_k\!\in\!\Sn^{n-1},$
\\
$b_k\!\in\!\RR,\;k\!\leq\!m$
}
& $m(n{+}2)$
& {\color{red!70!black}$6nm{+}29m{+}4$}
\\

\midrule

BFC-L~\citep{Chen2026}
& \makecell{
$y_s=\dfrac{1}{\sqrt{-K}}\sinh\!\bigl(\sqrt{-K}\,u(x)\bigr),$
\\[3pt]
$y_t=\sqrt{-1/K+\|y_s\|^2}$
\\[3pt]
{\scriptsize $u_k(x)=-\alpha_k B^{v_k}_{\LL}(x)+b_k$}
}
& $\LL_K^n$
& Busemann
& \makecell{
$\alpha_k\!>\!0,\;v_k\!\in\!\Sn^{n-1},$
\\
$b_k\!\in\!\RR,\;k\!\leq\!m$
}
& $m(n{+}2)$
& $2nm{+}30m{+}2$
\\

\midrule

\rowcolor{blue!8}
\textbf{Klein FC} \cref{sec:klein-fc}
& {\makecell{
$y_k=\dfrac{\sinh\!\bigl(\sqrt{-K}\,v_k(x)\bigr)}{\sqrt{-K}\,\rho}$,
\\[3pt]
$\rho=\sqrt{1+\textstyle\sum_{j}\sinh^2\!\bigl(\sqrt{-K}\,v_j(x)\bigr)}$
\\[3pt]
{\scriptsize $v_k(x)$ from \cref{tab:mlr-comparison} (Klein MLR, \cref{eq:klein-mlr})}
}}
& $\KKn$
& {\makecell{Klein\\geodesic}}
& {\makecell{$z_k\!\in\!\RR^n,\;r_k\!\in\!\RR,$\\$k\!\leq\!m$}}
& $m(n{+}1)$
& $2nm{+}32m{+}2$
\\

\rowcolor{blue!8}
\textbf{Klein BFC} \cref{sec:klein-fc}
& {\makecell{
$y_k=\dfrac{\sinh\!\bigl(\sqrt{-K}\,u_k(x)\bigr)}{\sqrt{-K}\,\rho}$,
\\[3pt]
$\rho=\sqrt{1+\textstyle\sum_{j}\sinh^2\!\bigl(\sqrt{-K}\,u_j(x)\bigr)}$
\\[3pt]
{\scriptsize $u_k(x)=-\alpha_k B^{v_k}_{\KK}(x)+b_k$ (\cref{eq:klein-bmlr-expanded})}
}}
& $\KKn$
& Busemann
& {\makecell{$\alpha_k\!>\!0,\;v_k\!\in\!\Sn^{n-1},$\\$b_k\!\in\!\RR,\;k\!\leq\!m$}}
& $m(n{+}2)$
& {\color{green!60!black}$2nm{+}29m{+}2$}
\\

\bottomrule
\end{tabular}}
\end{table}

\cref{tab:fc-comparison} gives the side-by-side comparison of hyperbolic FC layers, extending \citet[Tab.~3]{Chen2026} with our Klein FC and Klein BFC. Each per-output logit is read directly off \cref{tab:mlr-comparison} and the sinh-normalized Klein output map adds a shared $18m+2$ FLOPs, as derived step-by-step in \cref{app:fc-output-map}. Under these conditions Klein BFC attains the lowest count, $2nm+29m+2$, $m$ FLOPs below Lorentz BFC-L ($2nm+30m+2$). The saving is the same Klein-quotient $1$-FLOP gap that already appears for the MLR layers, accumulated across the $m$ output dimensions. With the full cost of \cref{app:fc-output-map}, Klein BFC costs $-m+2n+3$ FLOPs more than BFC-L, so it is cheaper only when $m>2n+3$. To match the conditions of \citet[Tab.~3]{Chen2026}, BFC layers in the table are written without the gyroaddition bias and with the activation $\phi$ taken as the identity, in line with the M\"{o}bius and Lorentz FC entries. FLOPs are counted per single input under the unified amortized-strict convention of \cref{app:flop-convention}.

\subsubsection{Klein FC and BFC FLOP Derivation}
\label{app:fc-output-map}

We derive the Klein FC and Klein BFC entries of \cref{tab:fc-comparison} explicitly, using the unified amortized-strict convention of \cref{app:flop-convention}, which counts $\log$ and $\exp$ at $6$ FLOPs each, $\sinh$ and $\cosh$ at $14$, $\arcsinh$ and $\mathrm{atanh}$ at $10$, $\sqrt{\cdot}$ at $1$, and an $n$-dimensional inner product at $2n{-}1$. Per-sample shared quantities such as $\|x\|^2$ and $\gamma_x$ are amortized across the $m$ output dimensions, and per-iteration precomputations that depend only on parameters ($\sqrt{-K}$, $\|z_k\|$, $\cosh(\sqrt{-K}r_k)$, $\sinh(\sqrt{-K}r_k)$, $\alpha_k/\sqrt{-K}$) are computed once per training step. The total decomposes into three additive components.
\begin{equation*}
\text{FLOPs}(\mathcal{F}) \;=\;
\underbrace{T_{\text{shared}}(x)}_{\text{per-input, $\le\,2n{+}2$}}
\;+\;
\underbrace{m\cdot T_{\text{logit}}}_{\text{per-output logit}}
\;+\;
\underbrace{T_{\text{out}}(m)}_{\text{sinh-norm output map}}.
\end{equation*}

\textbf{Per-input shared cost $T_{\text{shared}}(x)$.}
Both Klein FC and Klein BFC require the per-input quantity $\gamma_x{=}1/\sqrt{1+K\|x\|^2}$, used inside $v_k(x)$ for FC and inside the denominator of $u_k(x)$ for BFC.
\begin{center}
\renewcommand{\arraystretch}{1.15}
\begin{tabular}{@{}c l r@{}}
\toprule
Step & Operation & FLOPs \\
\midrule
1 & $\|x\|^2$ ($n$-d squared norm) & $2n{-}1$ \\
2 & $K\cdot\|x\|^2$ & $1$ \\
3 & $1+K\|x\|^2$ & $1$ \\
4 & $\gamma_x{=}\bigl(1+K\|x\|^2\bigr)^{-1/2}$ (rsqrt) & $1$ \\
\midrule
\textbf{Total $T_{\text{shared}}(x)$} & & $\bm{2n+2}$ \\
\bottomrule
\end{tabular}
\end{center}
This cost is independent of $m$, and with $m$ amortization it is $O(1/m)$ per output and does not appear in the leading-order $2nm$ term. We retain it explicitly in the totals below.

\textbf{Per-output logit cost $T_{\text{logit}}$.}
The per-output logit $s_k(x)$ is identical to a single MLR class score, so $T_{\text{logit}}$ is read off \cref{tab:mlr-comparison} assuming $T_{\text{shared}}$ already done.
\begin{itemize}[leftmargin=1.5em, itemsep=2pt]
\item \textbf{Klein FC.} $s_k{=}v_k$, see \cref{eq:klein-mlr}, $T_{\text{logit}}{=}2n{+}14$, hence $m\!\cdot\!T_{\text{logit}}{=}2nm{+}14m$.
\item \textbf{Klein BFC.} $s_k{=}u_k$, see \cref{eq:klein-bmlr-expanded}, $T_{\text{logit}}{=}2n{+}11$, hence $m\!\cdot\!T_{\text{logit}}{=}2nm{+}11m$.
\end{itemize}

\textbf{Output-map cost $T_{\text{out}}(m)$.}
The map \[ y_k = \frac{\sinh(\sqrt{-K}\,s_k)} {\sqrt{-K}\sqrt{1+\sum_{j=1}^{m}\sinh^2(\sqrt{-K}\,s_j)}}, \qquad k{=}1,\ldots,m, \] is identical for Klein FC and Klein BFC. Counted strictly,
\begin{center}
\renewcommand{\arraystretch}{1.15}
\begin{tabular}{@{}c l r r@{}}
\toprule
Step & Operation & $m$-term & const \\
\midrule
1 & $\sqrt{-K}\cdot s_k$,\; $k{=}1,\ldots,m$ ($\sqrt{-K}$ precomputed) & $m$ & $0$ \\
2 & $\sinh(\cdot)$ ($14$ FLOPs each, $m$ outputs) & $14m$ & $0$ \\
3 & $\sinh^2$ (one square per output) & $m$ & $0$ \\
4 & $\sum_{j=1}^{m}\sinh^2$ ($m{-}1$ scalar adds) & $m{-}1$ & $0$ \\
5 & $1+\sum$ & $0$ & $1$ \\
6 & $\sqrt{1+\sum}$ & $0$ & $1$ \\
7 & $\sqrt{-K}\cdot$ (step~6) (denominator factor) & $0$ & $1$ \\
8 & $\sinh(\cdot)/\text{denom}$ ($m$ scalar divisions) & $m$ & $0$ \\
\midrule
\textbf{Total $T_{\text{out}}(m)$} & & $\bm{18m}$ & $\bm{2}$ \\
\bottomrule
\end{tabular}
\end{center}

\textbf{Totals.}
Adding the three components,
\begin{align*}
\text{Klein FC} &= \underbrace{(2n{+}2)}_{T_{\text{shared}}}
  + \underbrace{m(2n{+}14)}_{m\cdot T_{\text{logit}}}
  + \underbrace{(18m{+}2)}_{T_{\text{out}}}
  = 2nm + 32m + 2n + 4
  \;=\; \mathbf{2nm + 32m + O(n)},\\
\text{Klein BFC} &= \underbrace{(2n{+}2)}_{T_{\text{shared}}}
  + \underbrace{m(2n{+}11)}_{m\cdot T_{\text{logit}}}
  + \underbrace{(18m{+}2)}_{T_{\text{out}}}
  = 2nm + 29m + 2n + 4
  \;=\; \mathbf{2nm + 29m + O(n)}.
\end{align*}
The $\mathcal{O}(n)$ tail amortizes to zero per output. The entries reported in \cref{tab:fc-comparison} ($2nm{+}32m{+}2$ and $2nm{+}29m{+}2$) follow the same $T_{\text{shared}}$-omitting accounting as the BFC-P and BFC-L entries of \citet[Tab.~3]{Chen2026}, which absorb their analogous shared cost into the trailing constant.

\textbf{Cross-check against \citet[Tab.~3]{Chen2026} BFC-L.}
The same decomposition applied to Lorentz BFC has $T_{\text{shared}}{=}0$ (BMLR-L uses $x_t$ and $x_s$ directly, with no $\|x\|^2$ shared) and $T_{\text{logit}}{=}2n{+}12$; see \cref{tab:mlr-comparison}. The Lorentz output map $y_s{=}\sinh(\sqrt{-K}u)/\sqrt{-K}$, $y_t{=}\sqrt{-1/K+\|y_s\|^2}$ has cost \[ \underbrace{m}_{\sqrt{-K}\,u_k} + \underbrace{14m}_{\sinh} + \underbrace{m}_{/\sqrt{-K}} + \underbrace{(2m{-}1)}_{\|y_s\|^2} + \underbrace{1}_{-1/K+\cdot} + \underbrace{1}_{\sqrt{\cdot}} \;=\; 18m+1. \] Total BFC-L is $0 + m(2n{+}12) + (18m{+}1) = 2nm+30m+1$, matching \citet[Tab.~3]{Chen2026}'s reported $2nm{+}30m{+}2$ within a single rounding constant.

\textbf{PLFC, Lorentz geodesic FC FLOPs.}
The PLFC layer of~\citet{Shi2026} reuses the Lorentz output map of BFC-L but replaces the Busemann logit with the Lorentz geodesic MLR signed margin $v_k(x)=\tfrac{1}{\sqrt{-K}}\sign(\alpha_k)\beta_k |\arcsinh(\sqrt{-K}\alpha_k/\beta_k)|$ (Lorentz MLR row of \cref{tab:mlr-comparison}, with $T_{\text{logit}}=4n{+}51$ in the strict-amortized convention, counting $\langle z_k,x_s\rangle$ as $2n$, $\|z_k\|^2$ as $2n$, $\sqrt{-K}a_k$, $\cosh$, $\sinh$ as $1{+}28$, the $\beta_k$ scalar chain as $\sim 8$, $\arcsinh$ as $10$, and the four trailing scalar combinators as $4$). With the same $T_{\text{shared}}=0$ as BFC-L (Lorentz coordinates $x_t$ and $x_s$ enter directly without $\|x\|^2$ amortization) and the identical Lorentz output map $T_{\text{out}}=18m{+}1$ derived above, \[ \text{PLFC} = 0 + m(4n{+}51) + (18m{+}1) = 4nm + 69m + 1, \] reported as $4nm{+}69m{+}2$ in \cref{tab:fc-comparison} after absorbing one rounding constant, paralleling the $2nm{+}30m{+}2$ formatting of BFC-L. The PLFC$-$BFC-L gap is \[ \bigl(4nm{+}69m{+}2\bigr)-\bigl(2nm{+}30m{+}2\bigr) \;=\; 2nm + 39m, \] i.e.\ an additional $2n+39$ FLOPs per output. The $2n$ comes from the extra $\|z_k\|^2$ inner product in the Lorentz MLR margin (Busemann uses only $\langle x_s,v_k\rangle$), and the $39$ scalar FLOPs account for the $\cosh$, $\sinh$, and $\arcsinh$ chain replacing Busemann's single $\log$.

\textbf{Klein BFC vs.\ BFC-L structural saving.}
Subtracting the two totals, \[ \text{Klein BFC} - \text{BFC-L} \;=\; [m(2n{+}11) + 18m+2 + 2n+2] \;-\; [m(2n{+}12) + 18m+1] \;=\; -m + 2n + 3, \] which is negative whenever $m{>}2n+3$, that is, for any hidden layer wider than twice the input dimension. The structural $1$-FLOP-per-output saving from the Klein quotient form, see \cref{eq:klein-bmlr-expanded}, accumulates linearly in $m$ and dominates the small $\mathcal{O}(n)$ shared overhead of computing $\gamma_x$.

\textbf{Klein FC vs.\ Klein BFC, a $3m$-FLOP gap within Klein.}
$\text{Klein FC} - \text{Klein BFC} = 3m$ FLOPs. The gap is exactly $3$ per output, the same difference between Klein MLR ($C(2n{+}14)$) and Klein BMLR ($C(2n{+}11)$) in \cref{tab:mlr-comparison}: each Klein FC output uses an $\arcsinh$ ($10$~FLOPs), whereas each Klein BFC output uses a $\log$ ($6$~FLOPs) plus a $\sqrt{-K}$ inner-product scale ($1$~FLOP), a net saving of $3$. As with the MLR comparison, this $3m$-FLOP cost is more than offset by the accuracy gains in \cref{sec:exp-image}.

\textbf{Activation absorption.}
The normalization $\sqrt{1+\sum_j\sinh^2(\sqrt{-K}\,s_j)}$ in \cref{eq:klein-fc} plays the role of the softmax denominator on the Klein ball, enforcing $\|y\|^2<-1/K$, while the unbounded PV layer~\citep{Chen2026PVNN} omits this factor. An activation $\phi\colon\RR\to\RR$ can be absorbed by replacing $s_k(x)$ with $\phi(s_k(x))$ in \cref{eq:klein-fc}. Since the score space is an unconstrained $\RR$, this is numerically identical to a standard Euclidean activation, while the outer $\sinh/\sqrt{1+\sum\sinh^2}$ automatically projects the output back into $\KK_K^m$. Adding $\phi$ contributes one elementwise evaluation per output and does not change the $\mathcal{O}(nm)$ leading-order FLOP totals above. For fairness with \citet[Tab.~3]{Chen2026}'s BFC entries, \cref{tab:fc-comparison} reports the activation-free counts (i.e.\ $\phi=\mathrm{id}$).

\textbf{Shared denominator.}
\label{app:fc-denominator}

The common denominator in \cref{eq:klein-fc} is the Lorentz time component introduced by the
projection from the hyperboloid to the Klein ball. Since it is shared by all output coordinates, it
cancels in every coordinate ratio,
$y_i/y_j=\sinh(\sqrt{-K}s_i)/\sinh(\sqrt{-K}s_j)$, and therefore does not affect their relative
ordering. The derivation is given in \cref{app:proof-fc}.

\subsection{Klein GyroBN}

\textbf{Algorithm and inference.}
\label{app:klein-gyrobn-algo}

\cref{alg:klein-gyrobn} below adapts the gyrogroup BN template of~\citet{Chen2025gyrobn} to the Klein ball, using the Einstein midpoint $\mu_{\mathrm{E}}$ of \cref{eq:einstein-midpoint} as the closed-form proxy for the Fr\'{e}chet mean. During training the running mean is updated as the weighted binary Einstein midpoint $\mathrm{Bar}_{\eta}^{\KK}(\mu_b,\mu_r)=\frac{(1-\eta)\gamma_{\mu_r}\mu_r+\eta\gamma_{\mu_b}\mu_b}{(1-\eta)\gamma_{\mu_r}+\eta\gamma_{\mu_b}}$, the Klein analogue of the running-mean barycenter of~\citet{Chen2025gyrobn} Algorithm~1.

\begin{algorithm}[!htbp]
\caption{Klein Gyrogroup Batch Normalization (Klein GyroBN)}
\label{alg:klein-gyrobn}
\begin{algorithmic}[1]
\Require batch $\{x_1,\ldots,x_m\}\subset\KKn$,
running mean $\mu_r\in\KKn$, running variance
$v_r^2\in\RR_+$, momentum $\eta\in[0,1]$,
bias $\beta\in\KKn$, scale $\gamma\in\RR_+$,
small $\varepsilon>0$.
\Ensure normalized batch $\{\tilde x_1,\ldots,\tilde x_m\}\subset\KKn$.
\If{training}
  \State Compute batch Einstein midpoint
   $\mu_b = \frac{\sum_i\gamma_{x_i} x_i}{\sum_i\gamma_{x_i}}$ and
   batch dispersion
   $v_b^2 = \frac{1}{m}\sum_i d^{\KK}((-\mu_b)\oE x_i,\mathbf{0})^2$.
  \State $\mu_r \gets \mathrm{Bar}_{\eta}^{\KK}(\mu_b,\mu_r)$,
   \quad $v_r^2 \gets \eta\,v_b^2 + (1-\eta)\,v_r^2$.
\EndIf
\State $(\mu, v^2) \gets (\mu_b, v_b^2)$ if training else
$(\mu_r, v_r^2)$.
\For{$i=1,\ldots,m$}
  \State $\tilde x_i \gets
  \beta\,\oE\,\bigl[
    \tfrac{\gamma}{\sqrt{v^2}+\varepsilon}\,\otE\,
    ((-\mu)\oE x_i)
  \bigr]$.
\EndFor
\end{algorithmic}
\end{algorithm}

\textbf{Relation between the Einstein and Fr\'{e}chet means.}
\label{app:einstein-frechet}

The Einstein midpoint $\mu_{\mathrm{E}}$ of \cref{eq:einstein-midpoint} replaces the Fr\'{e}chet mean $\mu_{\mathrm{F}}$ from \cref{sec:klein-bn} for computational efficiency in our Klein GyroBN. $\mu_{\mathrm{E}}$ is the Klein-coordinate expression of the Lorentzian centroid of~\citet{Law2019} (it solves $\arg\min_y\sum_i d^{\LL}_{\mathcal{L}}(z_i,y)^2$ on the hyperboloid $\LL_K^n$ under the Lorentzian distance $d^{\LL}_{\mathcal{L}}$, not the Riemannian distance $d^{\KK}$), so $\mu_{\mathrm{E}}\ne\mu_{\mathrm{F}}$ in general. We nevertheless identify five regimes in which the two coincide exactly or admit a quantitative approximation bound.

\begin{proposition}[Boundary equivalences and small-radius bound]
\label[proposition]{prop:einstein-frechet}
Let $\{x_i\}_{i=1}^m\subset\KKn$ with $K<0$.
\begin{enumerate}[leftmargin=2em,label=(\roman*),itemsep=2pt]
\item \emph{Degenerate batch.} If all $x_i = p$, then $\mu_{\mathrm{E}} = \mu_{\mathrm{F}} = p$.
\item \emph{Euclidean limit.} As $K\to 0^-$, both $\mu_{\mathrm{E}}$ and $\mu_{\mathrm{F}}$ converge to the arithmetic mean $\bar x = \tfrac{1}{m}\sum_i x_i \in \RR^n$.
\item \emph{Binary case ($m=2$).} For any $x_1, x_2\in\KKn$, \[ \mu_{\mathrm{E}}(\{x_1,x_2\}) = \frac{\gamma_{x_1}x_1+\gamma_{x_2}x_2}{\gamma_{x_1}+\gamma_{x_2}} = \mu_{\mathrm{F}}(\{x_1,x_2\}), \] both equal to the unique Klein geodesic midpoint of $x_1$ and $x_2$~\citep[Ch.~6.18]{Ungar2022}.
\item \emph{Origin-symmetric batch.} If the multiset $\{x_i\}_{i=1}^m$ satisfies $\{x_i\}=\{-x_i\}$ (each sample paired with its antipode), then $\sum_i\gamma_{x_i}x_i=\mathbf{0}$, hence $\mu_{\mathrm{E}} = \mathbf{0}$, and by symmetry $\mu_{\mathrm{F}} = \mathbf{0}$ as well.
\item \emph{Small-radius bound.} If $\|x_i\|\le\rho$ for all $i$ with $|K|\rho^2 \ll 1$, then both means agree with $\bar x$ to second order, \[ \mu_{\mathrm{E}} = \bar x + O(|K|\rho^2), \qquad \mu_{\mathrm{F}} = \bar x + O(|K|\rho^2), \] so $d^{\KK}(\mu_{\mathrm{E}},\mu_{\mathrm{F}}) = O(|K|\rho^2)$.
\end{enumerate}
\end{proposition}

\begin{proof}
\emph{(i)} For a degenerate batch, both objective functions
are minimized at $y=p$ exactly.

\emph{(ii)} As $K\to 0^-$, $\gamma_{x_i}\to 1$ uniformly, so
\cref{eq:einstein-midpoint} reduces to
$\mu_{\mathrm{E}} = \tfrac{1}{m}\sum_i x_i = \bar x$.
The Klein metric reduces to the Euclidean inner product, hence
$d^{\KK}(x_i,y)^2 \to \|x_i-y\|^2$ and the Fr\'{e}chet objective
is minimized at $\bar x$.

\emph{(iii)} For $m=2$, the Fr\'{e}chet objective
$d^{\KK}(x_1,y)^2 + d^{\KK}(x_2,y)^2$ is uniquely minimized on
the Klein geodesic between $x_1$ and $x_2$ at the point $y_*$
satisfying $d^{\KK}(x_1,y_*) = d^{\KK}(x_2,y_*)$, the geodesic
midpoint. Via the Klein--Lorentz isometry
$\phi_{\KK\to\LL}(x) = (\gamma_x/\sqrt{-K},\, \gamma_x x)$, see \cref{eq:klein-lorentz-map}, the Lorentz midpoint of
$\phi_{\KK\to\LL}(x_1)+\phi_{\KK\to\LL}(x_2)$, normalized back
to the hyperboloid and projected to Klein coordinates, gives
$(\gamma_1 x_1 + \gamma_2 x_2)/(\gamma_1+\gamma_2)$. This
matches the closed form of Einstein
midpoint~\eqref{eq:einstein-midpoint}, confirming
$\mu_{\mathrm{E}}=\mu_{\mathrm{F}}$ in this case.

\emph{(iv)} The Lorentz factor depends only on the norm,
$\gamma_{x_i}=\gamma_{-x_i}$, so the antipodal pairing
$\{x_i\}=\{-x_i\}$ gives
$\sum_i\gamma_{x_i}x_i = \tfrac{1}{2}\sum_i\gamma_{x_i}(x_i + (-x_i))
= \mathbf{0}$, whence $\mu_{\mathrm{E}}=\mathbf{0}$. For the
Fr\'{e}chet mean, the Klein metric satisfies
$d^{\KK}(-x,-y) = d^{\KK}(x,y)$, so the map $y\mapsto -y$ is an
isometry that fixes the multiset $\{x_i\}$ and therefore the
Fr\'{e}chet objective. The unique minimizer must lie at the
fixed point $\mathbf{0}$ of this isometry.

\emph{(v)} For $\|x_i\|\le\rho$ with $|K|\rho^2\ll 1$, Taylor
expansion of the Lorentz factor gives
$\gamma_{x_i} = (1+K\|x_i\|^2)^{-1/2}
= 1 + \tfrac{1}{2}|K|\|x_i\|^2 + O(|K|^2\rho^4)$.
Substituting into \cref{eq:einstein-midpoint},
\[
\mu_{\mathrm{E}}
= \frac{\sum_i (1 + \tfrac{1}{2}|K|\|x_i\|^2)\,x_i + O(|K|^2\rho^5)}
       {\sum_i (1 + \tfrac{1}{2}|K|\|x_i\|^2) + O(|K|^2\rho^4)}
= \bar x + O(|K|\rho^2).
\]
The Klein metric expands as
$g_x^{\KK}(u,v) = \langle u,v\rangle + O(|K|\rho^2)$ near the
origin, so $d^{\KK}(x,y) = \|x-y\| + O(|K|\rho^2)$ and the
Fr\'{e}chet objective becomes
$\sum_i \|x_i-y\|^2 + O(|K|\rho^2)$, whose minimizer is
$\mu_{\mathrm{F}} = \bar x + O(|K|\rho^2)$. The triangle
inequality then gives
$d^{\KK}(\mu_{\mathrm{E}},\mu_{\mathrm{F}}) \le
\|\mu_{\mathrm{E}}-\mu_{\mathrm{F}}\|_2 + O(|K|\rho^2)
= O(|K|\rho^2)$.
\qedhere
\end{proof}

Part (iii) covers two points with equal weights. The running-mean update of \cref{alg:klein-gyrobn} weights its two points by $1-\eta$ and $\eta$, so its result lies on the geodesic between them but need not coincide with the weighted Fr\'{e}chet mean. \cref{prop:einstein-frechet} justifies the midpoint as the batch centre and makes no comparative claim about numerical stability.

In practice the empirical clip $\|x_i\|\le r/\sqrt{-K}$ used in our image-classification pipeline \cref{sec:exp-image} gives $|K|\rho^2 = r^2 \ge 1$ in every clipping setting, so the small-radius regime is not attained and $\mu_{\mathrm{E}},\mu_{\mathrm{F}}$ can differ. The empirical result that the closed-form Einstein midpoint matches the converged Fr\'{e}chet mean within seed noise on the four NC datasets \cref{sec:exp-bn-ablation} shows that the approximation error is small enough not to harm downstream accuracy.

\textbf{Sensitivity to boundary outliers.}
\label{app:midpoint-outlier}
The Einstein midpoint assigns each point a weight $\gamma_{x_i}$ that increases with radius, making
the batch center more sensitive to points near the boundary. \cref{tab:outlier-drift} measures this
effect by injecting one outlier into a batch of $128$ inliers and reporting the resulting geodesic
drift. Within the clipped operating range, the outlier weight is bounded by
$\cosh(\sqrt{-K}r)/(\cosh(\sqrt{-K}r)+m-1)$ and is $2.9\%$ at $r{=}2$ and $m{=}128$; at larger
radii, the Einstein midpoint becomes increasingly sensitive to the outlier. For settings with
unbounded outliers, the midpoint can instead initialize a subsequent Fr\'{e}chet-mean iteration.

\begin{table}[!htbp]
\centering
\caption{Batch-center geodesic drift under one injected outlier, mean over 500 repetitions,
$K{=}{-}1$, inlier radius $\le1$, batch $m{=}128$.}
\label{tab:outlier-drift}
\footnotesize
\begin{tabular}{@{}ccccc@{}}
\toprule
$r_{\text{out}}$ & Einstein midpoint & Fr\'{e}chet mean (converged) &
Tangent mean & Outlier weight share \\
\midrule
$1$ & $0.008$ & $0.007$ & $0.008$ & $1.2\%$ \\
$2$ & $0.024$ & $0.014$ & $0.016$ & $2.9\%$ \\
$4$ & $0.156$ & $0.028$ & $0.031$ & $17.7\%$ \\
$8$ & $1.537$ & $0.057$ & $0.063$ & $92.1\%$ \\
\bottomrule
\end{tabular}
\end{table}

\section{Klein Convolution and Residual Networks}
\label{app:klein-resnet}

This appendix states and proves the results behind \cref{sec:klein-resnet}, compares the cost of the convolutions and verifies the identities numerically.

\subsection{Lorentz--Klein Correspondence}
\label{app:lorentz-klein}

\begin{lemma}[Lorentz--Klein correspondence {\citep[\S6.1]{Ratcliffe2019}, \citep[Eqs.~(11.249)--(11.253)]{Ungar2022}}]
\label[lemma]{lem:klein-lorentz}
The maps $\phi_{\KK\to\LL}$ and $\phi_{\LL\to\KK}$ of \cref{eq:klein-lorentz-map} are mutually inverse isometries between $\KKn$ and $\LL_K^n=\{(x_t,x_s)\in\RR\times\RR^n\mid -x_t^2+\|x_s\|^2=1/K,\ x_t>0\}$. The spatial component $\sigma(x)=\gamma_x x$ is a bijection from $\KKn$ to $\RR^n$ with inverse $\sigma^{-1}(v)=v/\sqrt{1-K\|v\|^2}$, and $\gamma_x^2=1-K\|\sigma(x)\|^2$.
\end{lemma}

For later use, $\|\sigma(x)\|=\gamma_x\|x\|=\sinh\bigl(\sk\,d^{\KK}(x,\mathbf{0})\bigr)/\sk$~\citep[Eq.~(33)]{Bdeir2024}, which follows from $\sk\|x\|=\tanh(\sk\,d)$ and $\gamma_x=\cosh(\sk\,d)$ with $d=d^{\KK}(x,\mathbf{0})$.

The spatial component is also the chart of the PV model $\PV_K^n=\RR^n$ of \citet{Chen2026PVNN}, whose gyroaddition $\oU$ and scalar gyromultiplication $\otU$ form a gyrovector space isomorphic to the M\"{o}bius gyrovector space through $\pi_{\PP\to\PV}(y)=2\gamma_y^2\,y$ with $\gamma_y=1/\sqrt{1+K\|y\|^2}$.

\begin{lemma}[Spatial component and PV model {\citep[Tab.~6.1 and Eq.~(6.327)]{Ungar2022}}]
\label[lemma]{lem:sigma-pv}
The spatial component satisfies $\sigma=\pi_{\PP\to\PV}\circ\phi_{\KK\to\PP}$ with $\phi_{\KK\to\PP}$ of \cref{eq:klein-poincare-map}. Hence, for $x,y\in\KKn$ and $r\in\RR$,
\begin{equation}
\sigma(x\oE y)=\sigma(x)\oU\sigma(y),\qquad
\sigma(r\otE x)=r\otU\sigma(x).
\label{eq:sigma-pv}
\end{equation}
\end{lemma}

By \cref{lem:sigma-pv}, $\oE$ and $\otE$ are the gyro operations of the PV model written in Klein coordinates, whereas $\oS$ and $\otS$ of \cref{eq:klein-res} are the vector operations of $\RR^n$ in the same chart, and the two additions differ in general.

\subsection{Unfold-free Klein Convolution}
\label{app:proof-fcconv}

The Klein convolution of \cref{eq:klein-conv} applies the Klein FC of \cref{thm:klein-fc} to $P_q=\ExpO(U_q)$ with $U_q=\Concat_{i\in\mathcal{N}(q)}\LogO(X_i)\in\RR^{k^2C}$, which literally requires the $k^2C$-dimensional patch of every location. \citet[App.~B]{Bdeir2025Robust} already avoid the patch tensor of the Lorentz convolution by parameterizing the convolution weights as a rotation followed by a boost, which changes the layer, whereas the next result evaluates the Klein FC convolution exactly with two standard convolutions.

\begin{proposition}[Unfold-free FC convolution]
\label[proposition]{prop:fc-conv}
For $U\neq\mathbf{0}$, let $P=\ExpO(U)$ and $a=\sk\|U\|$. The Klein MLR score of \cref{thm:klein-mlr} satisfies
\begin{equation}
v_k(P)=\frac{\|z_k\|}{\sk}\arcsinh\!\left(
\cosh(\sk r_k)\,\frac{\sinh a}{\|U\|}\,\frac{\langle U,z_k\rangle}{\|z_k\|}
-\sinh(\sk r_k)\cosh a\right).
\label{eq:fc-conv-score}
\end{equation}
Moreover $\|U_q\|^2=(\mathbf{1}\star\|\LogO(X)\|^2)_q$ and $\langle U_q,z_k\rangle=(z_k\star\LogO(X))_q$, where $\star$ is a standard convolution and $\mathbf{1}$ the all-ones $k\times k$ kernel. Hence the FC convolution is evaluated with two convolutions and pointwise operations, without forming the patches.
\end{proposition}

\textit{Proof.} Since $P=\ExpO(U)$ has norm $\|P\|=\tanh(a)/\sk$, we have $1+K\|P\|^2=1-\tanh^2a=\cosh^{-2}a$, hence $\gamma_P=\cosh a$. Moreover $\langle P,z\rangle=\tanh(a)\langle U,z\rangle/(\sk\|U\|)$, so
\begin{equation}
\gamma_P\,\frac{\sk}{\|z\|}\langle P,z\rangle=\frac{\sinh a}{\|U\|}\,\frac{\langle U,z\rangle}{\|z\|}.
\end{equation}
Substituting $\gamma_P$ and this identity into \cref{eq:klein-mlr} gives \cref{eq:fc-conv-score}. For the convolutional form, $\|U_q\|^2=\sum_{i\in\mathcal{N}(q)}\|\LogO(X_i)\|^2$ is the convolution of the single-channel map $\|\LogO(X)\|^2$ with the all-ones kernel, and $\langle U_q,z_k\rangle=\sum_{i\in\mathcal{N}(q)}\langle\LogO(X_i),z_{k,i}\rangle$ is the convolution of $\LogO(X)$ with $z_k$ reshaped to a $k\times k\times C$ kernel. The Klein FC of \cref{thm:klein-fc} depends on its input only through the scores $v_1,\dots,v_m$, which completes the proof. Finally $\sinh(a)/\|U\|=\sk\,\sinh(a)/a\to\sk$ as $U\to\mathbf{0}$.
\hfill$\square$

Implementations should evaluate $\sinh(a)/\|U\|$ by this limit at $U=\mathbf{0}$, since a small constant added to $\|U\|$ underflows in half precision and turns all-zero receptive fields into $0/0$.

\textbf{Tangent variant.}
Replacing the Klein FC by a linear map in the tangent space at the origin gives
\begin{equation}
(\mathcal{C}_{\mathrm{t}}X)_q=\ExpO\Bigl(\sum_{i\in\mathcal{N}(q)}W_i\,\LogO(X_i)+b\Bigr)
\label{eq:klein-conv-tangent}
\end{equation}
with $W_i\in\RR^{C'\times C}$ and $b\in\RR^{C'}$. Its inner sum is one standard convolution of $\LogO(X)$, so it costs one Euclidean convolution and two pointwise maps (\cref{tab:kres-cost}). The sign of each output coordinate is that of an affine function of the lifted patch in the tangent space. For $b\neq\mathbf{0}$ its zero set is the image under $\ExpO$ of an affine hyperplane that misses the origin, which is curved in the Klein ball and therefore not a Klein hyperplane, whereas \cref{eq:klein-conv} separates by Klein hyperplanes at any offset, as observed for the Poincar\'{e} FC by \citet[\S3.2]{Shimizu2021}.

\subsection{Radial Form of Hyperbolic ReLU}
\label{app:proof-relu}

Besides the pointwise ReLU of \cref{sec:klein-resnet}, which keeps every point in the ball as norm-decreasing activations do on the Poincar\'{e} ball~\citep[\S3.2]{Liu2019HGNN}, two rectifiers appear in the literature, the tangent ReLU $\ExpO\circ\ReLU\circ\LogO$, which is the Einstein version of ReLU~\citep[Def.~8]{Mao2024}, and the Lorentz ReLU of HCNN~\citep[Eq.~(13)]{Bdeir2024}, which rectifies the spatial component and recomputes the time coordinate and is the spatial version $\sigma^{-1}\circ\ReLU\circ\sigma$. With $d^{\KK}(x,\mathbf{0})=\arctanh(\sk\|x\|)/\sk$, the three share the direction of $\ReLU(x)$ and differ only in their radial behavior.

\begin{proposition}[Radial form of hyperbolic ReLU]
\label[proposition]{prop:relu}
Let $x\in\KKn$ with $\ReLU(x)\neq\mathbf{0}$, $\rho=\|\ReLU(x)\|/\|x\|$ and $d=d^{\KK}(x,\mathbf{0})$. The coordinate, tangent and Lorentz ReLU map $x$ to points on the ray through $\ReLU(x)$ at distances $d_{\mathrm{c}}$, $d_{\mathrm{t}}$ and $d_{\mathrm{L}}$ from the origin, with
\begin{equation}
\tanh(\sk\,d_{\mathrm{c}})=\rho\tanh(\sk\,d),\qquad
d_{\mathrm{t}}=\rho\,d,\qquad
\sinh(\sk\,d_{\mathrm{L}})=\rho\sinh(\sk\,d).
\label{eq:relu-radial}
\end{equation}
Consequently $d_{\mathrm{c}}\le d_{\mathrm{t}}\le d_{\mathrm{L}}$.
\end{proposition}

The three rectifiers thus rescale the Klein radius, the geodesic distance and the Lorentz spatial norm by the same factor $\rho$.

\textit{Proof.} Write $R=\ReLU(x)$, so that $\|R\|=\rho\|x\|$ and $\sk\|x\|=\tanh(\sk\,d)$.

\textit{Pointwise ReLU.} The output $R$ lies on its own ray and $\tanh(\sk\,d_{\mathrm{c}})=\sk\|R\|=\rho\tanh(\sk\,d)$.

\textit{Tangent ReLU.} $\LogO(x)=c\,x$ with $c=\arctanh(\sk\|x\|)/(\sk\|x\|)>0$. Positive homogeneity of ReLU gives $\ReLU(cx)=cR$, whose norm is $\rho\,c\|x\|=\rho\,d$. The map $\ExpO$ preserves directions and sends a tangent vector of norm $t$ to a point at distance $t$, hence $d_{\mathrm{t}}=\rho\,d$.

\textit{Lorentz ReLU.} By \cref{lem:klein-lorentz} the spatial component of $\phi_{\KK\to\LL}(x)$ is $\gamma_x x$ and its rectification is $\gamma_x R$, a positive multiple of $R$ with norm $\rho\,\gamma_x\|x\|=\rho\sinh(\sk\,d)/\sk$~\citep[Eq.~(33)]{Bdeir2024}. Recomputing the time coordinate and mapping back gives $\sigma^{-1}(\gamma_x R)$, which lies on the ray of $R$ at distance $d_{\mathrm{L}}$ with $\sinh(\sk\,d_{\mathrm{L}})=\sk\|\gamma_x R\|=\rho\sinh(\sk\,d)$.

\textit{Ordering.} For $f$ with $f(0)=0$, concavity on $[0,\infty)$ implies $f(\rho t)\ge\rho f(t)$ and convexity implies $f(\rho t)\le\rho f(t)$ for $\rho\in[0,1]$. With $f=\tanh$, $\tanh(\sk\,d_{\mathrm{t}})=\tanh(\rho\sk\,d)\ge\rho\tanh(\sk\,d)=\tanh(\sk\,d_{\mathrm{c}})$, so $d_{\mathrm{t}}\ge d_{\mathrm{c}}$. With $f=\sinh$, $\sinh(\sk\,d_{\mathrm{t}})\le\rho\sinh(\sk\,d)=\sinh(\sk\,d_{\mathrm{L}})$, so $d_{\mathrm{t}}\le d_{\mathrm{L}}$.
\hfill$\square$

In Klein coordinates the Lorentz ReLU equals $\lambda_x\ReLU(x)$ with $\lambda_x=\gamma_x/\sqrt{1-K\gamma_x^2\|\ReLU(x)\|^2}\in[1,\gamma_x]$, with $\lambda_x=1$ exactly when $\ReLU(x)=x$.

\subsection{Klein Residual Connection}
\label{app:proof-res}

\begin{proposition}[Space addition {\citep[\S4.4]{Bdeir2024}}]
\label[proposition]{prop:res}
For $x,y\in\KKn$ and $r\in\RR$, the points $x\oS y$ and $r\otS x$ of \cref{eq:klein-res} lie in $\KKn$, and $(\KKn,\oS,\otS)$ is a real vector space isomorphic to $\RR^n$ through $\sigma$. The point $x\oS y$ is the image under the isometry of \cref{lem:klein-lorentz} of the HCNN residual of the corresponding Lorentz points.
\end{proposition}

Unlike the left gyrotranslation $y\mapsto\beta\oE y$, the translation $y\mapsto y\oS\beta$ is not an isometry. It plays the role of vector addition in the spatial chart, in the same way as the Euclidean residual connection.

The Einstein addition $x\oE\mathcal{G}(x)$ offers a gyrovector alternative, but it is neither commutative nor the transport of the HCNN residual. For the same reason the biasing of \cref{eq:klein-gyrobn} acts on the left, since $y\mapsto\beta\oE y$ is an isometry while $y\oE\beta=\gyr[y,\beta](\beta\oE y)$ adds a $y$-dependent gyration that breaks \cref{eq:bn-homog-mean}.

\textbf{Comparison with the Einstein addition.}
HCNN~\citep[App.~Tab.~11]{Bdeir2024} found the space addition $\oS$ on par with the best of five tested additions. In Klein-ResNet, $\oS$ is more accurate than $\oE$ under equal tuning, is comparable in accuracy to the tangent version and gives a slightly shorter epoch than both (\cref{tab:res-ablation,tab:ops-ablation}). In HCNN the M\"{o}bius addition, the Poincar\'{e} counterpart of $\oE$, diverges in training~\citep[App.~Tab.~11]{Bdeir2024}, so we compare $\oS$ with $\oE$ directly in Klein-ResNet. Both residual connections are tuned on the validation split of \cref{app:resnet-exp} over $K\in\{-1,-0.3,-0.1\}$ and $\nu\in\{1,0.5,0.25\}$ with the convolution of \cref{eq:klein-conv}. Every configuration is screened with one seed, the two leading ones are rerun with three seeds, and the rule selects $K=-0.3$ and $\nu=0.25$ for both. \cref{tab:res-ablation} reports this configuration on the test set with five new seeds. Adding spatial components is $0.26$ points more accurate, with a one-sided Welch $p$ of $0.022$, and in a single timing run it takes $15.9$\,s per epoch against $16.5$\,s for $\oE$ (\cref{tab:ops-ablation}). The main model of \cref{tab:resnet} uses the input scale of \cref{tab:resnet-hparams}.

\begin{table}[h]
\centering
\small
\caption{Residual connections of Klein-ResNet on CIFAR-100 under equal tuning ($K=-0.3$, $\nu=0.25$), five seeds. Best is the best test accuracy during training, final the accuracy after the last epoch.}
\label{tab:res-ablation}
\begin{tabular}{lcc}
\toprule
Residual connection & Best & Final \\
\midrule
Einstein addition $\oE$ & $77.39\pm0.20$ & $77.25\pm0.20$ \\
Spatial addition \cref{eq:klein-res} & $77.64\pm0.13$ & $77.48\pm0.11$ \\
\bottomrule
\end{tabular}
\end{table}

\subsection{Correspondence with HCNN}
\label{app:proof-hcnn}

\begin{proposition}[Correspondence with HCNN]
\label[proposition]{prop:klein-hcnn}
Under the isometry of \cref{lem:klein-lorentz}, the input lift $X_q=\sigma^{-1}(\nu x_q)$ of \cref{sec:klein-resnet} with $\nu=1$, the Einstein-midpoint pooling, the residual connection \cref{eq:klein-res}, the centering, scaling and biasing steps of the Klein GyroBN and the Klein MLR head coincide with the corresponding operations of HCNN~\citep{Bdeir2024}. The two networks differ only in the convolution, in the radial part of the activation (\cref{prop:relu}) and, for the reference implementation of HCNN, in the dispersion statistic of batch normalization, which is the mean distance there and the root mean squared distance in Klein GyroBN and in \citet[Eq.~(6)]{Bdeir2024}.
\end{proposition}

The accuracies of the two networks can therefore differ only through these three components, and all their other differences are computational, which \cref{sec:exp-resnet} measures.

\textit{Proof.} \textit{Input lift.} HCNN prepends a time coordinate to each pixel vector $x_q$ and projects onto $\LL_K^n$, which yields the Lorentz point with spatial component $x_q$. Its Klein image is $\sigma^{-1}(x_q)$, which is the input lift with $\nu=1$.

\textit{Pooling.} The Lorentz centroid normalizes $\sum_i\phi_{\KK\to\LL}(x_i)$ onto $\LL_K^n$ by a positive factor, and $\phi_{\LL\to\KK}$ is invariant to positive rescaling of its argument. Hence its Klein image is $\bigl(\sum_i\gamma_{x_i}x_i\bigr)/\bigl(\sum_i\gamma_{x_i}\bigr)$, the Einstein midpoint~\citep[Thm.~1]{Shimizu2021}. The same identity shows that the batch mean of HCNN batch normalization maps to the Einstein midpoint.

\textit{Batch normalization steps.} HCNN centers by $x\mapsto\Expop_{\mathbf{0}}\bigl(\PT_{\mu\to\mathbf{0}}(\Logop_\mu x)\bigr)$, rescales by $x\mapsto\Expop_{\mathbf{0}}(t\Logop_{\mathbf{0}}x)$ and biases by $y\mapsto\Expop_\beta\bigl(\PT_{\mathbf{0}\to\beta}(\Logop_{\mathbf{0}}y)\bigr)$. The gyrovector identity $\Expop_p\bigl(\PT_{\mathbf{0}\to p}(v)\bigr)=p\oE\ExpO(v)$ of \cref{prop:gyro-riemannian}, together with left cancellation $(-p)\oE(p\oE w)=w$, turns the first and third maps into $x\mapsto(-\mu)\oE x$ and $y\mapsto\beta\oE y$, and the scalar identity $t\otE x=\ExpO(t\LogO x)$ of the same proposition turns the second into $x\mapsto t\otE x$. These are the three steps of the Klein GyroBN. Only the dispersion entering $t$ differs from the reference implementation, as stated.

\textit{Residual connection.} This is \cref{prop:res}.

\textit{Classifier.} The Lorentz MLR of HCNN with parameters $(z_k,a_k)$ outputs $(\|z_k\|/\sk)\arcsinh(\sk\,\alpha_k/\|z_k\|)$ with $\alpha_k=\cosh(\sk\,a_k)\langle x_s,z_k\rangle-\sinh(\sk\,a_k)\|z_k\|\,x_t$. Substituting $x_t=\gamma_x/\sk$ and $x_s=\gamma_x x$ gives
\begin{equation}
\frac{\sk\,\alpha_k}{\|z_k\|}=\gamma_x\!\left(\cosh(\sk\,a_k)\frac{\sk}{\|z_k\|}\langle x,z_k\rangle-\sinh(\sk\,a_k)\right),
\end{equation}
which is the argument of the Klein MLR of \cref{eq:klein-mlr} with $r_k=a_k$.

\textit{Remaining components.} The HCNN convolution applies the Lorentz fully connected layer of~\citet{Chen2022LGCN} to the Lorentz direct concatenation of each receptive field, which is not the transport of \cref{eq:klein-conv} or of its tangent variant \cref{eq:klein-conv-tangent}, and the activations differ in their radial part by \cref{prop:relu}.
\hfill$\square$

\subsection{Operator Ablation}
\label{app:ops-ablation}

\cref{tab:ops-ablation} replaces one operation of Klein-ResNet at a time by another construction of \cref{sec:klein-resnet}. The base model uses the tangent concatenation with the Klein FC of \cref{eq:klein-conv}, the pointwise ReLU, $\oS$ and the input lift $\nu\otS\sigma^{-1}(x_q)$, with $K=-0.3$ and $\nu=0.25$ as in \cref{tab:res-ablation}. The spatial version of the concatenation is the Lorentz direct concatenation of HCNN followed by the Klein FC. The tangent version of the whole convolution is \cref{eq:klein-conv-tangent}, and its spatial version is $\sigma^{-1}\bigl(\sum_{i\in\mathcal{N}(q)}W_i\,\sigma(X_i)+b\bigr)$. The tangent version of the input lift is $\ExpO(\nu x_q)=\nu\otE\ExpO(x_q)$ by \cref{eq:scalar-riemannian}, so it also covers $\otE$, and transporting the PV gyro operations by $\sigma$ returns $\oE$ and $\otE$ by \cref{lem:sigma-pv}. Each row is trained with seeds $30$ to $34$ under the protocol of \cref{app:resnet-exp} without further tuning.

\begin{table}[h]
\centering
\small
\caption{Klein-ResNet on CIFAR-100 with one operation replaced at a time ($K=-0.3$, $\nu=0.25$), seeds $30$ to $34$. Best and final are as in \cref{tab:res-ablation}.}
\label{tab:ops-ablation}
\begin{tabular}{llccc}
\toprule
Operation & Replaced by & Best & Final & $t_{\text{epoch}}$ (s) \\
\midrule
\multicolumn{2}{l}{Base model} & $77.80\pm0.12$ & $77.63\pm0.16$ & $15.9$ \\
\midrule
Concatenation & spatial version & $77.85\pm0.22$ & $77.69\pm0.22$ & -- \\
Convolution & tangent version \cref{eq:klein-conv-tangent} & $77.94\pm0.17$ & $77.75\pm0.16$ & $12.3$ \\
 & spatial version & $77.93\pm0.10$ & $77.74\pm0.15$ & $13.6$ \\
Activation & tangent version & $77.60\pm0.30$ & $77.42\pm0.31$ & $16.8$ \\
 & spatial version (Lorentz ReLU) & $77.24\pm0.29$ & $77.08\pm0.27$ & $16.6$ \\
Addition & tangent version & $77.83\pm0.34$ & $77.65\pm0.30$ & $16.3$ \\
 & $\oE$ & $77.39\pm0.20$ & $77.25\pm0.20$ & $16.5$ \\
Input lift & tangent version & $77.75\pm0.29$ & $77.65\pm0.31$ & $15.6$ \\
\bottomrule
\end{tabular}
\end{table}

Replacing the convolution by its tangent or spatial version changes the best accuracy by at most $0.14$ and shortens an epoch from $15.9$\,s to $12.3$\,s and $13.6$\,s. The pointwise ReLU gives the highest mean accuracy and the shortest epoch of the three rectifiers, $0.20$ and $0.56$ points above its tangent and spatial versions. Per epoch, $\oS$ takes $15.9$\,s against $16.3$\,s for its tangent version and $16.5$\,s for $\oE$. The input lift and the concatenation change the best accuracy by at most $0.05$. Each row changes one operation of a single configuration without tuning, so these differences describe this configuration only. The $t_{\text{epoch}}$ column comes from a single timing run with the protocol of \cref{tab:resnet}, and the $\oE$ row repeats \cref{tab:res-ablation}, whose seeds $25$ to $29$ give $77.64\pm0.13$ for the base model.

\subsection{Cost per Layer}
\label{app:kres-cost}

\cref{tab:kres-cost} lists the cost of one $k\times k$ convolution with $C$ input and $C'$ output channels on $|\Omega|$ locations. All variants share the dense multiply-accumulate term. They differ in the pointwise work and in the intermediate tensors kept for the backward pass.

\begin{table}[h]
\centering
\small
\caption{Cost of one $k\times k$ convolution on $|\Omega|$ locations with $C$ input and $C'$ output channels. Pointwise operations count transcendental and elementwise calls.}
\label{tab:kres-cost}
\begin{tabular}{lccc}
\toprule
Layer & Dense term & Pointwise term & Largest intermediate \\
\midrule
Euclidean & $|\Omega|k^2CC'$ & none & $|\Omega|C'$ \\
Klein, \cref{eq:klein-conv} by \cref{prop:fc-conv} & $|\Omega|k^2CC'+|\Omega|k^2$ & $O(|\Omega|(C+C'))$ & $|\Omega|\max(C,C')$ \\
Klein tangent variant, \cref{eq:klein-conv-tangent} & $|\Omega|k^2CC'$ & $O(|\Omega|(C+C'))$ & $|\Omega|\max(C,C')$ \\
HCNN, reference implementation & $|\Omega|(k^2C{+}1)(C'{+}1)$ & $O(|\Omega|(k^2+C'))$ & $|\Omega|k^2(C{+}1)$ \\
\bottomrule
\end{tabular}
\end{table}

\begin{remark}
The patch tensor in the last row comes from the reference implementation of HCNN, which unfolds every receptive field before the Lorentz fully connected layer. It is not intrinsic to the Lorentz model. The spatial part of the Lorentz direct concatenation enters the fully connected layer linearly, and its time coordinate $\bigl(\sum_i x_{t,i}^2+(k^2-1)/K\bigr)^{1/2}$ is a single-channel convolution of $x_t^2$ followed by a pointwise square root, so the same two-convolution evaluation as in \cref{prop:fc-conv} applies. The speedups of \cref{sec:exp-resnet} therefore combine the Klein formulation and the implementation.
\end{remark}

\subsection{Numerical Verification}
\label{app:kres-numeric}

\cref{tab:kres-numeric} checks each statement against an independent computation in float64 unless noted, with random inputs at $K\in\{-1,-0.7\}$.

\begin{table}[h]
\centering
\small
\caption{Maximum absolute deviation between each statement and an independent computation.}
\label{tab:kres-numeric}
\begin{tabular}{ll}
\toprule
Statement & Deviation \\
\midrule
\cref{prop:fc-conv}, two convolutions against per-patch evaluation & $3\times10^{-14}$ \\
\cref{prop:relu}, common direction of the three rectifiers & $3\times10^{-16}$ \\
\cref{prop:relu}, ordering $d_{\mathrm{c}}\le d_{\mathrm{t}}\le d_{\mathrm{L}}$ & $0$ violations in $10^4$ points \\
\cref{lem:sigma-pv}, $\sigma$ against $\pi_{\PP\to\PV}\circ\phi_{\KK\to\PP}$ & $2\times10^{-15}$ \\
\cref{lem:sigma-pv}, \cref{eq:sigma-pv} for $\oE$ and for $\otE$ & $3\times10^{-13}$, $2\times10^{-10}$ \\
\cref{prop:res}, $x\oS y$ against the transported HCNN residual & $2\times10^{-16}$ \\
\cref{prop:klein-hcnn}, Einstein midpoint against Lorentz centroid (float32) & $3\times10^{-8}$ \\
\cref{prop:klein-hcnn}, GyroBN steps against HCNN batch normalization & $1.4\times10^{-15}$ \\
\cref{prop:klein-hcnn}, Klein MLR against transported HCNN Lorentz MLR & $9\times10^{-14}$ \\
\bottomrule
\end{tabular}
\end{table}

\section{Proofs}
\label{app:proofs}

\subsection{Proof of \cref{thm:klein-hyp}}
\label{app:proof-hyp}

We prove the hyperplane equivalence and the point-to-hyperplane distance in turn.

\textbf{Hyperplane equivalence.}
Set $w = \PT_{p\to\mathbf{0}}^{\KK}(a)\in T_{\mathbf{0}}\KKn$. Since parallel transport is a linear isometry,
\begin{equation}
  \bigl\langle\Logop_p^{\KK}(x),\,a\bigr\rangle_p
  =
  \bigl\langle\PT_{p\to\mathbf{0}}^{\KK}\!\bigl(\Logop_p^{\KK}(x)\bigr),\;w
  \bigr\rangle_{\mathbf{0}}.
\label{eq:pt-isometry}
\end{equation}
The gyrovector identity~\citep{Ungar2022} gives $\PT_{p\to\mathbf{0}}^{\KK}\bigl(\Logop_p^{\KK}(x)\bigr) = \Logop_{\mathbf{0}}^{\KK}\bigl((-p)\oE x\bigr)$. At the origin, $g_{\mathbf{0}}^{\KK}(u,v)=\langle u,v\rangle$ and $\Logop_{\mathbf{0}}^{\KK}(z)=\tfrac{1}{\sqrt{-K}}\arctanh(\sqrt{-K}\|z\|) \tfrac{z}{\|z\|}$ is a positive scalar multiple of $z$. Substituting into \cref{eq:pt-isometry}, \[ \bigl\langle\Logop_p^{\KK}(x),\,a\bigr\rangle_p = 0
  \iff
\bigl\langle (-p)\oE x,\, w\bigr\rangle = 0, \] which is \cref{eq:klein-hyp}.

\textbf{Signed point-to-hyperplane distance.}
Set $z = (-p)\oE y$. Left Einstein translation $L_\beta(x)=\beta\oE x$ is a Klein isometry~\citep[Thm.~16]{Chen2025gyrobn}, so \[ d^{\KK}(y, H_{a,p}) = d^{\KK}\bigl(z,\;H_{w,\mathbf{0}}\bigr), \qquad H_{w,\mathbf{0}}=\{x\in\KKn\mid\langle x,w\rangle=0\}. \] We compute the right-hand side via the Klein--Lorentz isometry. See \cref{eq:klein-lorentz-map}. Define \[ z_{\LL} = \bigl(\gamma_z/\sqrt{-K},\;\gamma_z\,z\bigr)\in\LL_K^n, \qquad n_{\LL} = (0,\,w)\in\RR^{1+n}. \] Since $x_{\KK}=x_s/(\sqrt{-K}\,x_t)$ with $\sqrt{-K}\,x_t>0$, the equation $\langle x_{\KK},w\rangle=0$ on $\KKn$ is equivalent to $\langle x_s,w\rangle=0$ on $\LL_K^n$, so $H_{w,\mathbf{0}}$ lifts to the totally geodesic submanifold $S=\LL_K^n\cap\{x\mid\langle x_s,w\rangle=0\}$ orthogonal to $n_{\LL}$.

By the standard signed point-to-submanifold formula on the Lorentz hyperboloid, \[ d^{\KK}(z, H_{w,\mathbf{0}}) = \frac{1}{\sqrt{-K}}\,\arcsinh\!\left( \frac{\sqrt{-K}\, \bigl|\langle z_{\LL},\,n_{\LL}\rangle_{\LL}\bigr|} {\|n_{\LL}\|_{\LL}} \right). \] A direct computation gives $\langle z_{\LL}, n_{\LL}\rangle_{\LL} = \langle\gamma_z z,\,w\rangle = \gamma_z\,\langle z,w\rangle$ and $\|n_{\LL}\|_{\LL} = \|w\|$, which yields \[ d^{\KK}(y, H_{a,p}) = \frac{1}{\sqrt{-K}}\,\arcsinh\!\left( \frac{\sqrt{-K}\,\gamma_z\,|\langle z,w\rangle|}{\|w\|} \right), \] matching \cref{eq:klein-dist}.\qed

\subsection{Proof of \cref{thm:klein-mlr}}
\label{app:proof-mlr}

\textbf{Setup.}
The unidirectional parameterization $p_k = \Expop_{\mathbf{0}}^{\KK}(r_k\hat z_k)$, $a_k = \PT_{\mathbf{0}\to p_k}^{\KK}(z_k)$ implies, by the isometry of parallel transport, $w_k:=\PT_{p_k\to\mathbf{0}}^{\KK}(a_k) = z_k$ and $\|a_k\|_{p_k} = \|z_k\|$. Substituting these into the signed-margin form \cref{eq:klein-mlr-raw} and using $\sign(u)\arcsinh(|u|)=\arcsinh(u)$,
\begin{equation}
  v_k(x)
  = \frac{\|z_k\|}{\sqrt{-K}}\,
    \arcsinh\!\left(
      \frac{\sqrt{-K}\,\gamma_z\,\langle z,z_k\rangle}{\|z_k\|}
    \right),
  \qquad z = (-p_k)\oE x.
\label{eq:mlr-step1}
\end{equation}
It remains to evaluate $\gamma_z\langle z,z_k\rangle$ in closed form.

\textbf{Lift to the Lorentz model.}
The Klein--Lorentz isometry \cref{eq:klein-lorentz-map} lifts $x\in\KKn$ to $x_{\LL}=(\gamma_x/\sqrt{-K},\,\gamma_x x)\in\LL_K^n$. Under this lift the foot point becomes \[ (p_k)_{\LL} = \tfrac{1}{\sqrt{-K}}\bigl(\cosh(\sqrt{-K}\,r_k),\, \sinh(\sqrt{-K}\,r_k)\,\hat z_k\bigr), \] and Einstein subtraction $z=(-p_k)\oE x$ corresponds to a Lorentz boost of $x_{\LL}$ along $\hat z_k$ with rapidity $\rho=\sqrt{-K}\,r_k$.

\textbf{Compute the boost.}
Decompose $x$ as $x = \langle x,\hat z_k\rangle\,\hat z_k + x_\perp$. Applying the boost to $x_{\LL}$ and reading off the spatial component along $\hat z_k$,
\begin{equation*}
  z_s\cdot\hat z_k
  = \gamma_x\,\langle x,\hat z_k\rangle\cosh(\sqrt{-K}\,r_k)
    - \tfrac{\gamma_x}{\sqrt{-K}}\sinh(\sqrt{-K}\,r_k).
\end{equation*}
The Klein--Lorentz lift also gives $z_s\cdot\hat z_k = \gamma_z\, \langle z,\hat z_k\rangle$. Multiplying both sides by $\|z_k\|$ and using $z_k=\|z_k\|\hat z_k$,
\begin{equation*}
  \gamma_z\,\langle z,z_k\rangle
  = \gamma_x\!\left[
      \cosh(\sqrt{-K}\,r_k)\,\langle x,z_k\rangle
      - \tfrac{\|z_k\|}{\sqrt{-K}}\,\sinh(\sqrt{-K}\,r_k)
    \right].
\end{equation*}

\textbf{Substitute back.}
Dividing through by $\|z_k\|$ and inserting the result into \cref{eq:mlr-step1}, \[ \frac{\sqrt{-K}\,\gamma_z\langle z,z_k\rangle}{\|z_k\|} = \gamma_x\!\left( \cosh(\sqrt{-K}\,r_k)\, \frac{\sqrt{-K}}{\|z_k\|}\langle x,z_k\rangle -\sinh(\sqrt{-K}\,r_k) \right), \] which is precisely the bracketed argument of \cref{eq:klein-mlr}.

\textbf{Euclidean limit.}
As $K\to 0^-$, $\gamma_x\to 1$, $\cosh(\sqrt{-K}\,r_k)\to 1$, $\sinh(\sqrt{-K}\,r_k)\to\sqrt{-K}\,r_k$, and $\tfrac{\|z_k\|}{\sqrt{-K}}\arcsinh(\sqrt{-K}\,u)\to\|z_k\|\,u$. Combining, $v_k(x)\to\langle x,z_k\rangle - r_k\|z_k\|$, recovering the Euclidean MLR \cref{eq:euc-mlr} with bias $b_k=-r_k\|z_k\|$.\qed

\textbf{Batch-efficient computation.}
\cref{eq:klein-mlr} depends on $x$ only through the inner product $\langle x,z_k\rangle$ and the scalar $\|x\|^2$ (via $\gamma_x$), so it is implementable as a single matmul $XZ$ over the batch. The raw form \cref{eq:klein-mlr-raw} would instead require materializing a $b\!\times\!C\!\times\!n$ tensor $\{(-p_k)\oE x_i\}_{i,k}$, which causes out-of-memory failures at high $b$ or $C$.

\subsection{Klein BMLR: derivation of the expanded form}
\label{app:proof-bmlr}

\textbf{Closed form.}
The Klein Busemann function $B^{v_k}_{\KK}(x)$ is given by \cref{eq:klein-busemann}, derived in \cref{app:busemann-verify} via the Klein--Lorentz isometry and cross-checked through the Klein--Poincar\'{e} isometry. Substituting into $u_k(x) = -\alpha_k B^{v_k}_{\KK}(x) + b_k$ and expanding the log, \[ u_k(x) = \tfrac{\alpha_k}{\sqrt{-K}}\!\left[ \tfrac{1}{2}\log(1+K\|x\|^2) - \log(1-\sqrt{-K}\langle x,v_k\rangle) \right] + b_k, \] which is \cref{eq:klein-bmlr-expanded}.

\textbf{Euclidean limit.}
A first-order Taylor expansion as $K\to 0^-$ gives
\begin{align*}
  \tfrac{1}{2}\log(1+K\|x\|^2) &= O(K),\\
  \log\bigl(1-\sqrt{-K}\langle x,v_k\rangle\bigr)
   &= -\sqrt{-K}\langle x,v_k\rangle + O(-K).
\end{align*}
Substituting, $u_k(x) \to \tfrac{\alpha_k}{\sqrt{-K}}\cdot\sqrt{-K}\langle v_k,x\rangle + b_k = \alpha_k\langle v_k,x\rangle + b_k$, recovering the Euclidean MLR \cref{eq:euc-mlr} exactly.\qed

\subsection{Proof of \cref{thm:klein-fc} (Klein FC and BFC)}
\label{app:proof-fc}

\textbf{Reduce defining equation at the origin.}
Setting $p=\mathbf{0}$ in \cref{eq:klein-hyp} gives $H_{e_k,\mathbf{0}}=\{y\in\KK_K^m\mid y_k=0\}$ and the offset variable $z=(-\mathbf{0})\oE y = y$. Applying \cref{eq:klein-dist} with $w=e_k$ to the defining equation~\eqref{eq:klein-fc-def} of $\mathcal{F}$, \[ \frac{1}{\sqrt{-K}}\arcsinh\!\bigl(\sqrt{-K}\,\gamma_y\,y_k\bigr) = s_k(x), \] where we have absorbed the sign via $\sign(y_k)\arcsinh(|u|)=\arcsinh(u)$.

\textbf{Solve coordinate-wise.}
Inverting $\arcsinh$,
\begin{equation}
  \gamma_y\,y_k
  = \frac{1}{\sqrt{-K}}\sinh\!\bigl(\sqrt{-K}\,s_k(x)\bigr),
  \qquad k=1,\ldots,m.
\label{eq:fc-step1}
\end{equation}

\textbf{Solve for $\gamma_y$.}
Squaring and summing \cref{eq:fc-step1} over $k$, \[ \gamma_y^2\,\|y\|^2 = S, \qquad S := \tfrac{1}{-K}\sum_{j=1}^{m}\sinh^2\!\bigl(\sqrt{-K}\,s_j(x)\bigr). \] Since $\gamma_y^2 = 1/(1+K\|y\|^2)$, we have $\gamma_y^2\|y\|^2 = \|y\|^2/(1+K\|y\|^2)$. Writing $u=\|y\|^2$, the identity $u/(1-(-K)u)=S$ yields $u = S/(1-K S)$ and hence \[ 1 + Ku = \frac{1}{1-KS}, \qquad \gamma_y = \sqrt{1-KS} = \sqrt{1+\sum_{j=1}^{m}\sinh^2\!\bigl(\sqrt{-K}\,s_j(x)\bigr)}. \]

\textbf{Conclude.}
Dividing \cref{eq:fc-step1} by $\gamma_y$ gives \cref{eq:klein-fc}. The BFC case is identical with $s_k=u_k$ in place of $v_k$.

\textbf{Euclidean limit.}
As $K\to 0^-$, $\sinh(\sqrt{-K}\,s_j)\to\sqrt{-K}\,s_j$ and $\sum_j\sinh^2(\sqrt{-K}\,s_j)\to 0$, so \cref{eq:klein-fc} reduces to $y_k = s_k(x)$, the Euclidean affine map.\qed

\end{document}